\documentclass[review]{elsarticle}

\journal{Journal of \LaTeX\ Templates}

\usepackage{graphics}

\usepackage{xspace}
\usepackage{amsmath, amssymb,latexsym,amsfonts,amstext}
\usepackage{multirow}
\usepackage{listings, graphicx}
\usepackage{wrapfig}
\usepackage{color}
\usepackage{float}
\usepackage{hyperref}

\usepackage[caption=false,font=footnotesize]{subfig}

\usepackage{xcolor}
\usepackage{xparse}

\NewDocumentCommand{\codeword}{v}{%
\texttt{\textcolor{blue}{#1}}%
}
\newcommand\tab[1][1cm]{\hspace*{#1}}

\begin{document} \sloppy

\begin{frontmatter}

\title{Image Segmentation using Multi-Threshold technique by Histogram Sampling}

\author{Amit Gurung\corref{mycorrespondingauthor}}

\ead{rajgurung777@gmail.com}
\cortext[mycorrespondingauthor]{Corresponding author}

\author{Sangyal Lama Tamang}
\address{Department of Computer Science,}
\address{Martin Luther Christian University, Shillong, Meghalaya, India}




\begin{abstract}

The segmentation of digital images is one of the essential steps in image processing or a computer vision system. It helps in separating the pixels into different regions according to their intensity level. A large number of segmentation techniques have been proposed, and a few of them use complex computational operations. Among all, the most straightforward procedure that can be easily implemented is thresholding. In this paper, we present a unique heuristic approach for image segmentation that automatically determines multilevel thresholds by sampling the histogram of a digital image. Our approach emphasis on selecting a valley as optimal threshold values. We demonstrated that our approach outperforms the popular Otsu's method in terms of CPU computational time.  We demonstrated that our approach outperforms the popular Otsu's method in terms of CPU computational time. We observed a maximum speed-up of $35.58\times$ and a minimum speed-up of $10.21\times$ on popular image processing benchmarks. To demonstrate the correctness of our approach in determining threshold values, we compute PSNR, SSIM, and FSIM values to compare with the values obtained by Otsu's method. This evaluation shows that our approach is comparable and better in many cases as compared to well known Otsu's method.

\end{abstract}



\end{frontmatter}


\section{Introduction} \label{sec:introduction}


In most computer vision systems, one of the essential preprocessing tasks is the image segmentation. The reliability of the outputs depends on the quality of the input image provided by the image preprocessing. Thus, research is progressing in the direction of enhancing the quality of input images to eliminate noise, visual artifacts, and redundancy of information. One of the most used techniques to handle these issues is Image Segmentation. It is the process of grouping pixels into different groups or segments in an image. Each such group represents an object in an image providing a better understanding of the objects in the given image.

Recently, image segmentation have been applied in a number of areas such as Medical Imaging for the detection of brain tumor or the study of brain development of neonatal brain from Magnetic Resonance Imaging (MRI) scanning \cite{babu2016contrast,li2011level,shi2010neonatal,joseph2014brain}, improvement of irregularity detection in biometric fingerprint \cite{fingerprint_el2011design}, landscape analysis of remotely sensed satellite images \cite{GIS_based_Img_seg_devereux2004efficient}, also for object detection in still and moving images \cite{object_detection_yamaoka2006image} .


The approach of image segmentation can be broadly categorized into  discontinuity-detection and similarity-detection based methods \cite{kaur2014various}. The former is an approach of segmenting an image into regions based on discontinuity, whereas the later segments image into regions based on the similarity of pixels. Image segmentation can be achieved by a number of varying techniques, some of these are 1) thresholding, 2) clustering-based, 3) edge-based, 4) region-based, 5) watershed-based methods, 6) partial differential equation-based and 7) Artificial Neural Network (ANN)-based segmentation methods.

One of the simplest image segmentation technique is \textit{thresholding}. In this method, a threshold value is chosen to segment an image. All pixel values above or below the threshold value are classified as object or as a background. When only a single threshold value is used to segment image, it is known as \textit{global thresholding}, and when multiple threshold values are used to segment one or more objects, it is referred to as \textit{local thresholding} techniques.
\textit{Clustering} in image segmentation is a technique of thresholding, in which an image is partitioned into $K$-clusters. For each $K$-clusters, a cluster center is chosen randomly (or using a heuristic method). A pixel is assigned to a particular cluster based on the minimum distance between the pixel and the cluster center. The distance metric is usually based on features such as pixel color, intensity, texture, etc. These processes are iterated to compute appropriate cluster centers until convergence is achieved. Segmentation of image is achieved by mapping these clusters back to the original spatial domain \cite{Survey_img_seg_fu1981survey}.
However, \textit{edge-based} image segmentation is based on the theory that segmentation can be achieved by detecting discontinuity of pixels lying on the boundary between different regions. Gray histogram and gradient based method are two main edge-based segmentation methods \cite{kang2009comparative}.
The \textit{region-based} segmentation approach is based on the partitioning of the image into different regions according to a set of predefined criteria \cite{Region_based_shi2000normalized}. 
Another segmentation method, \textit{watershed-based} image segmentation replicates the process of rainfall in a real landscape. In a gray-scale landscape, light and dark intensity pixel are considered as hills and hollows of a gray-scale image. When an imaginary rainfall occurs in a gray-scale landscape, the rain flows from high altitude (gray level area) to some low lying (gray level) region. This flow creates watersheds or catchment basins. A gray-scale landscape is then segmented or partitioned into regions according to watersheds \cite{bleau2000watershed}.
When looked into a supervised segmentation along with a training data set a little or incomplete knowledge of the problem is required. Artificial Neural Networks (ANN) is a technique of supervised image segmentation. ANN are networks of interconnected parallel processing units. ANN partitions the image into multiple segments where all pixels in a partition holds some similar characteristics. As any other supervised learning model, ANN can learn by examples \cite{ANN_indira2011image}.
Extracting the desired object of interest from an image has always been the fundamental and most important task in image segmentation. When considering partial differential equations (PDE) for image segmentation, PDE always considers images as continuous objects. Due to the flexible structure, PDE converts images into initial and boundary conditions and later obtains the segmentation result as the solution of the equation \cite{PDE_wei2016image}.

A thresholding-based method is considered to be the simplest among all the known techniques. The problem is to determine a threshold value (for global thresholding) that divides the pixels into different classes. Two famous classic works are attributed to Otsu \cite{otsu1979threshold} and Kapur et al. \cite{kapur1985new}. The core idea behind Otsu's method is to maximize the between-class variance of gray levels. Kapur et al. propose the maximization of histogram entropy of segmented classes to select the optimal threshold value. Both these methods (Otsu's and Kapur's) can be easily extended for multilevel thresholding. However, they are inefficient in determining optimal thresholds due to the exponential growth in the computational complexity of the algorithm. The precision of the algorithm also decreases as the number of thresholds increases \cite{sathya2011optimal}

An approach that is in line with Kapur's work is \textit{minimization of cross entropy} commonly referred to as MCET. The work was initially presented by Solomon Kullback in \cite{kullback1997information}. MCET was considered as an extension to Kapur's work.  However, due to the computational complexity of determining the optimal threshold, the problem remains.
To address this problem, researchers propose a large number of meta-heuristic optimization algorithms. Some of these optimization algorithms minimizes the cross-entropy \cite{yin2007multilevel,horng2010multilevel,horng2011multilevel,oliva2014multilevel}, while others maximize the Kapur's entropy \cite{sathya2011optimal,bhandari2014cuckoo,khairuzzaman2017multilevel,bhandari2015modified} (or maximize Otsu's between class variance \cite{sathya2011optimal,khairuzzaman2017multilevel, bhandari2015modified}) to determine optimal threshold values. Segmentation of image is one of the most essential and preliminary steps in many applications related to computer vision and image processing. These meta-heuristic optimization algorithms converge to the optimal solution faster than the exhaustive search. However, when the dimension of the problem (i.e., the number of thresholds to be found out) increases, there is a proportional increase in the search time. The problem becomes worst for an image having higher dimensions (image size).

Image segmentation based on the histogram of an image is a popular thresholding technique. A histogram of an image consists of a number of peaks and valleys, and each valley separates a region or an object from its background. When there are only two distinct peaks in a histogram, it forms a bi-modal histogram. A valley between the two peaks forms an optimal global threshold value. However, when more than two peaks exist, global thresholding may not serve well. It requires more than one thresholds or a multilevel thresholding technique is a need.
In this paper, we propose a heuristic method of image segmentation using the multi-threshold technique by sampling the histogram of a digital image. Our algorithm is designed for a gray-scale image of $n$-levels. The algorithm consists of three main steps. First, it iterates over the $n$-levels and determines all valleys from the histogram so as to emphasis the resultant threshold as valley \cite{ng2006automatic}. Second, the histogram is equally partitioned into $r$-regions and determine points having minimum value (\textbf{Frequency} in a histogram see Figure \ref{fig:lenaHistogram}) within each region. The goal of this step is to select the minimum point within a region. The only two possibilities for this point is the lowest valley or a descending slope in the region. The advantage of this step is two-fold 1) it helps to eliminate a local minima problem within a region, which is a serious issue in most optimization algorithm and 2) it helps to select threshold point in a uniformly distributed fashion. Finally, the third step is to choose these optimal threshold values obtained in the previous two steps. Candidate points are formed by choosing common points in the two prior steps (valley points from the first step and minimum points from the second step).  We adopt an ad-hoc approach of clustering the candidate points and select a mean of the cluster as an optimal value. The number of clusters is the number of threshold values to be determined. To emphasis valley as the threshold, we select the immediate next candidate point to the mean of the cluster. The implementation of this approach and benchmarks reported in the paper can be downloaded from \url{https://sites.google.com/view/imagesegmentation/downloads}.

The rest of the paper is organized as follows. Section \ref{sec:prelims} provides the necessary basic concepts in image processing. In Section \ref{sec:methodology}, we
present our algorithm of multilevel thresholding. In Section \ref{sec:results}, we provide the experimental results to illustrate the performance of our approach compared to the most popular multilevel thresholding method. We conclude in Section \ref{sec:conclusion}.
\section{Preliminaries}\label{sec:prelims}
We propose an algorithm for image segmentation using the multi-thresholding technique for digital images. Our approach is mainly based on the histogram generated from the gray-scale image of the given image.

\begin{figure*}[!htb] 
	\centering{}
	\subfloat[Binary Image \label{fig:binaryImage}]
	{\includegraphics[scale=0.52]{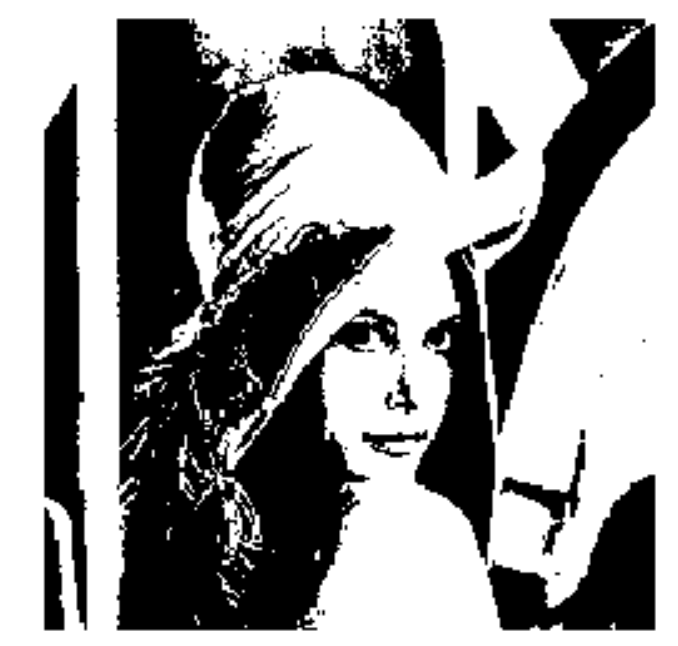}}
	\quad
	\subfloat[gray-scale Image \label{fig:grayImage}]
	{\includegraphics[scale=0.52]{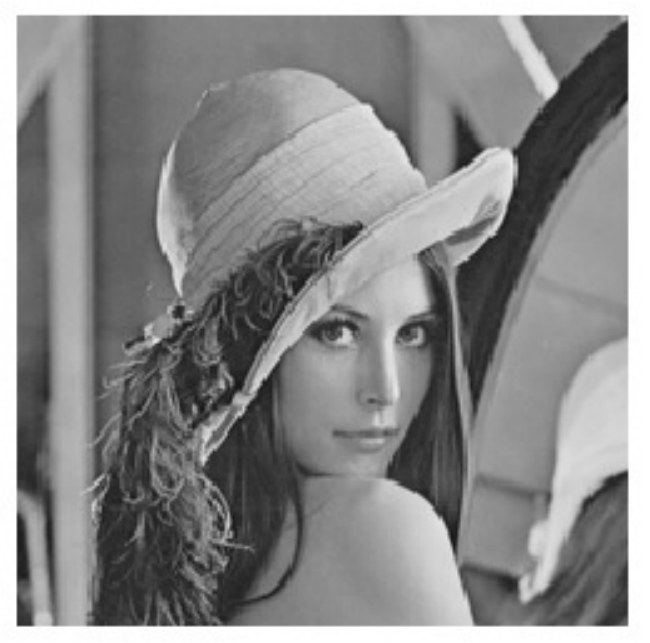}} 
	\quad
	\subfloat[RGB Image \label{fig:rgbImage}]
	{\includegraphics[scale=0.52]{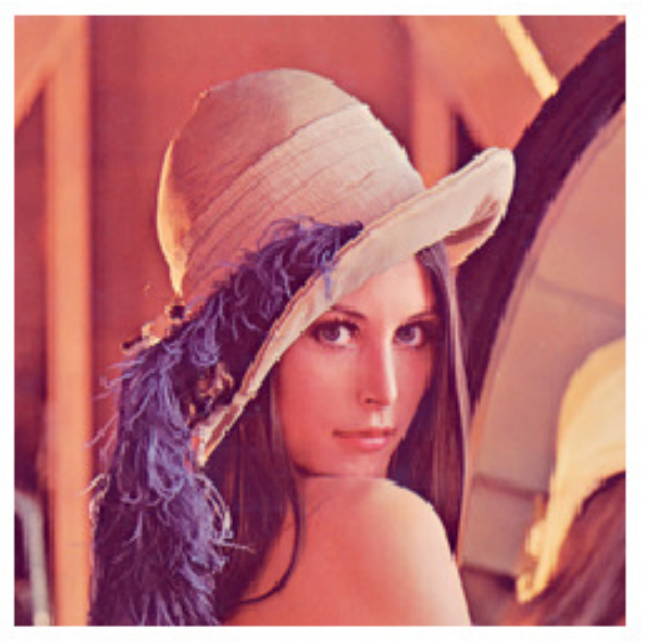}}
	\caption{Types of Digital Images for the benchmark image of a girl Lena. \label{fig:DigitalImages}}
\end{figure*}

\subsection{Digital Image} \label{subsec:DigitalImage}

Digital images are two-dimensional (2D) images defined as some function f(x,y), where x and y are known as spatial or plane coordinates. Digital images are transformed images from analog media to electronic data which can be saved, organized, retrieved and restored through electronic devices \cite{pornimaT}. They can be broadly classified into three different types on the basis of their size and range of pixel values as: 
\begin{enumerate}
	\item \textbf{Binary Image:} Digital images with only two possible values for every pixel is known as binary images. Each pixel will be stored as a single bit i.e., 0 or 1. Figure \ref{fig:binaryImage} shows the binary image generated from the original color image of the girl Lena, the most commonly used image benchmark in the field of image processing.
	\item \textbf{Gray-scale Image:} Gray-scale images are 8-bit images giving a possible range of pixel values from $L \in [0,255]$. The pixel values in a gray-scale image represent the brightness of the pixel. Typically zero is taken to be black, and 255 is taken to be white. The gray-scale image obtained from the color image of the girl Lena is shown in Figure \ref{fig:grayImage}.
	\item \textbf{Color Image:} It is also known as an RGB image, where R, G, and B stands for the primary color red, green, and blue, respectively. The RGB image is a system for representing the color to be used in a computer display as a two-dimensional array of small integers. Each of these integers represents a pixel value for an image. An RGB image has three-pixel values, one for each of red, green, and blue colors. The RGB image of the benchmark image Lena is shown in Figure \ref{fig:rgbImage}.
\end{enumerate}

Color images are converted to equivalent gray-scale using the standard formula \cite{poynton2012digital}
\begin{equation}\label{eq:color2gray}
I_{gray}(i,j)=[0.2989 0.5870 0.1140] \times \begin{bmatrix}
R(i,j)\\ 
G(i,j)\\ 
B(i,j)
\end{bmatrix}
\end{equation}
where $R(i,j)$, $G(i,j)$ and $B(i,j)$ are respectively red, green and blue pixel values of the color image. $I_{gray}(i,j)$ is the equivalent gray-scale value computed as a weighted sum of these three components.
These gray-scale images, in turn, can be easily converted to a binary image by applying a global thresholding technique.
\begin{equation}
I_{bw}(i,j) =\left\{\begin{matrix}
1 \qquad if \; \; I_{gray}(i,j)< th \\
\; \; 0 \qquad if \; \; I_{gray}(i,j)>= th
\end{matrix}\right.
\end{equation}
$th$ is the chosen global threshold value and $I_{bw}(i,j)$ is the corresponding binary value generated for the selected threshold for the image.

\subsection{Image Histogram} \label{sec:ImageHistogram}
For visualization of a target object from the background image, the histogram-based thresholding technique is the most commonly used approach for image segmentation, in digital image processing \cite{ismail2009simple}. A histogram is a graph consisting of x- and y-axis, where the x-axis is the gray level pixel values, and the y-axis gives the number of pixels (or frequency) corresponding to the gray levels. 
\begin{figure}[!htb]
	\centering
	\includegraphics[scale=0.45]{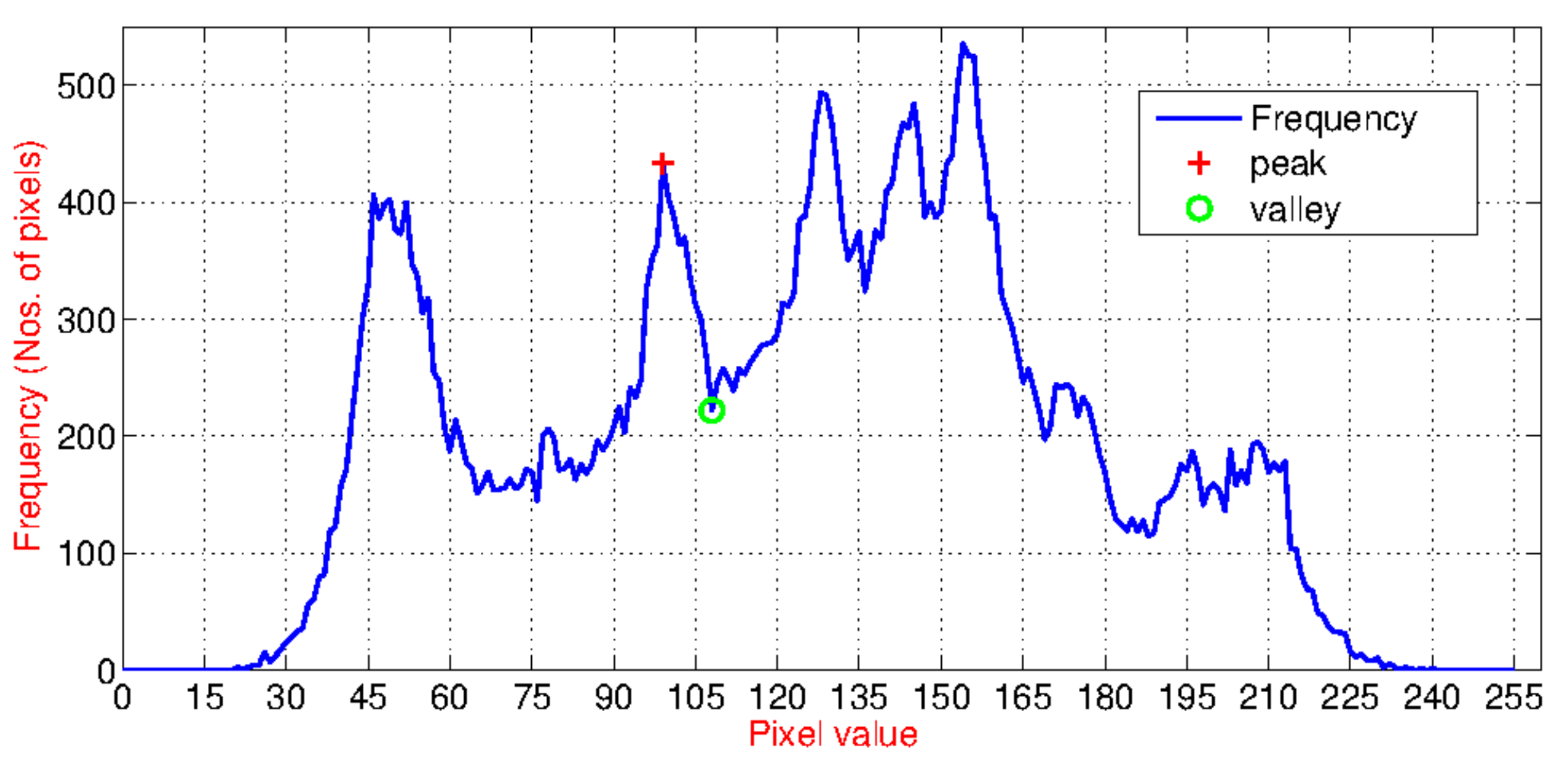}
	\caption{Histogram of a gray-scale image of Lena. The symbol {\color{red}\textbf{+}} (coloured in red) and {\color{green}\textbf{o}} (coloured in green) shows a \textit{peak} and a \textit{valley} point in a curve.}
	\label{fig:lenaHistogram}
\end{figure}
Figure \ref{fig:lenaHistogram} shows a histogram plot of the gray-scale image of Lena. The histogram plot is a nonlinear curve. In this paper, we call a \textit{peak} to a point representing the highest frequency in a curve and a \textit{valley} to the point denoting the least frequency in a curve (see Figure \ref{fig:lenaHistogram}).

\subsection{Multilevel thresholding}
Determining the best threshold for any digital image is computationally expensive. The three most popular method that can be found in the literature for global thresholding which subsequently extended for multilevel or local thresholding is presented in the following subsections.  

\subsubsection{Maximizing variance between classes}
Otsu proposes to maximize the between class variance in order to determine the best threshold value for image segmentation \cite{otsu1979threshold} (object from the background). Let $h$ represent the histogram of the image in gray-scale such that $h(i)=n_i/N$ and $N=n_1+n_2+ ... + n_L$ where $n_i$ is the number of pixels in level $i$.
Then, the probabilities of class occurrence $[1,...,(th-1)]$ and $[th,...,L]$ are given by
\begin{equation}\label{eq:omegas}
\omega_0(th) = \sum_{i=1}^{th-1} h(i) \qquad and \qquad \omega_1(th) = \sum_{i=th}^{L} h(i)
\end{equation}
Whereas the class mean is represented by
\begin{equation}\label{eq:mus}
\mu_0(th) = \sum_{i=1}^{th-1} i h(i)/\omega_0 \qquad and \qquad \mu_1(th) = \sum_{i=th}^{L} i h(i)/\omega_1
\end{equation}
Therefore, to determine the best threshold value $th$ Otsu proposes to maximize the between-class variance denoted by 
\begin{equation}
\sigma_{B}^{2}(th) =\omega_0 \omega_1 (\mu_1 - \mu_0)^2
\end{equation}
This method can be easily extended to support multilevel thresholding. For instance, when the number of thresholds to be determine is two ($th_1, th_2$), three classes or probability distributions are formulated as $X_1=[1,...,(th_1 - 1)]$, $X_2=[th_1,...,(th_2 - 1)]$ and  $X_3=[th_2,...,L]$. Accordingly, the optimal thresholds is now a function of two variables $th_1, th_2$ which is computed as 
\begin{equation}
\arg \max \{\sigma_{B}^{2}(th_1,th_2)\}
\end{equation}

\subsubsection{Maximizing entropy}
Kapur et al. presented an algorithm based on the concept of entropy to segment digital image. Let $h(i)$ bears the same meaning as in Equation \ref{eq:omegas}. To determine two thresholds (say $t=[th_1,th_2]$), where $1< th_1 < th_2 < L$, the probabilities of class occurrence $[1,...,(th_1 - 1)]$, $[th_1,...,(th_2 - 1)]$ and $[th_2,...,L]$ are given by
\begin{equation}\label{eq:omegas2}
\omega_0(t) = \sum_{i=1}^{th_1-1} h(i) ; \quad \omega_1(t) = \sum_{i=th_1}^{th_2 - 1} h(i) \quad and \quad
\omega_2(t) = \sum_{i=th_2}^{L} h(i)
\end{equation}

Then, the entropies for each of these classes are given by
\begin{equation} \label{eq:entropy1}
\begin{split}
H_0(t) = - \sum_{i=1}^{th-1} \frac{h(i)}{\omega_0} \ln \left( \frac{h(i)}{\omega_0} \right) \\
H_1(t) = - \sum_{i=th_1}^{th_2 - 1} \frac{h(i)}{\omega_1} \ln \left( \frac{h(i)}{\omega_1} \right) \\
H_2(t) = - \sum_{i=th_2}^{L} \frac{h(i)}{\omega_2} \ln \left( \frac{h(i)}{\omega_2} \right)
\end{split}
\end{equation}
The optimal thresholds are computed by maximizing the sum of the entropies. For global threshold, the optimal threshold is given by $\sigma_{W}(th_1) = H_0(t) + H_1(t)$ here $th_2=L$. Multilevel thresholding, $t=[th_1,th_2]$ optimal thresholds are obtained by maximizing  $\sigma_{W}(th_1,th_2) = H_0(t) + H_1(t) + H_2(t)$.

\subsubsection{Maximizing cross entropy}
The cross-entropy between two probabilistic distribution is the measure of the statistical difference in uncertainty in the outcome of the experiment when data is transmitted from one distribution to another. Kullback's  cross-entropy is given as \cite{kullback1997information}:
\begin{equation}
\varphi (X,Y) = \sum_{i=1}^{N}x_i \log\left ( \frac{x_i}{y_i}  \right )
\end{equation}
where $X=\{x_1,x_2,...,x_N\}$ and $Y=\{y_1,y_2,...,y_N\}$ are the two probability distribution. A high value of $\varphi$ represents more uncertainty in the distribution process.

In a digital image, to segment an image into an object and a background (i.e., into two partitions), a threshold value $th$ is chosen. For efficient image segmentation, an optimal threshold is computed by minimizing the cross-entropy given by \cite{li1993minimum}:
\begin{equation} \label{eq:mcet1}
\eta (th) = \sum_{i=1}^{th-1} ih(i) \log \left( \frac{\sum_{i=1}^{th-1} ih(i)}{\sum_{i=1}^{th-1} h(i)} \right) + \sum_{i=th}^{L} ih(i) \log \left( \frac{\sum_{i=th}^{L} ih(i)}{\sum_{i=th}^{L} h(i)} \right)
\end{equation}
where $h$ is the histogram of the image.
The computational complexity for determining a single threshold value is $O(L^2)$. However, this complexity increases to $O(L^{n+1})$ for `$n$' threshold values. 
To compute an optimal threshold using exhaustive search, we compute Equation \ref{eq:mcet1} \textbf{for all $th \in [1,L]$}, and select the $th$ where $\eta (th)$ is the minimum of all. This is computationally an expensive operation, to reduce this complexity, \cite{tang2011improved} presented an improvement by introducing recursive programming to support multilevel thresholding. Let $[th_1,th_2,...,th_n]$ be the set of thresholds to be determined, $th_0 < th_1 < th_2,...,th_n < th_{n+1}$, where $th_0=1$ and $th_{n+1}=L+1$ are dummy thresholds introduced for convenience. The objective function to be minimize, to obtain an optimal thresholds can be represented as:
\begin{equation}\label{eq:recursiveMCET}
\eta (th_1,th_2,...,th_n) = \sum_{i=1}^{n+1}m^1(th_{i-1},th_i) \log \left( \frac{m^1(th_{i-1},th_i)}{m^0(th_{i-1},th_i)}\right)
\end{equation}
where $m^0$ and $m^1$ are the values of zero-moment and first-moment points computed as 
\begin{equation*}
m^0(a,b) = \sum_{i=a}^{b-1} h(i)  \qquad  and \qquad  m^1(a,b) = \sum_{i=a}^{b-1} ih(i) 
\end{equation*}

Equation \ref{eq:recursiveMCET} reduces the computational complexity from $O(L^{n+1})$ to $O(L^{n})$, `$n$' being the number of thresholds to be determined. However, this reduction did not do any better when `$n$' is large. The literature presents a large number of meta-heuristic optimization algorithms (mentioned earlier in Section \ref{sec:introduction}). This algorithm applies heuristic techniques on these three popular thresholding methods to converge to the optimal solution in fewer iterations. However, there is no clear winner in this race. Segmentation is the most important step in all image processing and computer vision. Therefore, determining an efficient segmentation technique is still a recent research area in digital image processing. 

\section{Multilevel Thresholding using Histogram Sampling}\label{sec:methodology}
We propose an approach of determining multiple threshold values from a given image represented as a gray-scale image. When the input image is a color image, it is converted into gray-scale using Equation \ref{eq:color2gray}. 

\subsection{Proposed Algorithm}
We present below our proposed algorithm as five major steps:

\textbf{Step-1: } We obtain the normalized histogram $h$ of the input image represented as  
$h(i)=n_i/N$ for $N=n_1+n_2+ ... + n_L$,  where $n_i$, is the number of pixels in level $i$. $L$ is the pixel of an image representing the highest gray level intensity.

\textbf{Step-2: } Let $X_v$ be the set of all valley points between [1, $L$]. A valley point is a pair $(i, h(i))$. We scan the histogram of the image and obtain all pixel-level that represents a \textit{valley} in the histogram and construct \textbf{setA} as\\
\tab \lstinline{for} $i$ in 1 to $L$ \\ 
\tab \tab \codeword{setA} = \{$i$ : $h(i) \in X_v$\} \\ Note that as mentioned in Section \ref{sec:ImageHistogram}, a \textit{valley} is a point representing the lowest frequency of a pixel-level in a curve. A simple method to determine all valleys in the histogram is a gradient search method. In this case, gradient descent is used to find all discrete local minima.

\textbf{Step-3: } We sample the histogram into $r$-regions or partitions of equal size, and determine pixel-level having the lowest frequency in each of these partitions. We decide the number of partitions as \\
\tab \codeword{r} $= L/s$ for $s \in $ \{ $x$ : ($L$ \% $x$) is zero, and $x>1$ \}\\
where \% is the modulo division operator. 
We computed \textbf{setB}, to obtain the set of all minimum points in the histogram $h$, for each $r$-partitions as follows:\\
\tab \codeword{setB} = $\arg \underset{r_{i-1} \le i < r_i}{min} \{h(i)\}$ ; $r_i$ is the $i^{th}$ partition. \\
When $L=256$, the possible values for $r$ are \textbf{2/4/8/16} and the partition sizes can be \textbf{128/64/32/16}. 
In this paper, we chose 32 equal partitions. However, when the required number of thresholds is 32 or more, 64 or higher partitions size can be selected.
The goal of this step is to select the minimum point within a region. The only two possibilities for this point is either the lowest valley or the last spot in a descending slope in that region. The advantage of this step is; first, it helps to eliminate a local minima problem within a region. A local minima problem is a severe concern in most optimization algorithms. Secondly, this partitioning of histograms helps to uniformly distribute the candidate thresholds in the histogram of a digital image.

\textbf{Step-4: } We now create new set \textbf{setC} using sets \textbf{setA} obtained in \textbf{Step-2} and \textbf{setB} in \textbf{Step-3}.
\textbf{setC} contains the elements that are common in both $setA$ and $setB$, computed as \\
\tab \codeword{setC} = $setA \cap setB$

\textbf{Step-5: } Thus, \textbf{setC} contains the candidate threshold values. 
Now, based on the number of threshold values to be determined, we can select appropriate thresholds from the candidate set \textbf{setC}. Let $t$ be the number of thresholds to be determined.
We adopt a very naive approach of grouping the candidate points into $t$ clusters and select the mean of the cluster as optimal value. The number of clusters formed is based on the number of threshold values to be determined. We emphasis thresholds as valley points and select the immediate next candidate point to the mean of the cluster (for the same reason as mentioned earlier).

For computational efficiency, we perform \textbf{Step-2} and \textbf{Step-3} under the same scan of the histogram $h$. Moreover, the histogram of an image is already a probability distribution function (PDF), and our approach do not depend on the normalized histogram. Therefore, we may skip the computation involved in \textbf{step-1}, instead just use the histogram obtained from the input image.

\subsection{Complexity Analysis}
The computational time of our algorithm for computing multilevel thresholds are constant. However, for most algorithms, the computational time increases with an increase in the number of thresholds to be determined. Step-1 is the most expensive step in our algorithm, which computes the histogram of an image. In the worst case, the time to compute a histogram is $O(N^2)$, assuming the height and width of the image are equal to $N$. The next two steps are computed in a single \lstinline{for} loop of size $L$, this can be done in $O(L)$, where $L$ is the highest intensity level of the pixel. Step-4 compares the elements of sets \textbf{setA} and \textbf{setB}, this requires $O(max(a,b))$, where $a$ and $b$ are the number of elements in the two sets, $a,b < L$. Finally, Step-5 requires at most $r$ iterations which compute the mean of candidate thresholds for each $t$ clusters. This computation can be done in $O(r)$, where $r$ and $t$  are the number of partitions and thresholds to be determined, respectively. The number of partitions $r$ is usually constant, and  $t < r < L$; therefore, an increase in the size of $t$ does not affect the computational time.

\subsection{Illustration}
\begin{figure}[!htbp]
	\centering
	\includegraphics[width=1.0\linewidth]{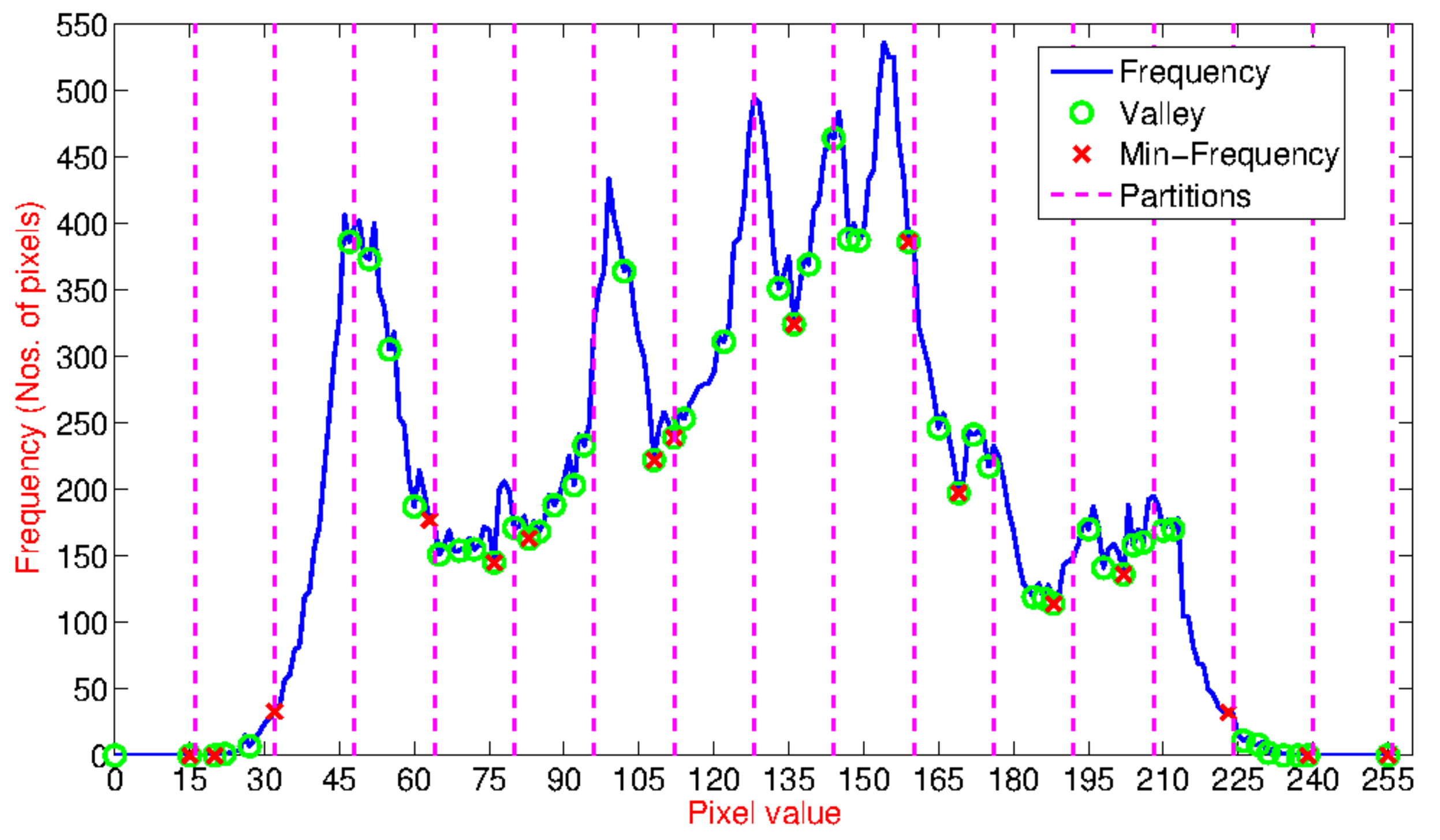}
	\caption{Illustration of our approach on the image of Lena.}
	\label{fig:HistogramAlgorithm}
\end{figure}

We illustrate our approach with the help of Figure \ref{fig:HistogramAlgorithm}. The algorithm begins by generating a histogram of the image in \textbf{Step-1}. \textbf{Step-2} generates all the \textit{valley} points in the histogram and create set \textbf{Set-A}. In the figure, all points represented by the {\color{green}green circle} are valley. In \textbf{Step-3}, the algorithm partitions the histogram into equal-partitions, in this example it is divided into 16 partitions. The {\color{magenta}dotted lines(- -) in magenta} colour denotes the partitions. In each of these partitions, the pixel having the least frequency is chosen to form a set \textbf{setB}. These points are marked as {\color{red}cross (x) in red} colour. \textbf{Step-4} determines only those points that are common in both \textbf{setA} and \textbf{setB} and call this set as \textbf{setC}. In the example, we obtain 13 such points when 16 partitions are chosen. Finally, \textbf{Step-5} returns the required number of threshold values from the set \textbf{setC} based on a very naive clustering approach.

\subsection{Segmentation using Multi-thresholding}
We segment the image into multiple segments by using threshold values obtained from the proposed algorithm. In this work, we use the following approach to segment image into three classes using two thresholds:
\begin{equation}
I_{seg}(i,j) =\left\{
\begin{matrix}
I_{gray}(i,j)  \qquad \qquad if \; \; I_{gray}(i,j) \le th_1 \\
\; \; th_1     \qquad \qquad if \; \; th_1 < I_{gray}(i,j) \le th_2 \\
I_{gray}(i,j) \qquad \qquad if \; \; I_{gray}(i,j) > th_2 \\
\end{matrix}\right.
\end{equation}
When the number of thresholds are more than two we use the following approach to generate segmented image:
\begin{equation}
I_{seg}(i,j) =\left\{
\begin{matrix}
I_{gray}(i,j) ,  \qquad  \qquad   \qquad  \qquad if \; \; I_{gray}(i,j) \le th_1 \\
th_1 ,  \quad if \; \; th_{i-1} < I_{gray}(i,j) \le th_i, \quad i=2,3,...,t-1 \\
I_{gray}(i,j) ,  \qquad  \qquad   \qquad  \qquad if \; \; I_{gray}(i,j) > th_t \\
\end{matrix}\right.
\end{equation}

\section{Experiments} \label{sec:results}

We have implemented our proposed algorithm in MATLAB.
In the text that follows, we refer to our histogram-based algorithm as AMTIS, abbreviating \textbf{A}utomatic \textbf{M}ultilevel \textbf{T}hresholding for \textbf{I}mage \textbf{S}egmentation. We present an evaluation of the algorithm on various standard benchmarks commonly used in the literature. Performance of our proposed algorithm in comparison to the popular Otsu's method is reported. We use MATLAB's built-in function \textit{multithresh}, which implements Otsu's method of multilevel thresholding \cite{otsu1979threshold}.

\begin{table}
\centering
	\begin{tabular}{|c|c|c|}
		\hline Sl. No. & Benchmark & Size (width $\times$ height)\\ 
		\hline 1 & Lena &  $220 \times 220$ \\ 
		\hline 2 & Cameraman & $256 \times 256$ \\ 
		\hline 3 & Hunter & $512 \times 512$  \\ 
		\hline 4 & Baboon & $512 \times 512$ \\ 
		\hline 5 & Fruits & $512 \times 512$  \\ 
		\hline 6 & Mountain & $640 \times 480$  \\ 
		\hline 7 & Airplane & $512 \times 512$   \\ 
		\hline 8 & Boat & $512 \times 512$  \\ 
		\hline 9 & FingerPrint\_1 & $300 \times 300$  \\ 
		\hline 10 & FingerPrint\_2 & $300 \times 300$  \\ 
		\hline 11 & Blonde (Lady Zelda) & $512 \times 512$  \\ 
		\hline 12 & Frozen Franz Joshef & $4531 \times 6005$ \\ 
		\hline 
	\end{tabular} 
	\caption{Image benchmarks and their sizes in pixels.} \label{tab:benchmarkSize}
\end{table}

\subsection{Benchmarks} We use standard image benchmarks that are popular in image processing. The dimensions of the benchmarks are shown in Table \ref{tab:benchmarkSize}. Some of these images are obtained from the USC-SIPI image database. The image \textit{Frozen Franz Josef} is taken from \href{https://earthobservatory.nasa.gov/images/76883/frozen-franz-josef-land}
{\url{https://earthobservatory.nasa.gov/images/76883/frozen-franz-josef-land}}. Franz Josef is located 600 miles from the North Pole. It is always covered with ice even during the summer. The image is a satellite image made from a combination of visible and near-infrared wavelengths. The two fingerprint images are taken from \href{http://bias.csr.unibo.it/fvc2000/download.asp}{\url{http://bias.csr.unibo.it/fvc2000/download.asp}}, using the \textit{DB1\_B.zip} and the benchmark files are \textit{101\_1.tif} and \textit{105\_2.tif} respectively.

\subsection{Results}
The experiments were performed on AMD FX(TM)-6100 Six-Core Processor, 3.3GHz, with 8 GB RAM. The results are an average of 20 runs. The number of thresholds evaluated is 2, 3, 4, and 5 in line with the results presented in the related literature \cite{oliva2019image,horng2011multilevelExperiment,maitra2008hybrid}. We obtain threshold values using our algorithm AMTIS and generate a segmented image. To verify the quality of the segmented image, we compute the peak-to-signal ratio (PSNR), the structure-similarity index (SSIM) and feature similarity index (FSIM).
The PSNR is a measure to determine the quality of the reconstructed image (in this case the segmented image $I_{seg}$) in comparison to the original image ($I_{gray}$) using the root mean square error (RMSE) as:
\begin{equation}
\begin{split}
PSNR = 20 log_{10}\left ( \frac{Max_p}{RMSE} \right ), (dB) \\
RMSE = \sqrt{\frac{\sum_{i=1}^{m}\sum_{j=1}^{n}(I_{gray}(i,j) - I_{seg}(i,j))}{m\times n}}
\end{split}
\end{equation}
where $Max_p$ is the highest intensity value of a pixel. The unit of measurement is in decibel (dB). When the intensity value is represented using 8-bit, $Max_p=255$. A higher PSNR value is desired for better quality \cite{welstead1999fractal}.

SSIM \cite{wang2004image} is used to measure the structural similarity between the original image and the segmented image computed using the Equation \ref{eq:ssim}. Like PSNR, for a better segmentation quality, a higher value of SSIM is desired.
\begin{equation} \label{eq:ssim}
\begin{split}
SSIM (I_{gray}, I_{seg}) = \frac{(2\mu_{I_{gray}}\mu_{I_{seg}} + C1)(2\sigma_{I_{gray}}\mu_{I_{seg}} + Cc)}{(\mu_{I_{gray}}^{2} + \mu_{I_{seg}}^{2} + C1)(\sigma_{I_{gray}}^{2} + \sigma_{I_{seg}}^{2} + C2)}, \\
\sigma_{I_{gray}I_{seg}} = \frac{1}{N+1}\sum_{i=1}^{N}({I_{gray}}_{i} - \mu_{I_{gray}})({I_{seg}}_{i} - \mu_{I_{seg}})
\end{split}
\end{equation}
where $\sigma_{I_{gray}I_{seg}}$ is the standard deviation, $C1$, $C2$ are constant values used to avoid instability when ($\mu_{I_{gray}}^{2} + \mu_{I_{seg}}^{2}$) approaches to zero.

FSIM \cite{zhang2011fsim,bhandari2014cuckoo} calculates the similarity between two images: in this case, the original gray-scale image and the segmented image. As PSNR and SSIM, the higher value is interpreted as a better performance of the thresholding method. The FSIM is then defined as:
\begin{equation}
FSIM =\frac{\sum_{x\in \Omega} S_L(x) PC_m(x)}{\sum_{x\in \Omega}PC_m(x)}
\end{equation}
where,
\begin{equation}
\begin{split}
S_L(x) = S_{PC}(x)S_{G}(x), \\
S_{PC}(x) = \frac{2PC_{1}(x)PC_{2}(x) + T_1}{{PC_{1}^2}(x)+ {PC_{2}^{2}}(x) + T_1}, \\
S_{G}(x) = \frac{2G_{1}(x)G_{2}(x) + T_2}{{G_{1}^2}(x)+ {G_{2}^{2}}(x) + T_2}
\end{split}
\end{equation}
The gradient magnitude of the image, $G$ is given by:
\begin{equation}
G=\sqrt{G_x^2 + G_y^2}
\end{equation}
and PC is the phase congruence, expressed as 
\begin{equation}
PC(x) = \frac{E(x)}{ \varepsilon + \sum n A_n(x)}
\end{equation}
The local amplitude on the scale of $n$ is $A_n(w)$ and $E(w)$ is taken to be the magnitude of the response vector in $w$ on $n$. The term $\varepsilon$ is a positive constant. For a better segmentation quality, a higher value of FSIM is desired.

\begin{table*}
	\centering
	\includegraphics[width=1.0\linewidth]{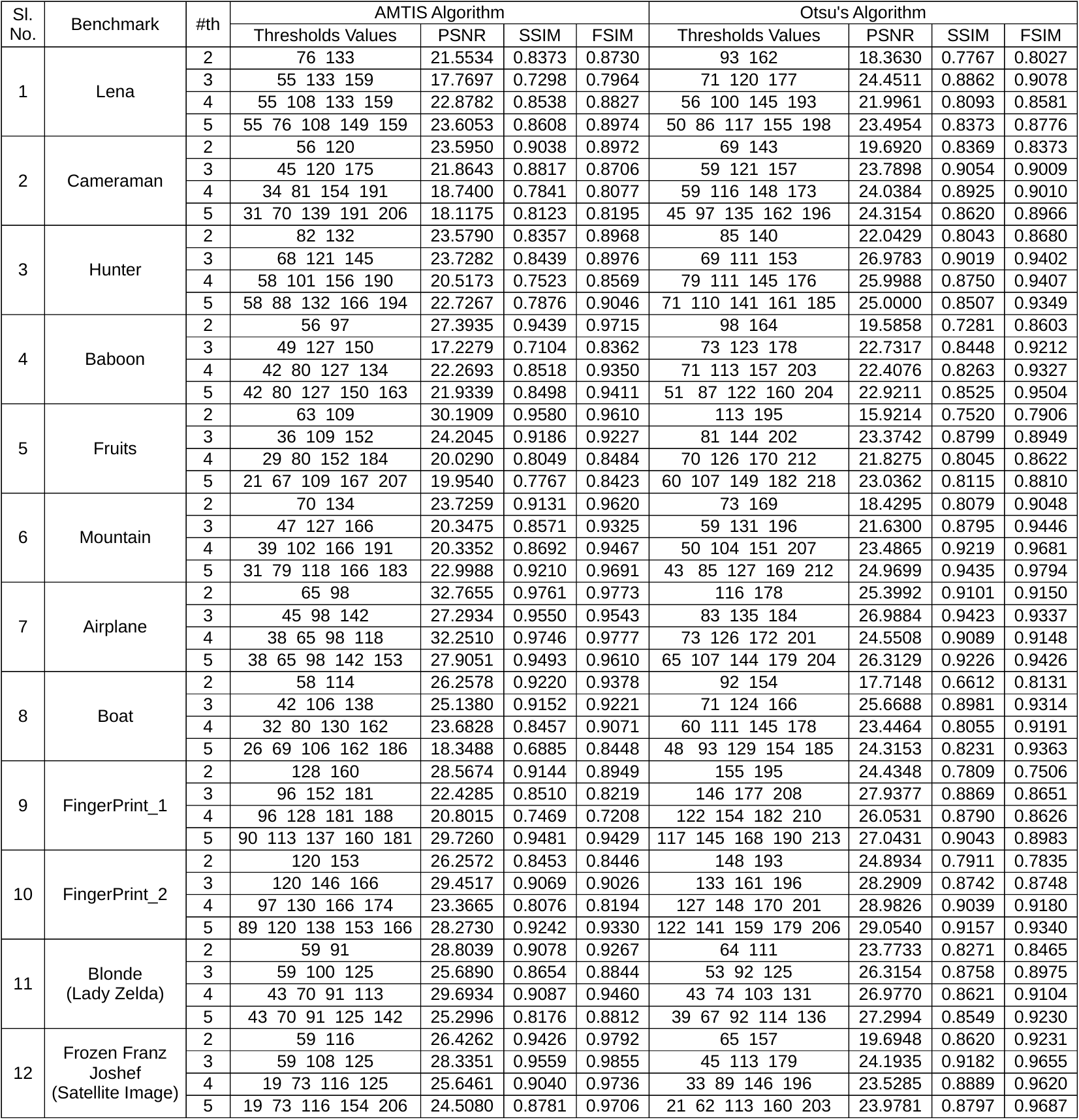}
	\caption{Comparison of results obtained by AMTIS to that of Otsu's Method.}
	\label{fig:incompleteData}
\end{table*}

Table \ref{fig:incompleteData} shows the performance comparison of our proposed algorithm (AMTIS) to that of Otsu's method. The algorithm is compared to the generated optimal threshold values and the values of PSNR, SSIM, and FSIM. We observed that the results obtained by our proposed algorithm AMTIS are comparable and better in many cases in comparison to the popular Otu's method.

\begin{table*}
	\centering
	\includegraphics[scale=0.9]{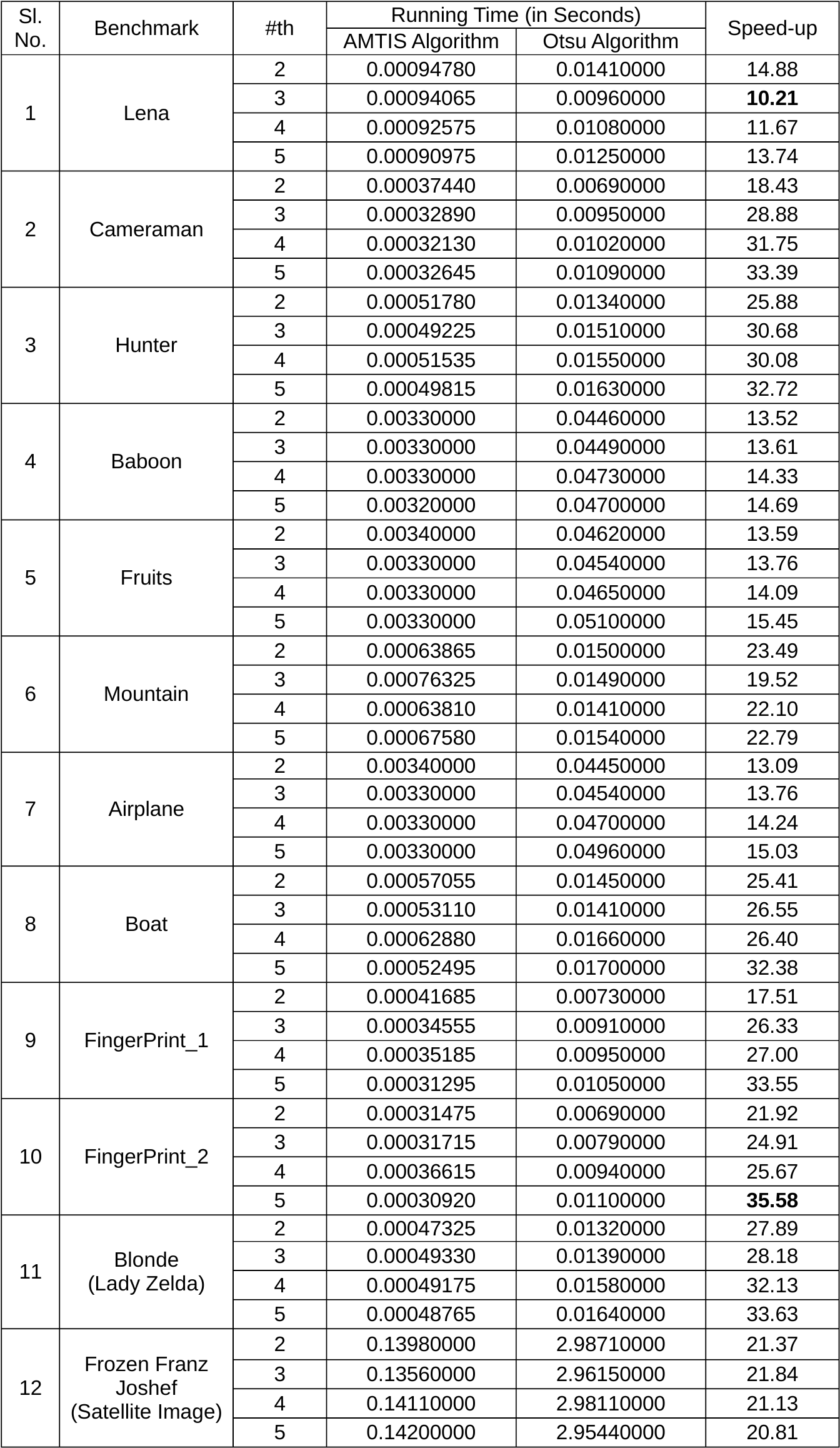}
	\caption{Performance speed-up on various benchmarks using AMTIS compared to Otsu's Method in MATLAB.}
	\label{fig:completeData}
\end{table*}

In a few benchmarks, AMTIS fails to compute optimal threshold values. This drawback is because we adopt a naive approach in selecting the thresholds, one way to improve this is by devising appropriate clustering technique. However, our algorithm outperforms Otsu's method in CPU computational time, as evident in Table \ref{fig:completeData}. A maximum speed-up of $35.58\times$ is observed for \textit{FingerPrint\_2} benchmark and a minimum speed-up of $10.21\times$ for the most popular benchmark, \textit{Lena} respectively. 


\begin{figure*}[p] 
	\centering{}
	\subfloat[threshold=2 \label{fig:Lenath2}]
	{\includegraphics[scale=0.45]{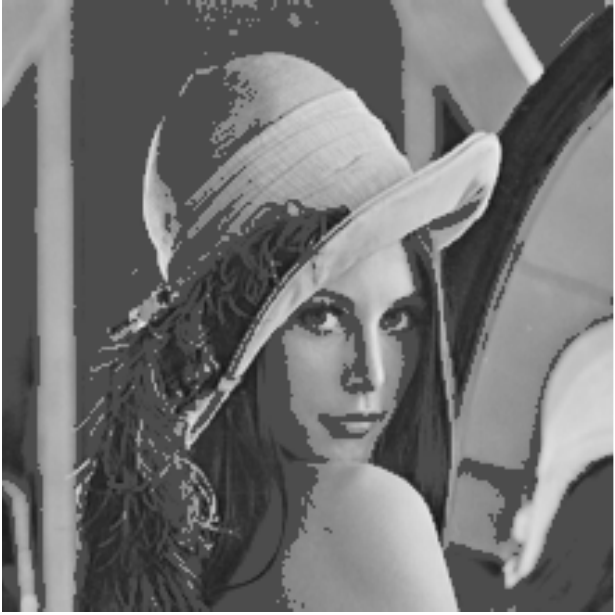}} \:
	\subfloat[threshold=3 \label{fig:Lenath3}]
	{\includegraphics[scale=0.45]{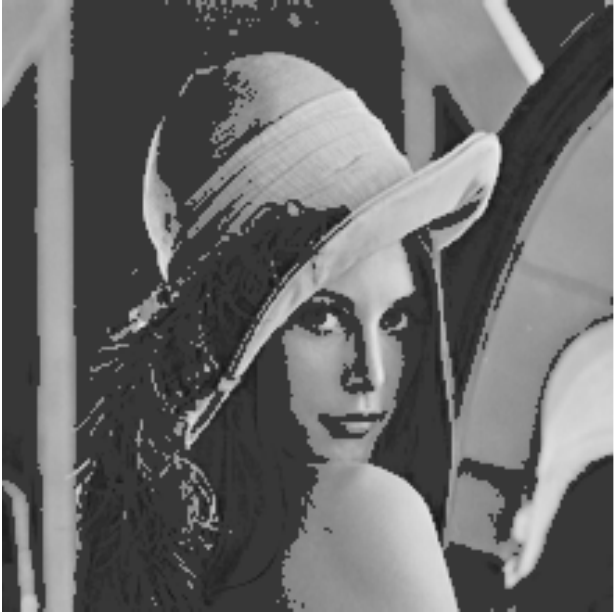}} \:
	\subfloat[threshold=4 \label{fig:Lenath4}]
	{\includegraphics[scale=0.45]{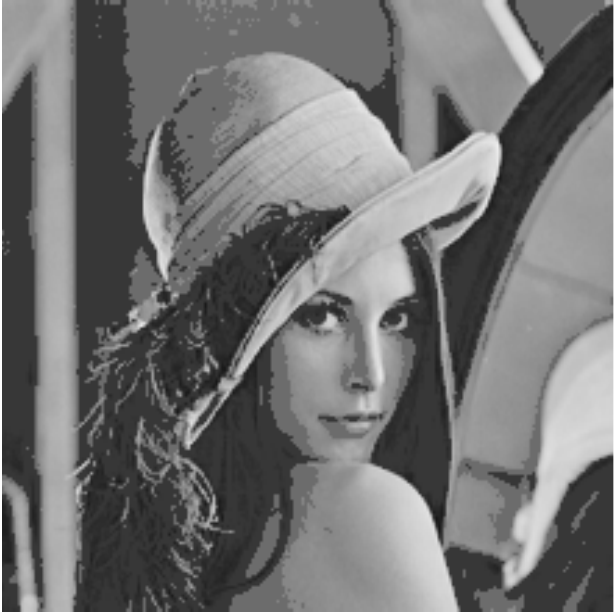}} \:
	\subfloat[threshold=5 \label{fig:Lenath5}]
	{\includegraphics[scale=0.452]{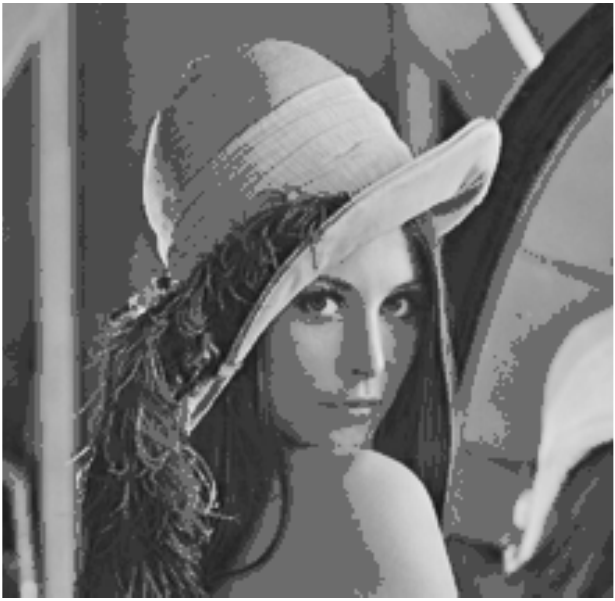}}
	
	\subfloat[threshold=2 \label{fig:Lenath2Histogram}]
	{\includegraphics[scale=0.2]{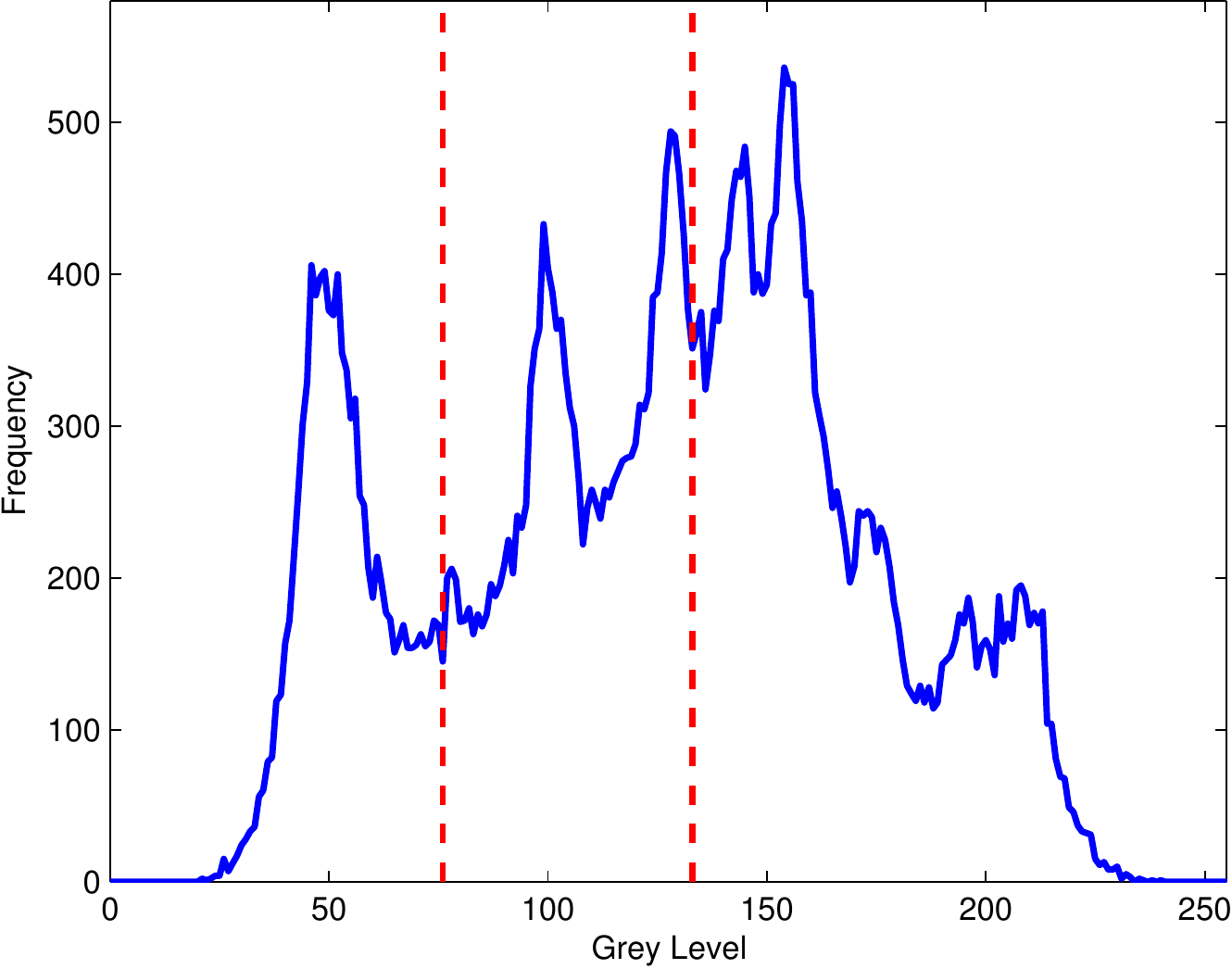}} \:
	\subfloat[threshold=3 \label{fig:Lenath3Histogram}]
	{\includegraphics[scale=0.22]{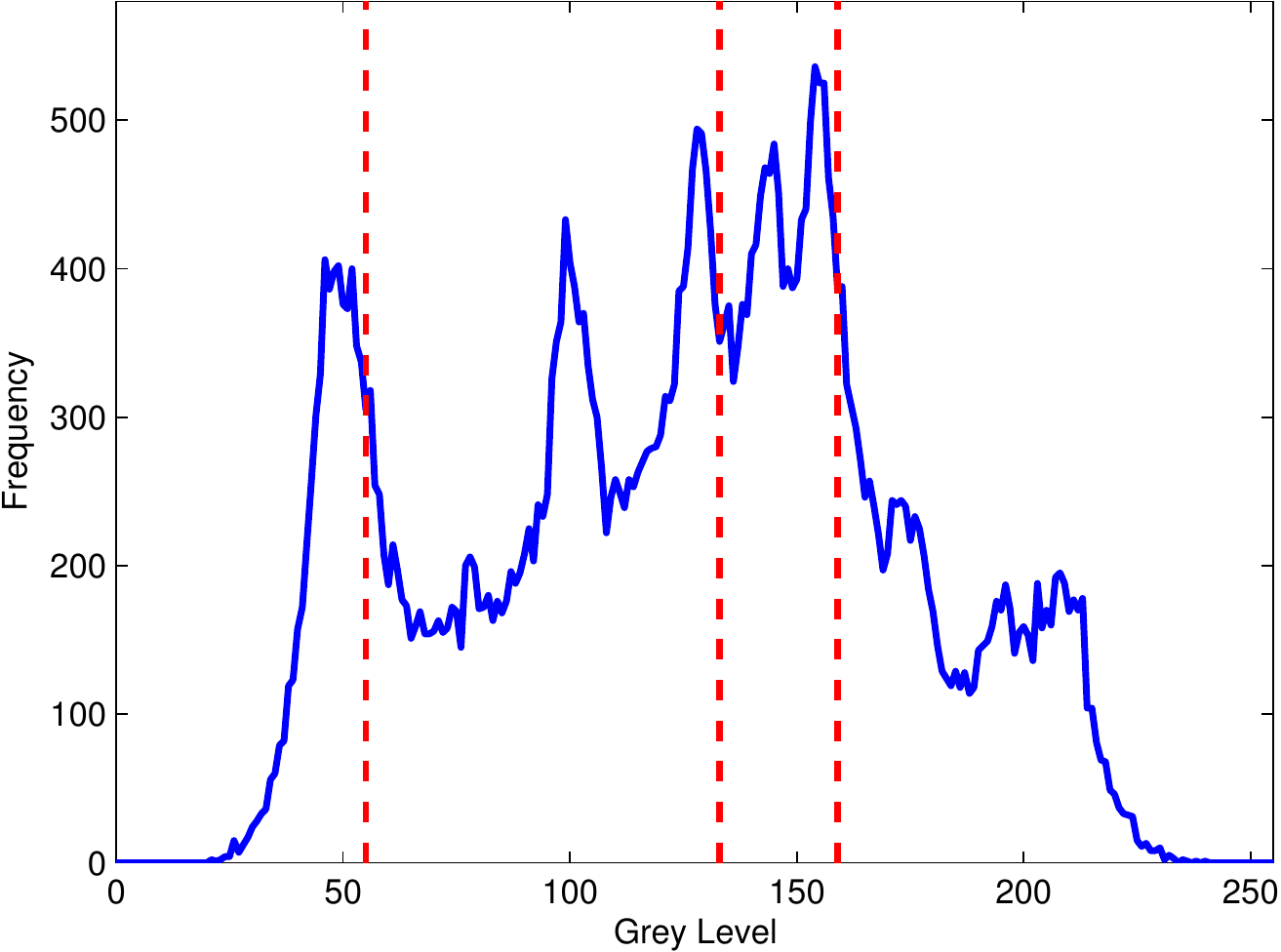}}  \:
	\subfloat[threshold=4 \label{fig:Lenath4Histogram}]
	{\includegraphics[scale=0.2]{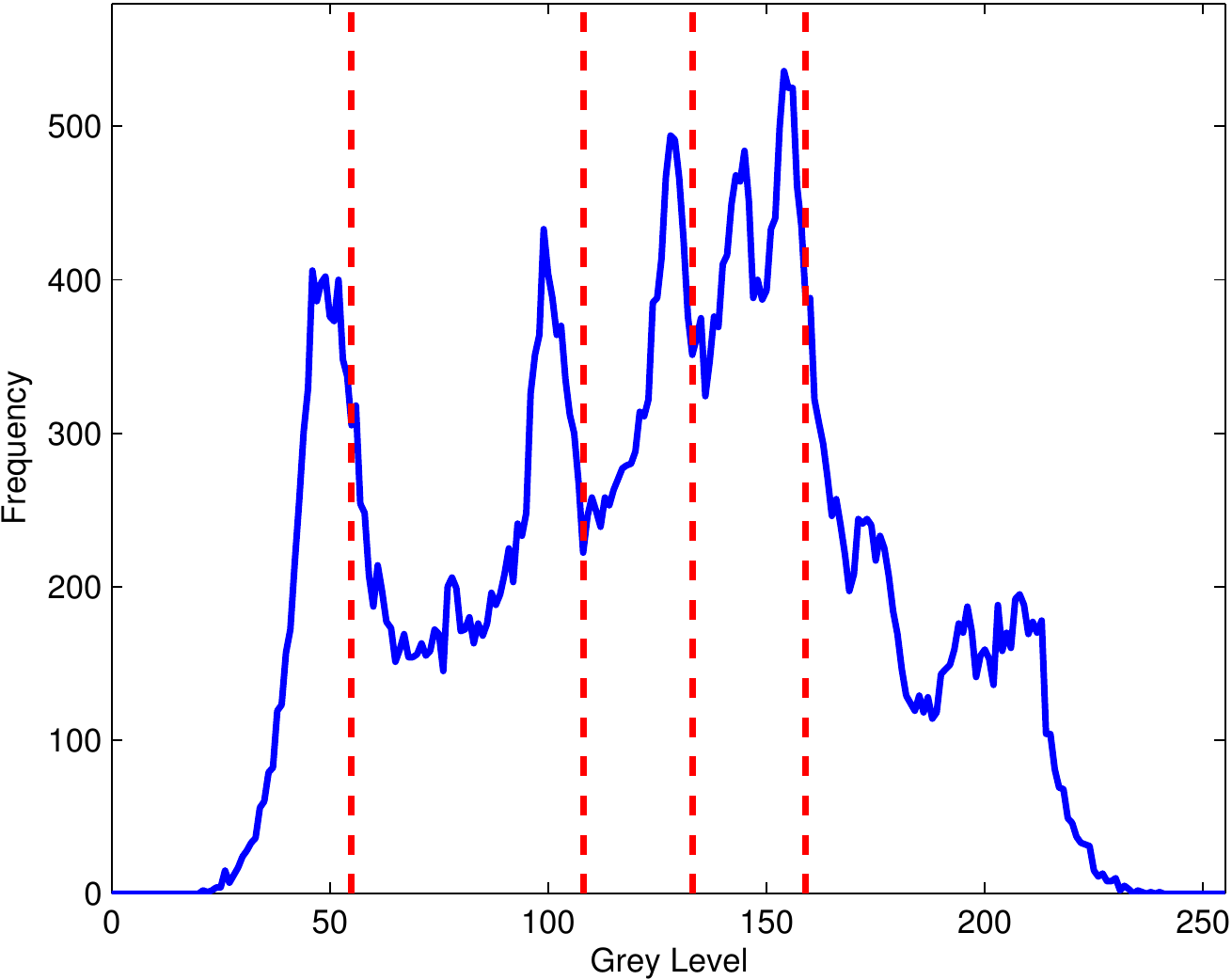}} \:
	\subfloat[threshold=5 \label{fig:Lenath5Histogram}]
	{\includegraphics[scale=0.2]{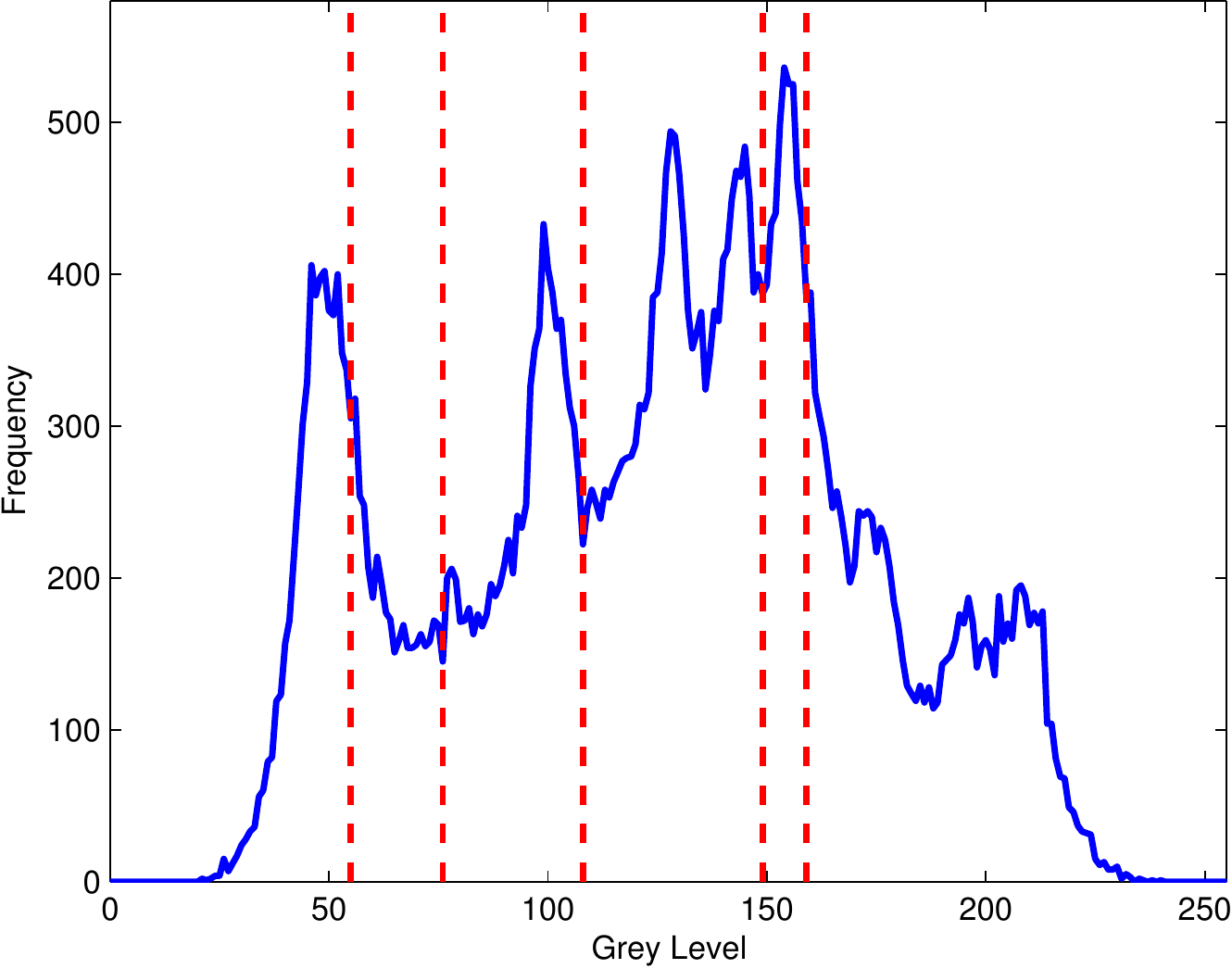}}
	
	\subfloat[threshold=2 \label{fig:Cameramanth2}]
	{\includegraphics[scale=0.37]{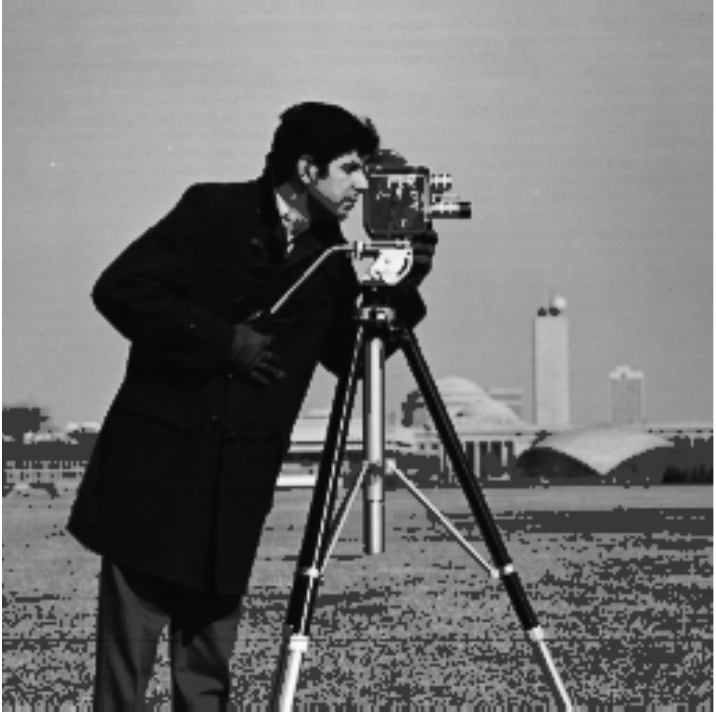}} \:
	\subfloat[threshold=3 \label{fig:Cameramanth3}]
	{\includegraphics[scale=0.37]{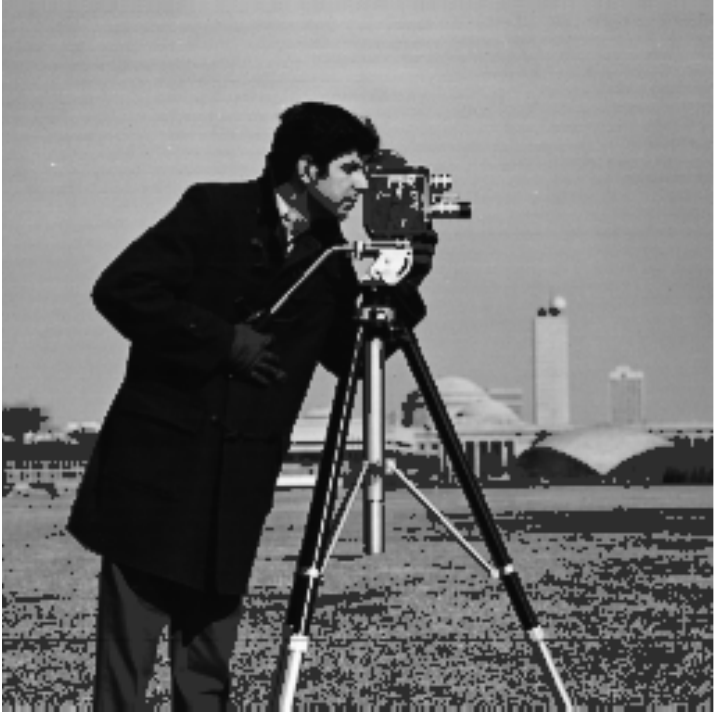}}  \:
	\subfloat[threshold=4 \label{fig:Cameramanth4}]
	{\includegraphics[scale=0.37]{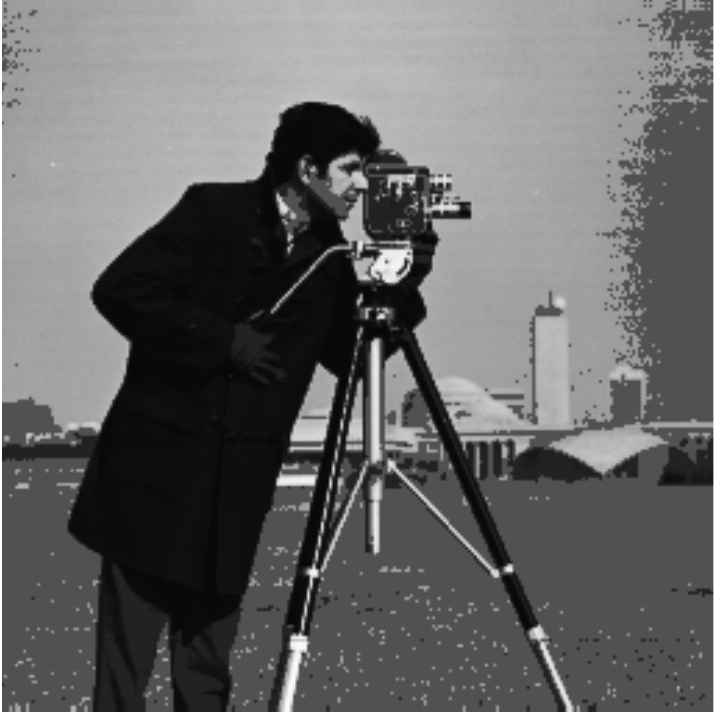}} \:
	\subfloat[threshold=5 \label{fig:Cameramanth5}]
	{\includegraphics[scale=0.37]{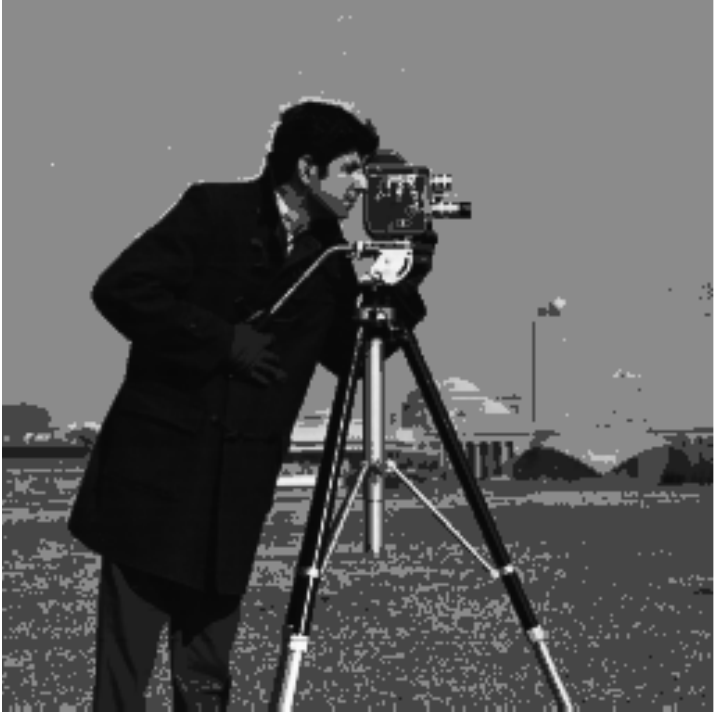}}
	
	\subfloat[threshold=2 \label{fig:Cameramanth2Histogram}]
	{\includegraphics[scale=0.2]{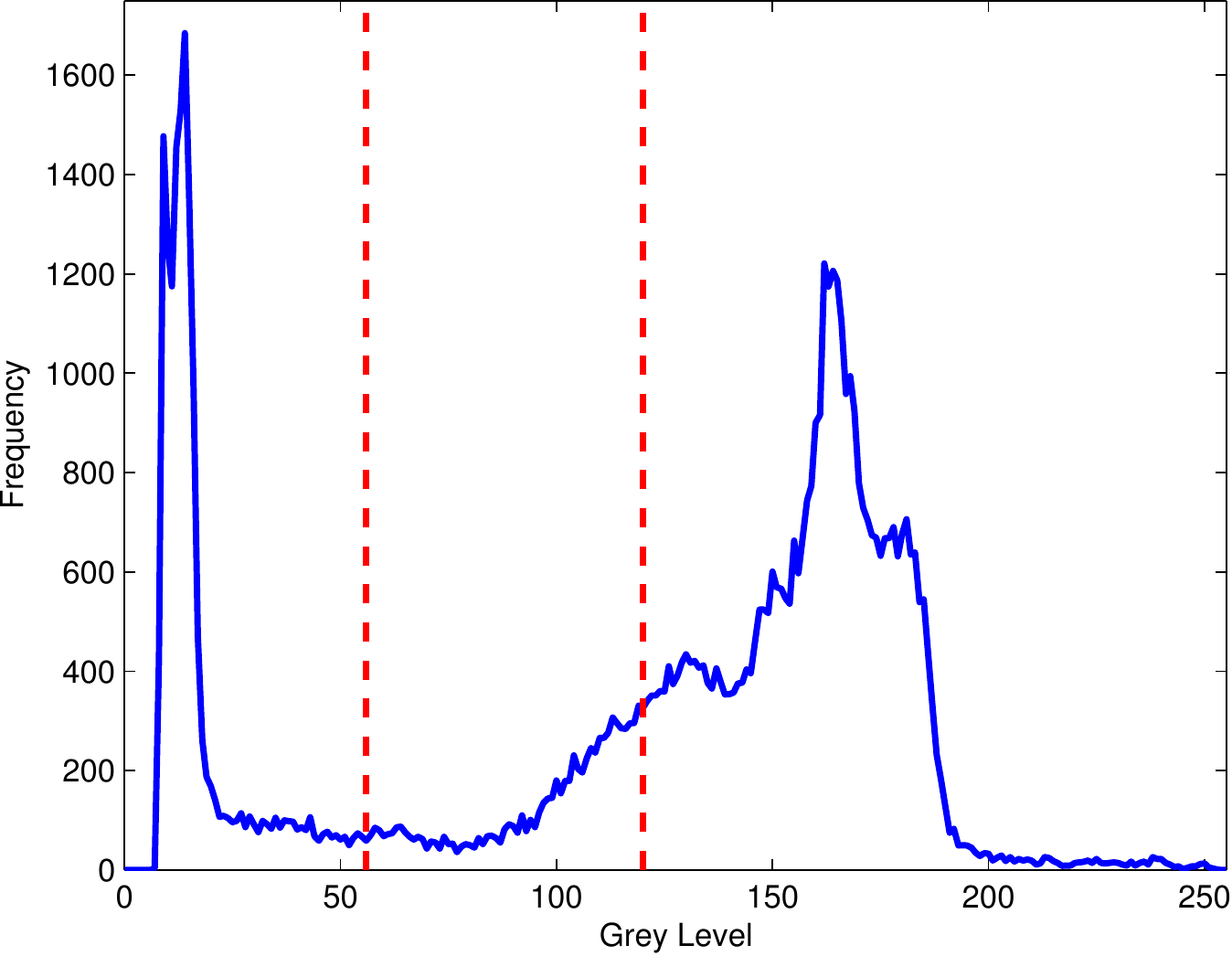}} \:
	\subfloat[threshold=3 \label{fig:Cameramanth3Histogram}]
	{\includegraphics[scale=0.2]{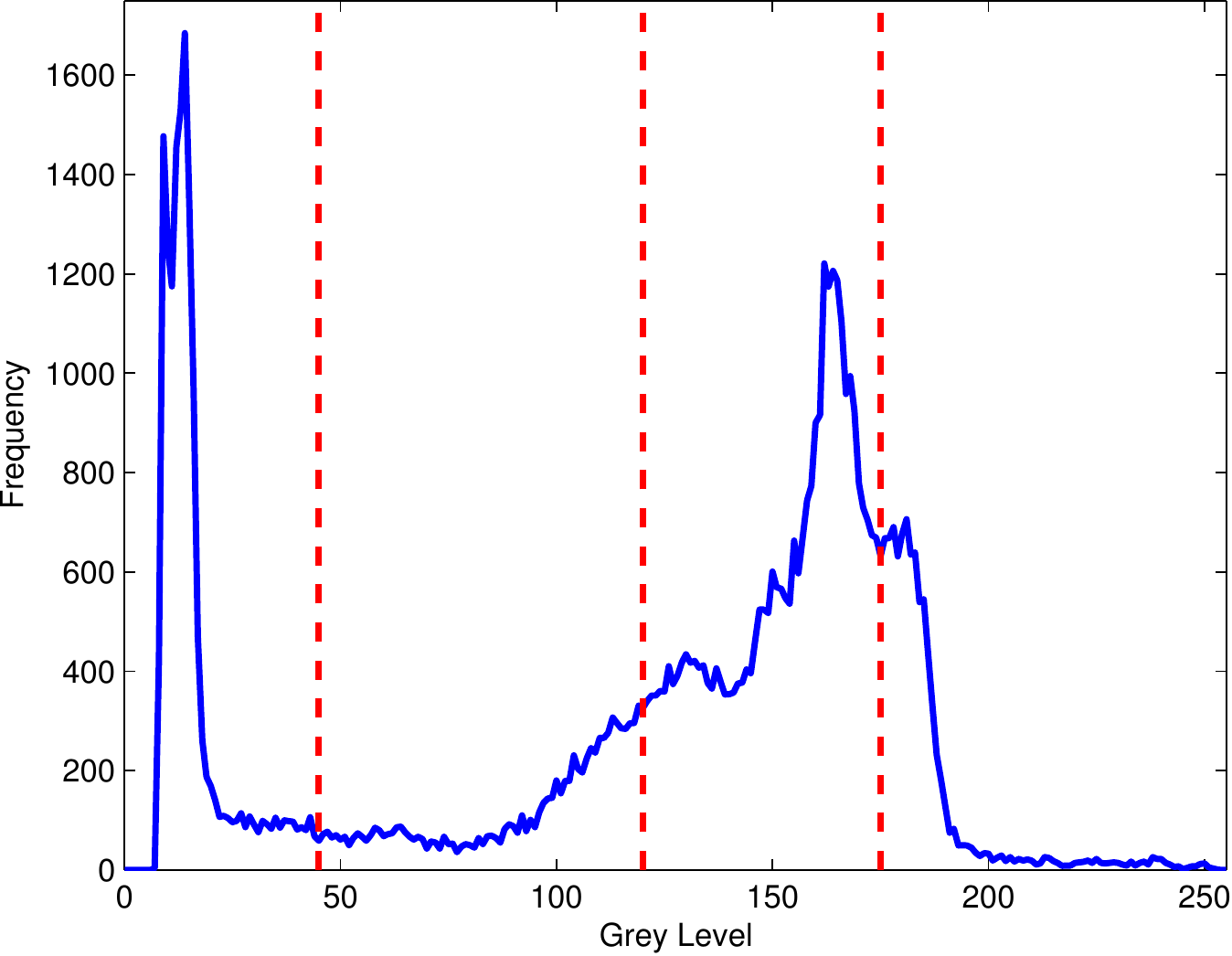}}  \:
	\subfloat[threshold=4 \label{fig:Cameramanth4Histogram}]
	{\includegraphics[scale=0.2]{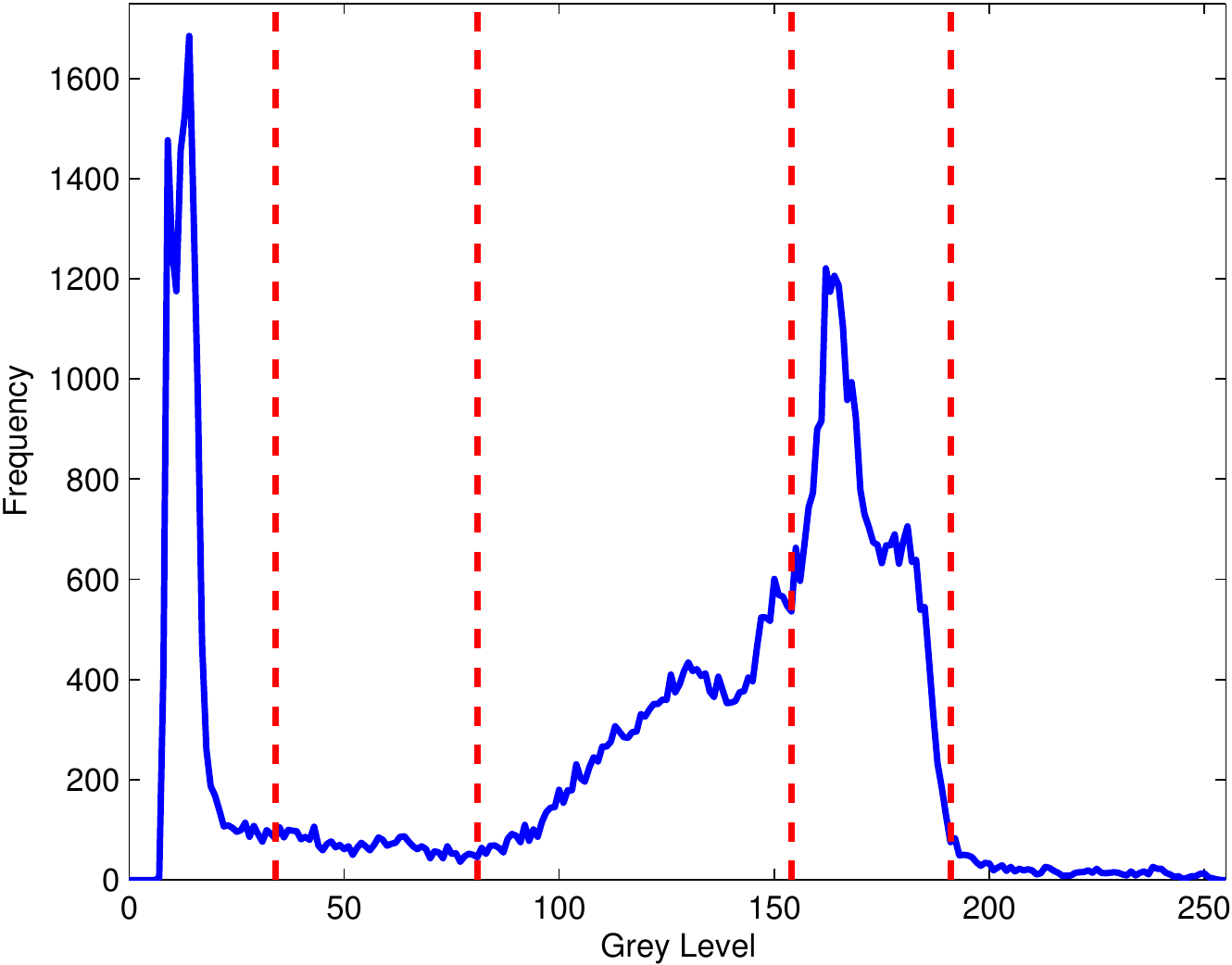}} \:
	\subfloat[threshold=5 \label{fig:Cameramanth5Histogram}]
	{\includegraphics[scale=0.2]{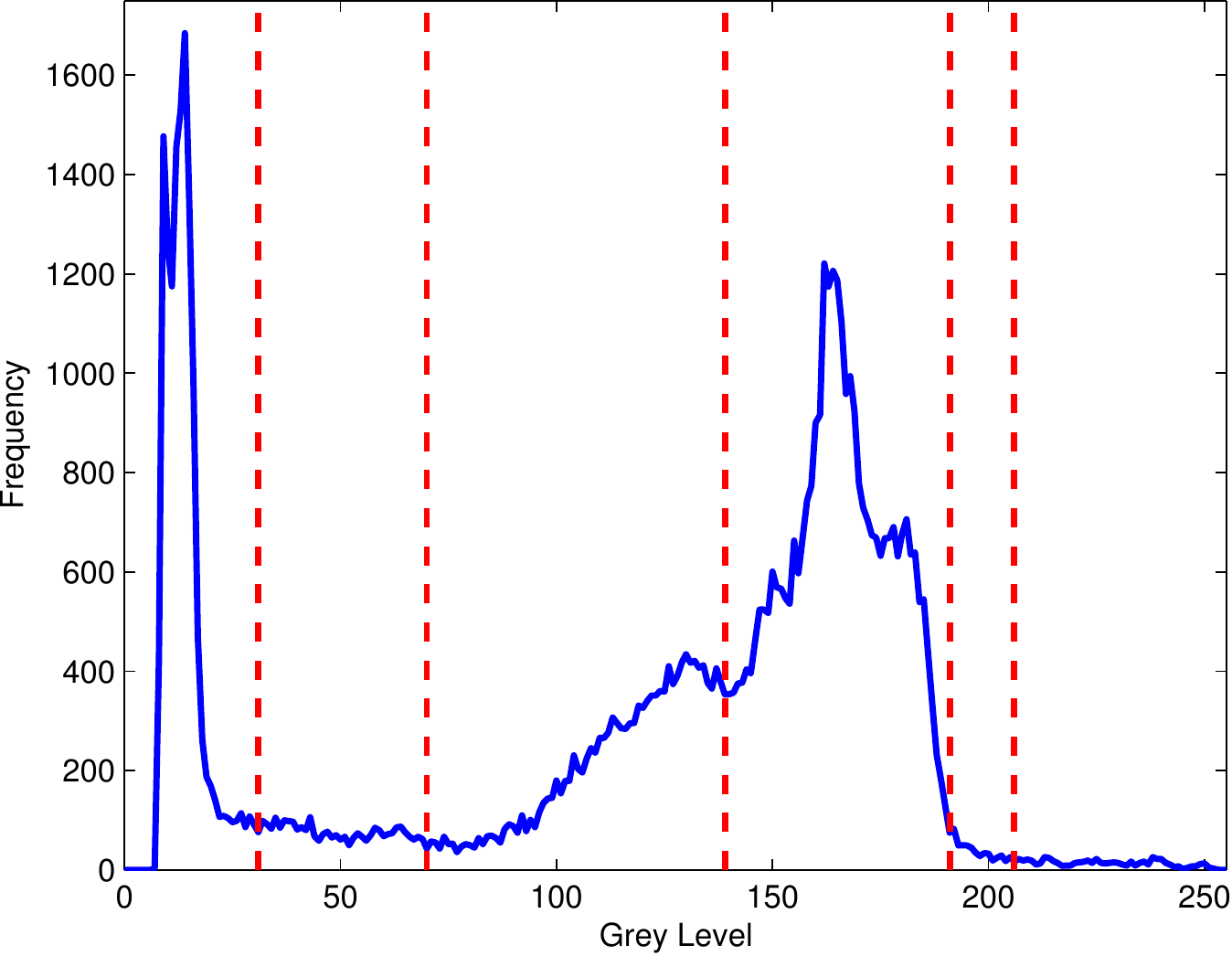}}    
	
	\subfloat[threshold=2 \label{fig:Hunterth2}]
	{\includegraphics[scale=0.19]{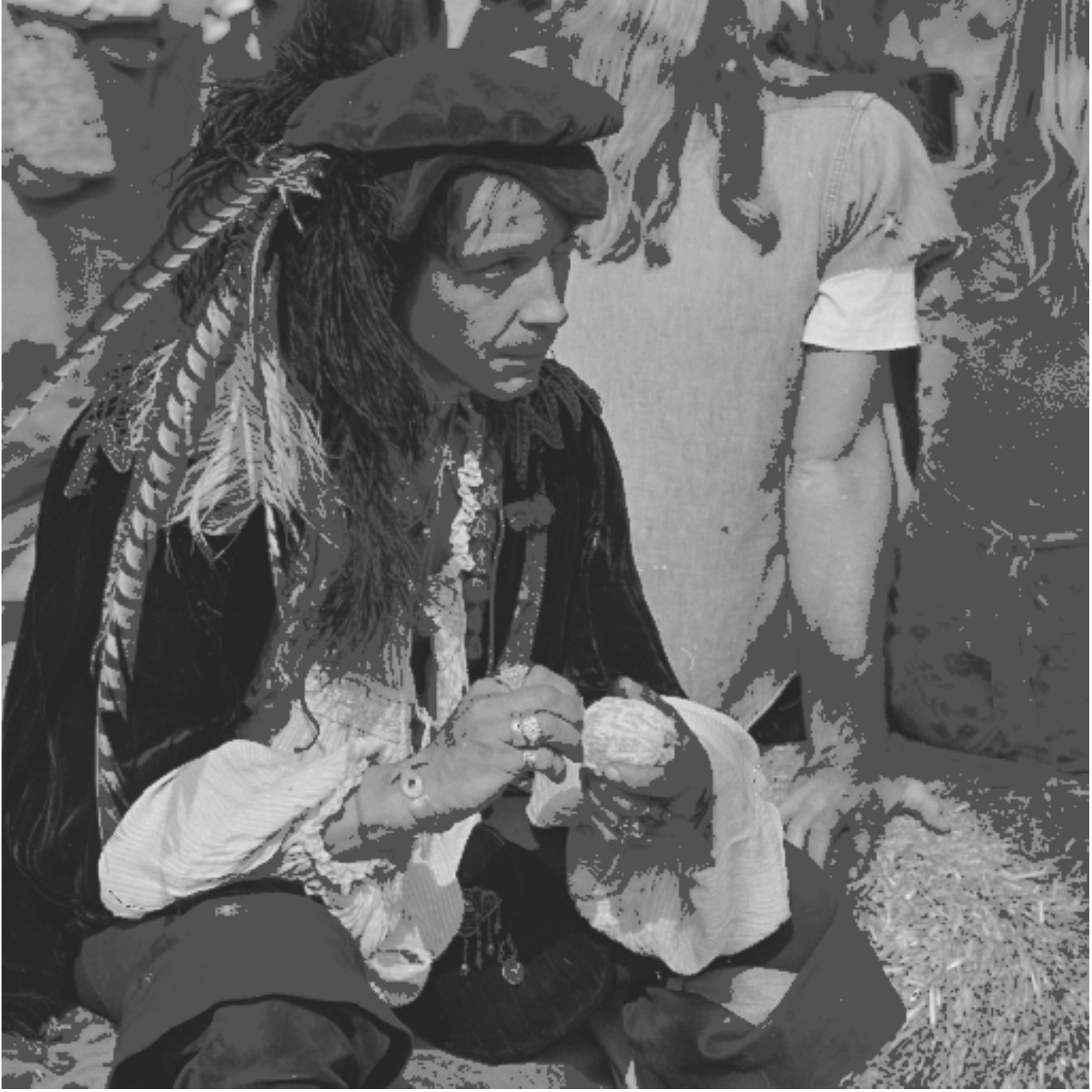}} \:
	\subfloat[threshold=3 \label{fig:Hunterth3}]
	{\includegraphics[scale=0.20]{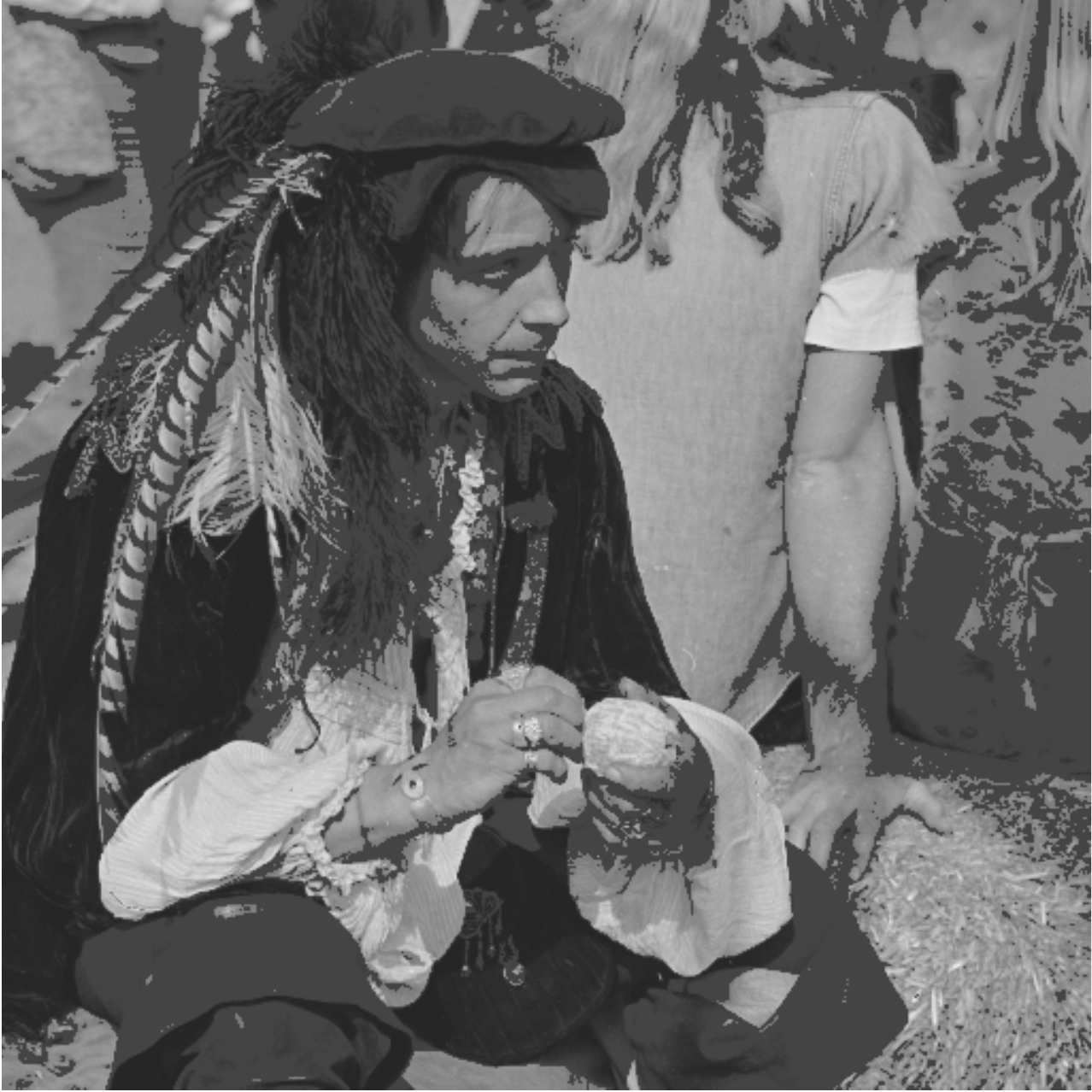}}  \:
	\subfloat[threshold=4 \label{fig:Hunterth4}]
	{\includegraphics[scale=0.19]{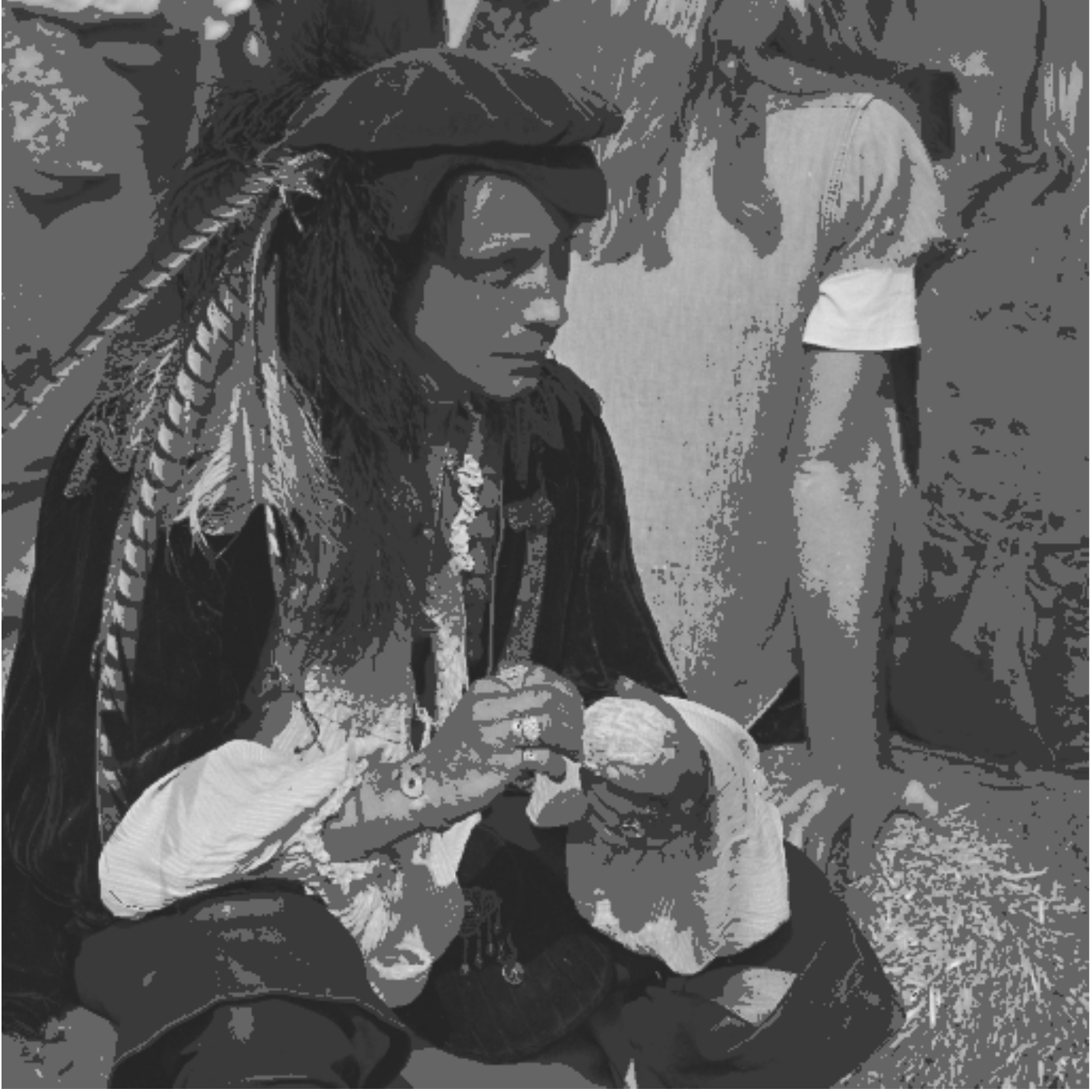}} \:
	\subfloat[threshold=5 \label{fig:Hunterth5}]
	{\includegraphics[scale=0.19]{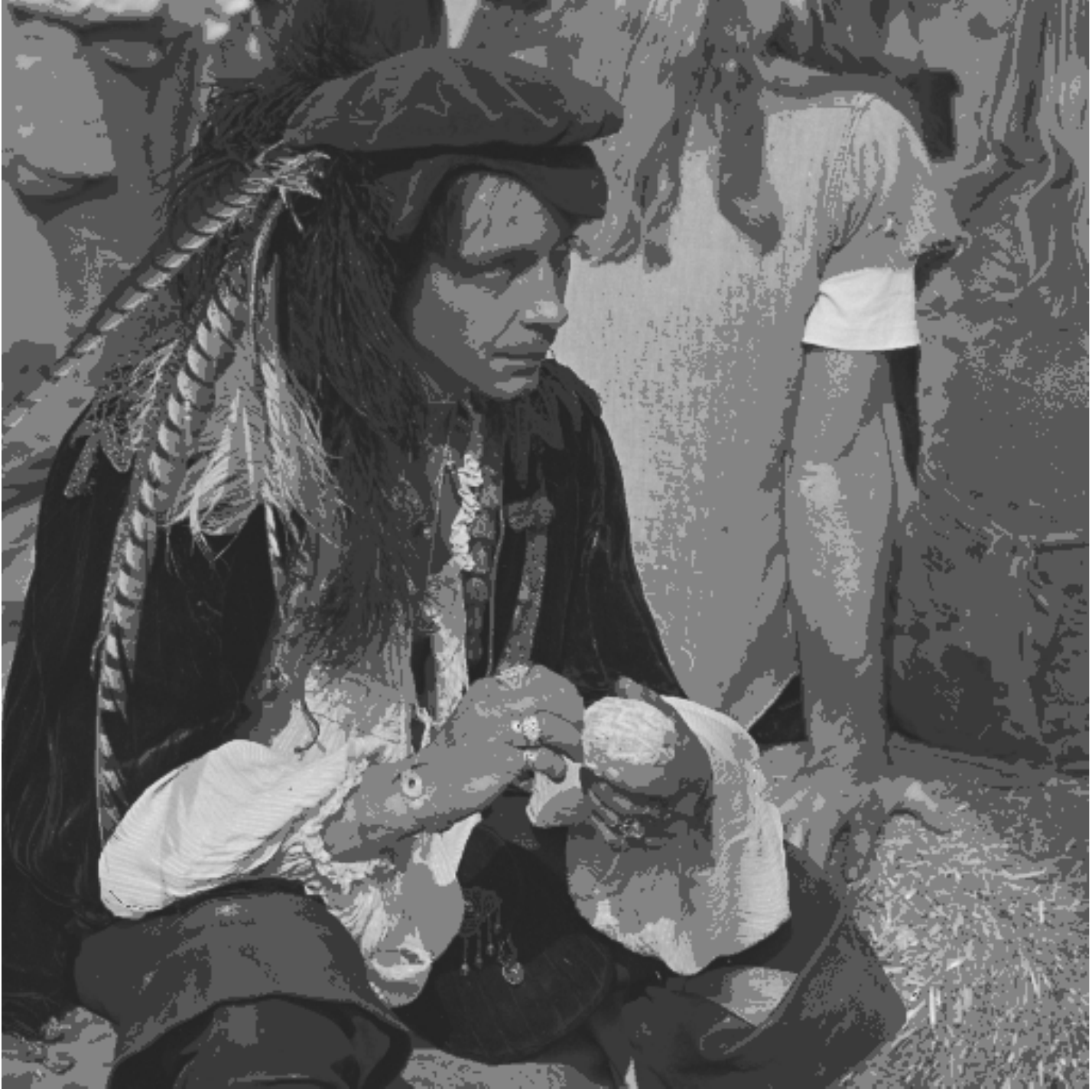}}
	
	\subfloat[threshold=2 \label{fig:Hunterth2Histogram}]
	{\includegraphics[scale=0.21]{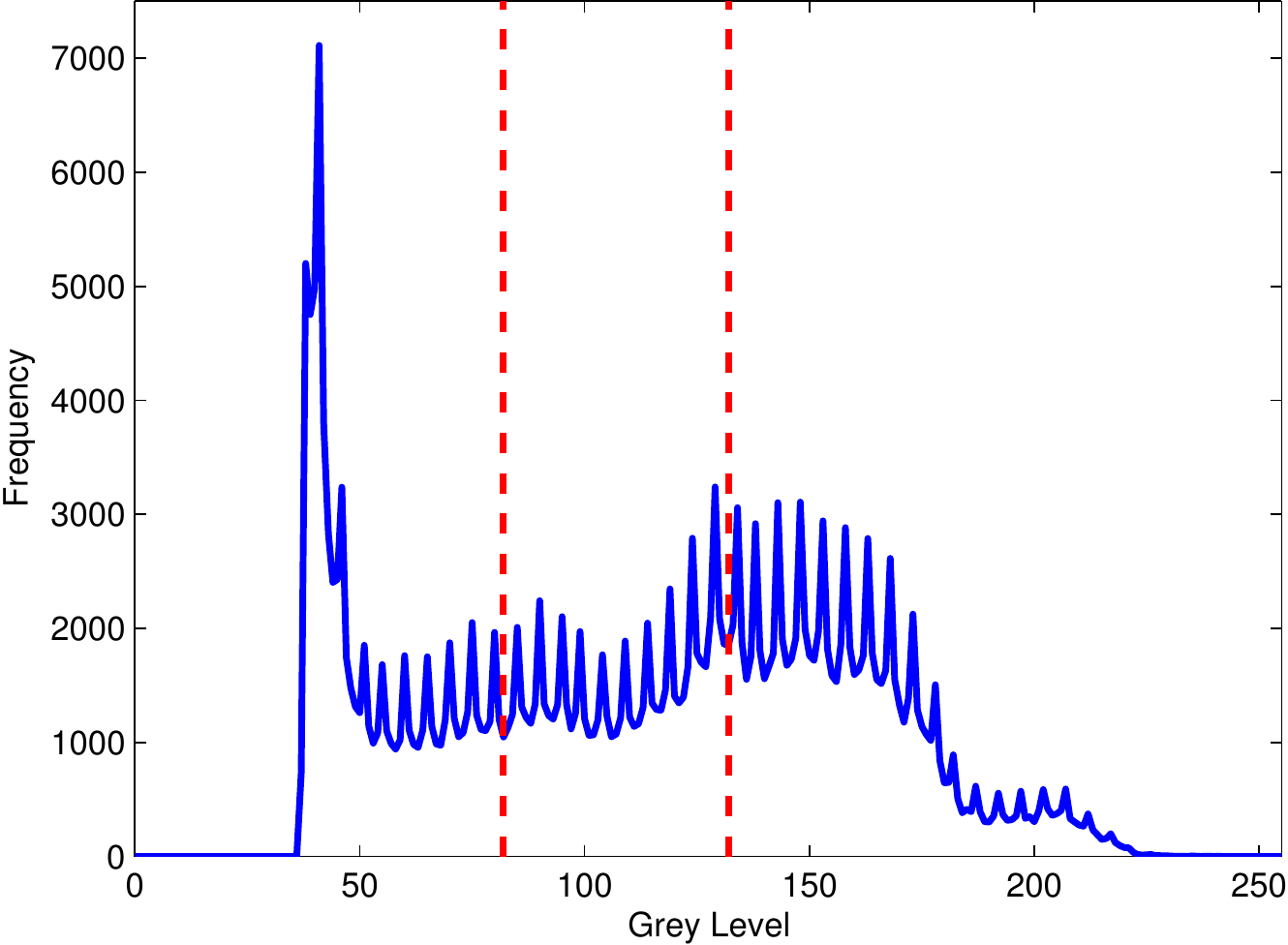}} \:
	\subfloat[threshold=3 \label{fig:Hunterth3Histogram}]
	{\includegraphics[scale=0.20]{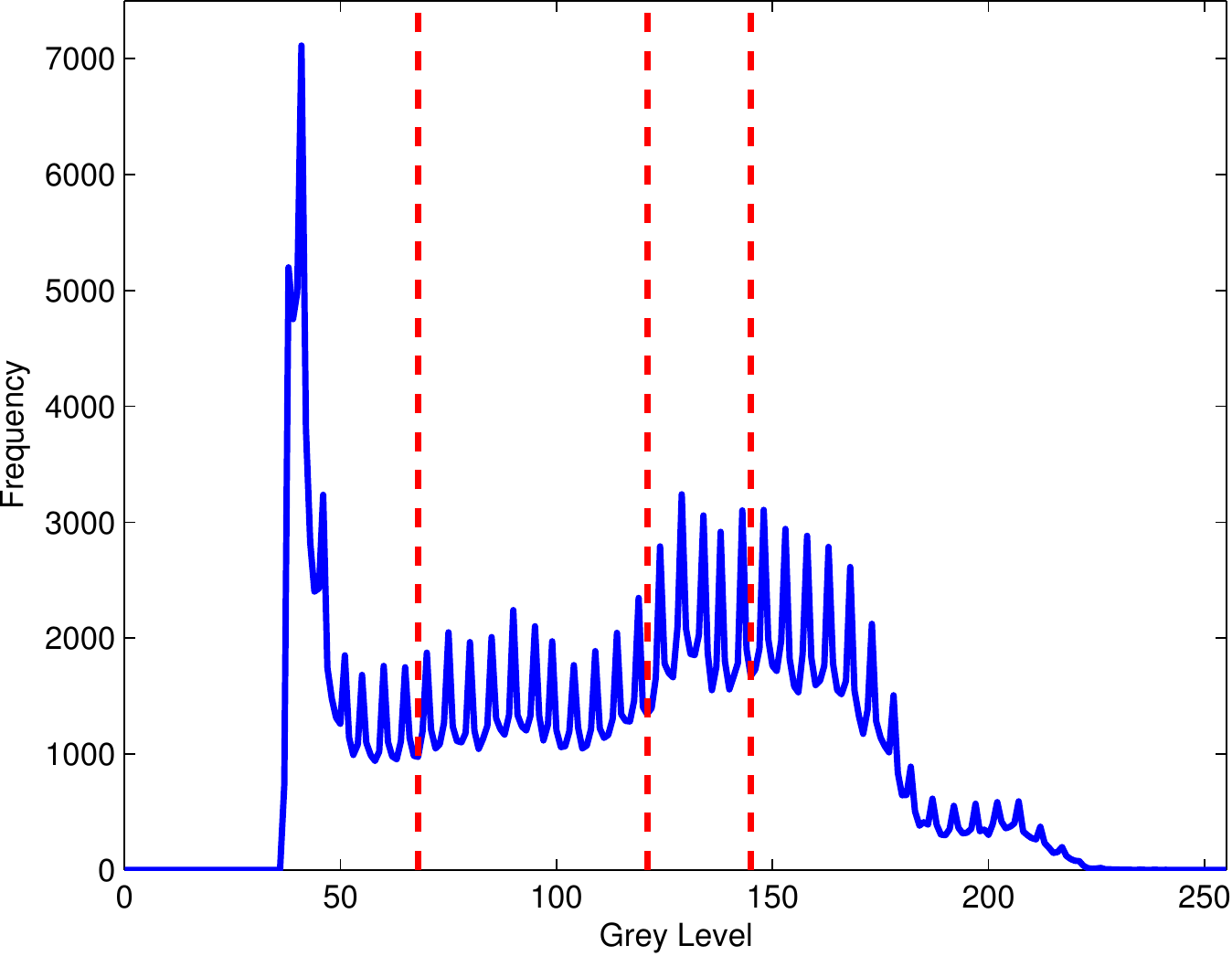}}  \:
	\subfloat[threshold=4 \label{fig:Hunterth4Histogram}]
	{\includegraphics[scale=0.20]{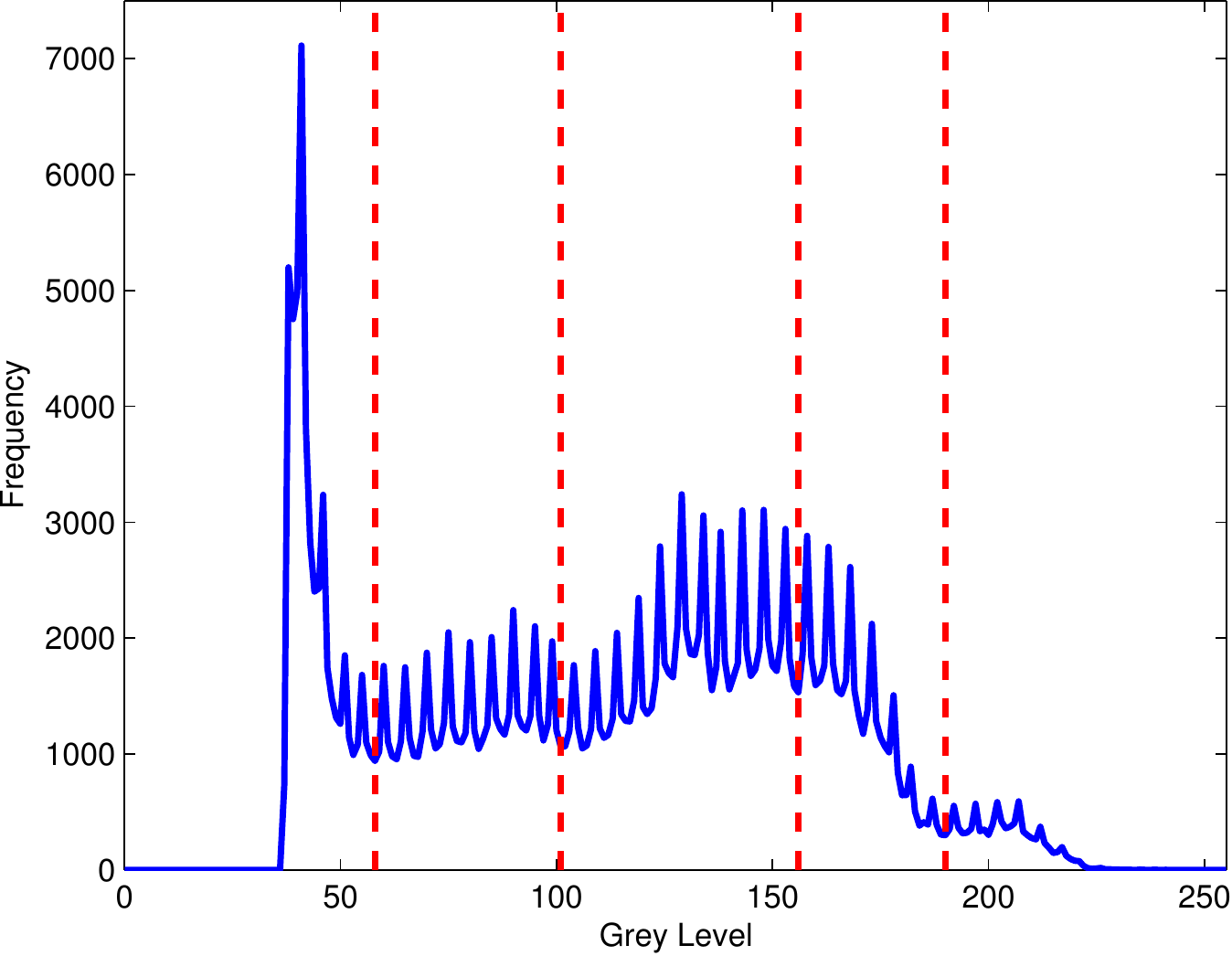}} \:
	\subfloat[threshold=5 \label{fig:Hunterth5Histogram}]
	{\includegraphics[scale=0.20]{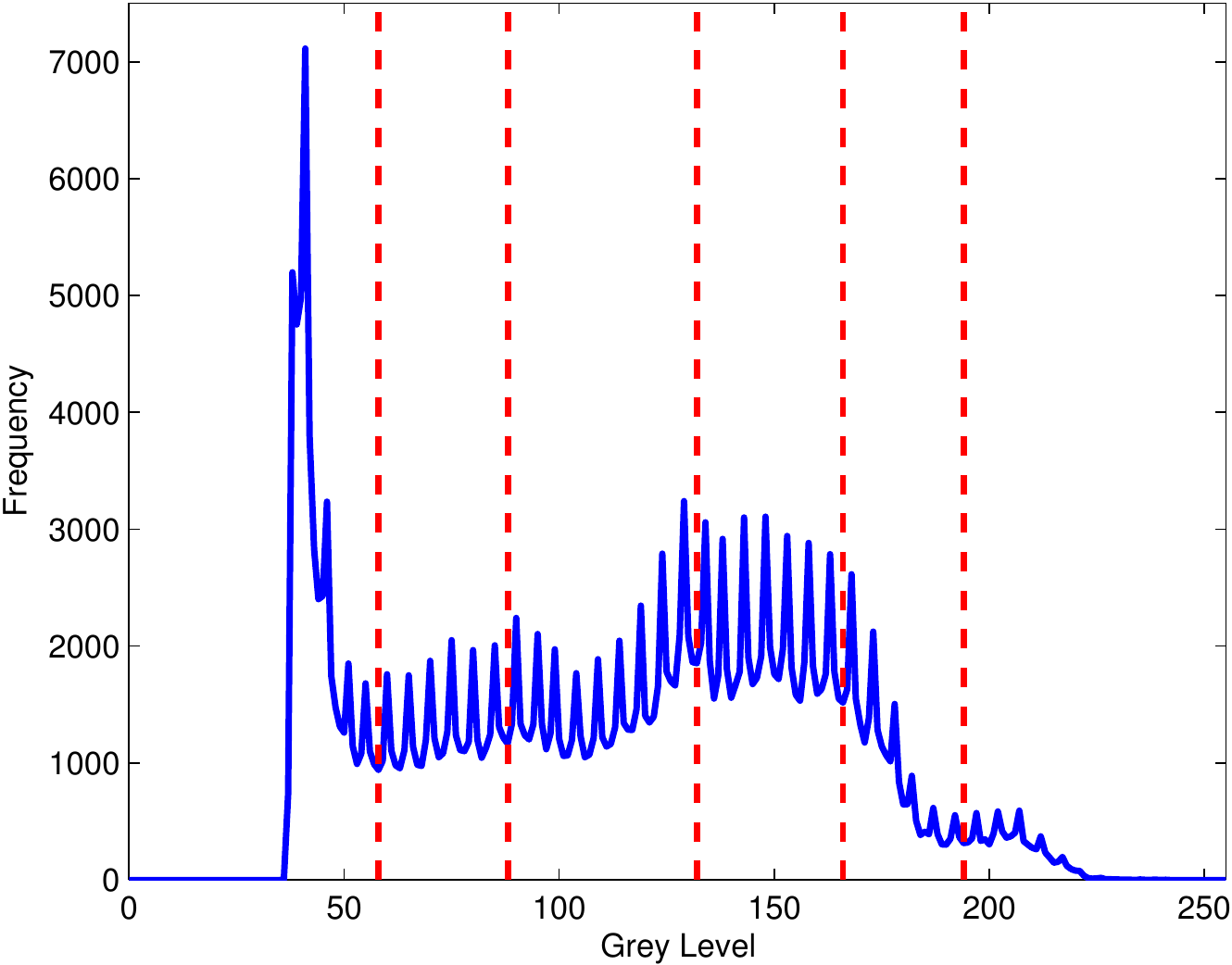}}
	
	\caption{Result obtained using our approach on the benchmark Lena, Cameraman and Hunter. \label{fig:thresholding}}
\end{figure*}

\begin{figure*}[p] 
	\centering{} 
	\subfloat[threshold=2 \label{fig:Baboonth2}]
	{\includegraphics[scale=0.195]{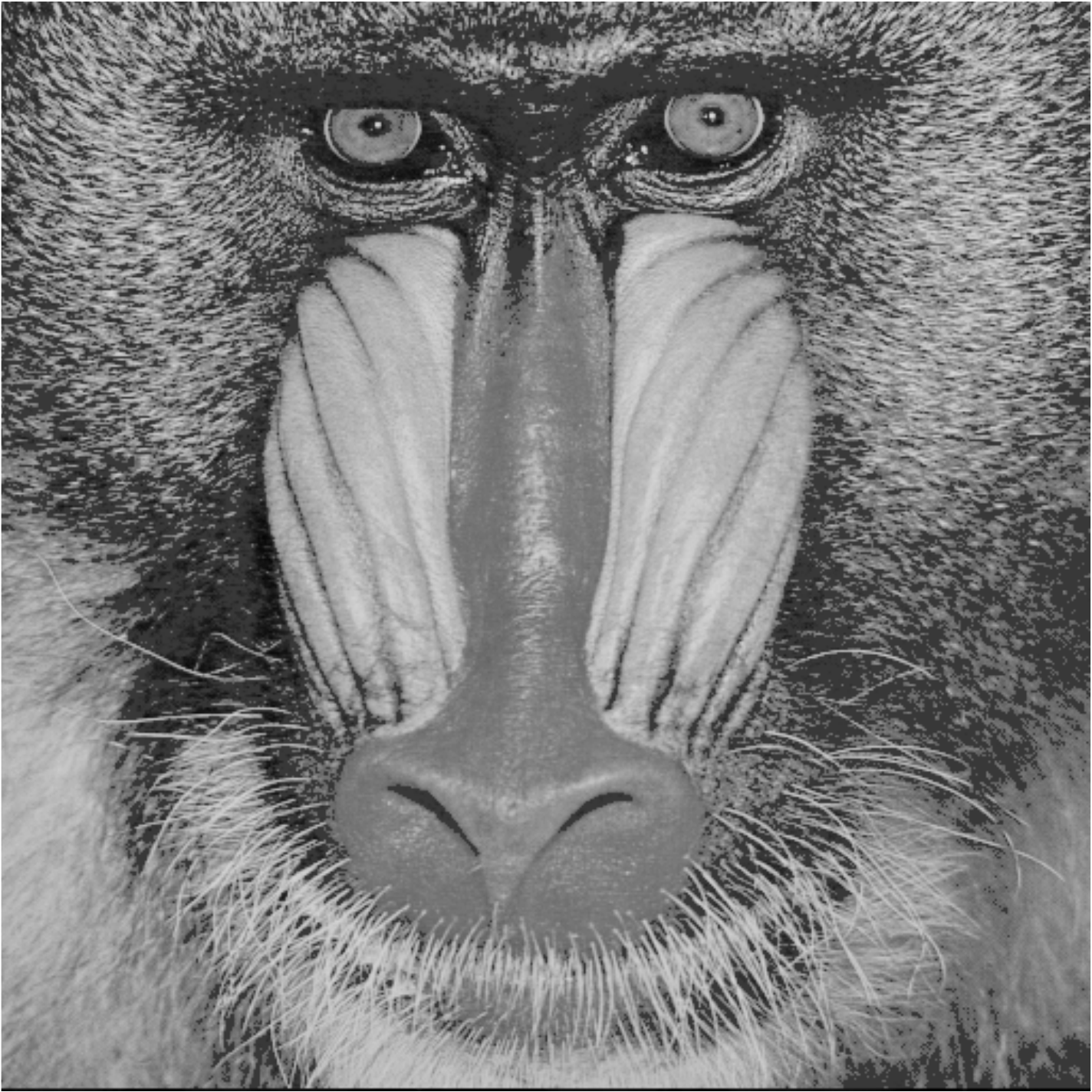}} \:
	\subfloat[threshold=3 \label{fig:Baboonth3}]
	{\includegraphics[scale=0.195]{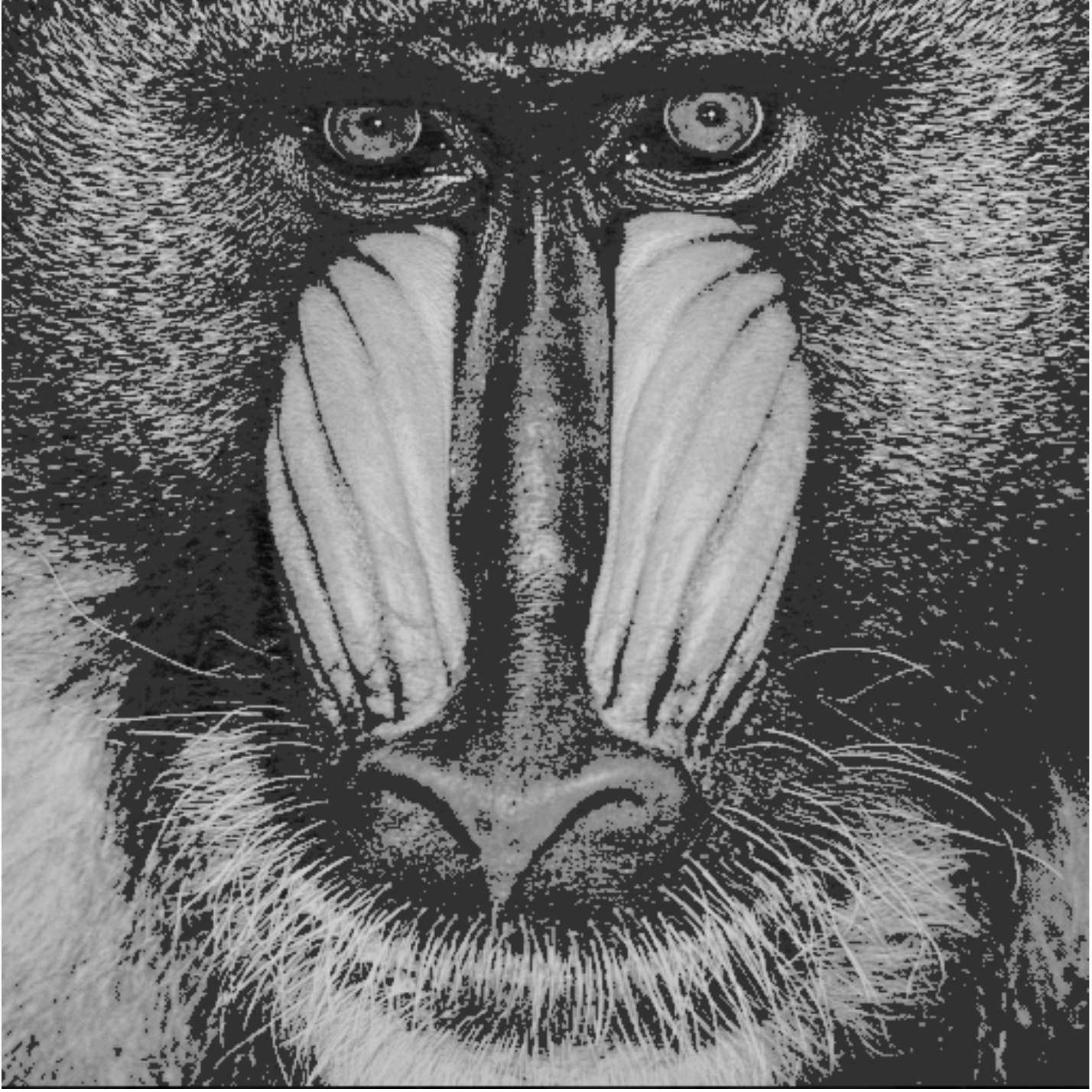}}  \:
	\subfloat[threshold=4 \label{fig:Baboonth4}]
	{\includegraphics[scale=0.195]{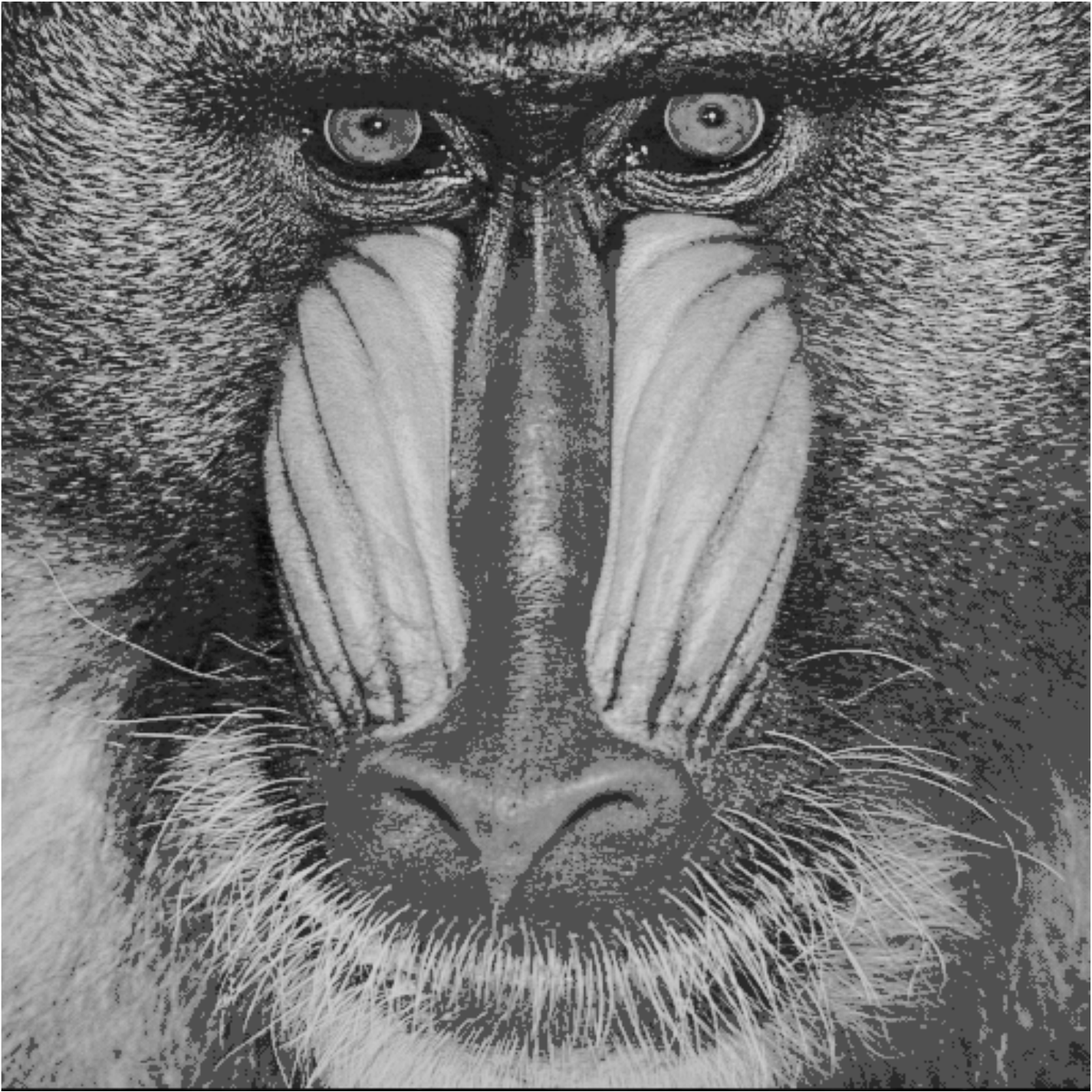}} \:
	\subfloat[threshold=5 \label{fig:Baboonth5}]
	{\includegraphics[scale=0.195]{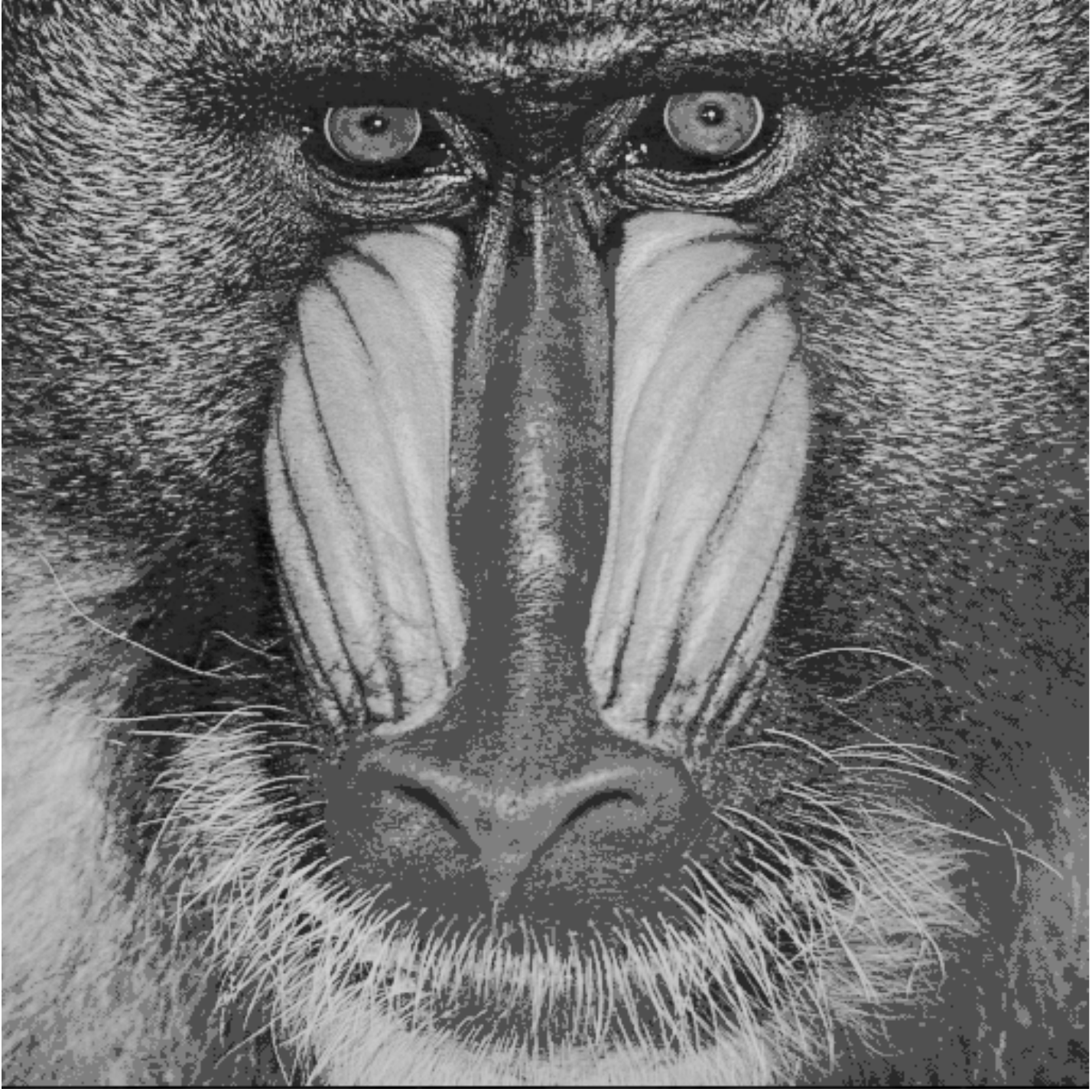}}     
	
	\subfloat[threshold=2 \label{fig:Baboonth2Histogram}]
	{\includegraphics[scale=0.21]{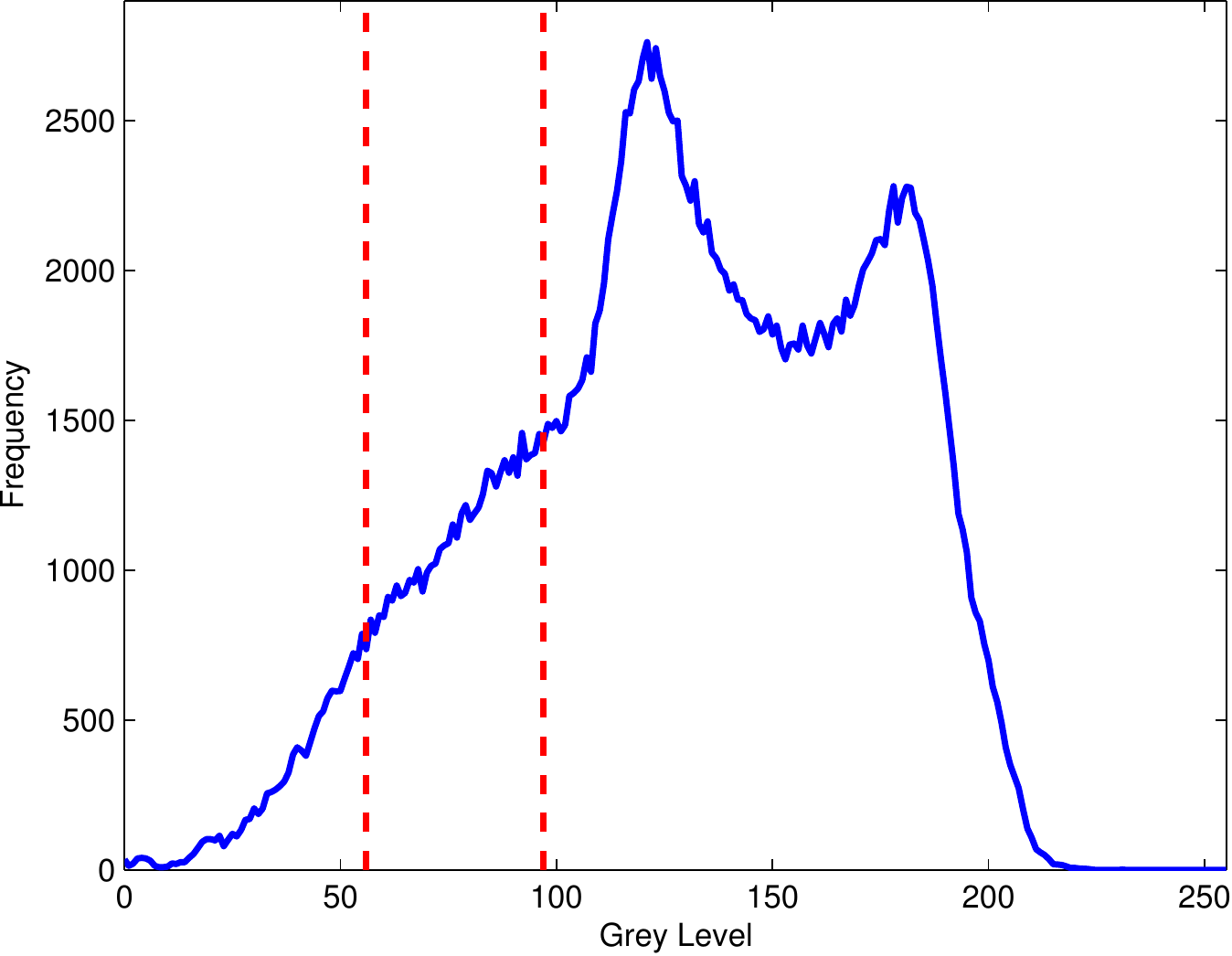}} \:
	\subfloat[threshold=3 \label{fig:Baboonth3Histogram}]
	{\includegraphics[scale=0.21]{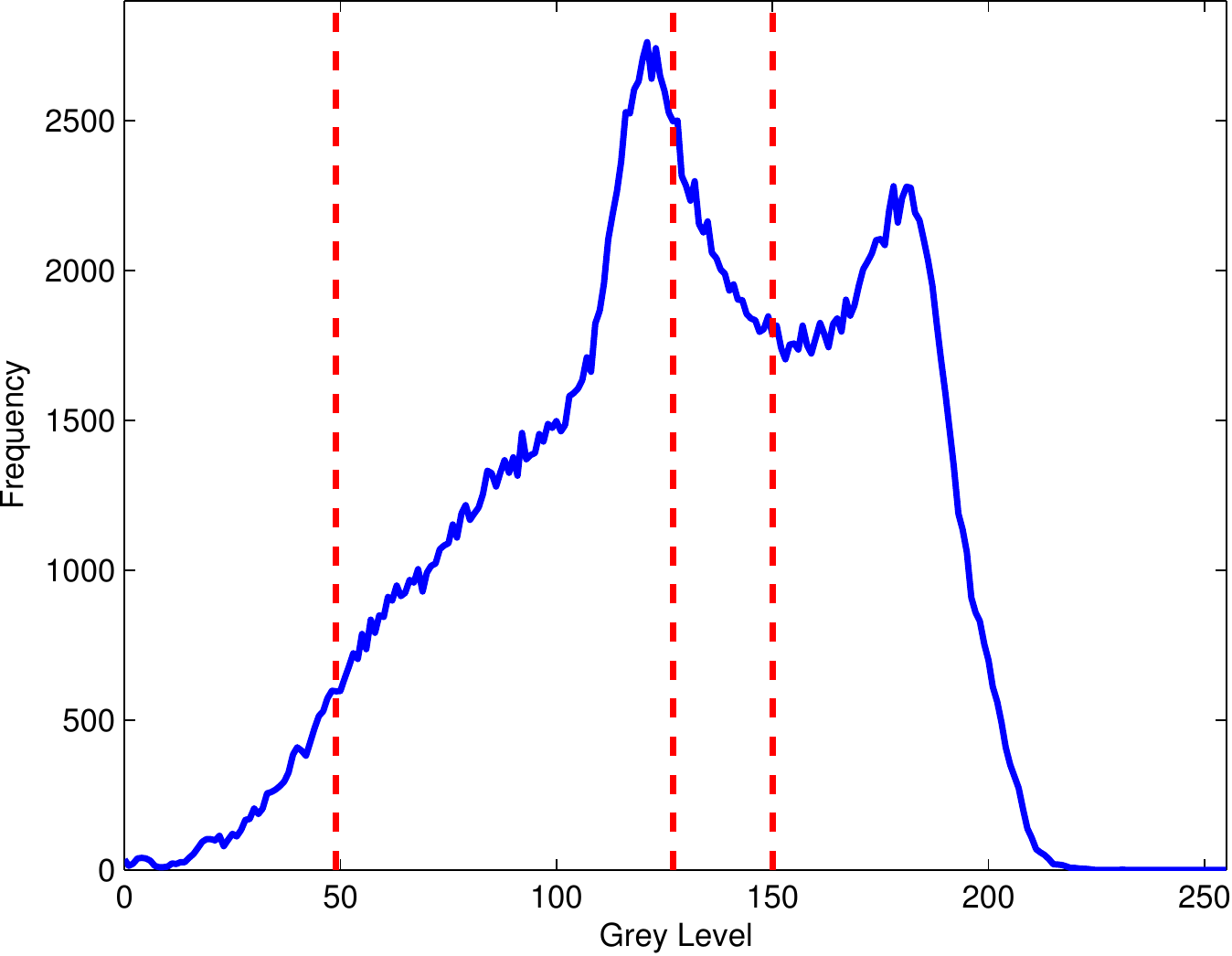}}  \:
	\subfloat[threshold=4 \label{fig:Baboonth4Histogram}]
	{\includegraphics[scale=0.21]{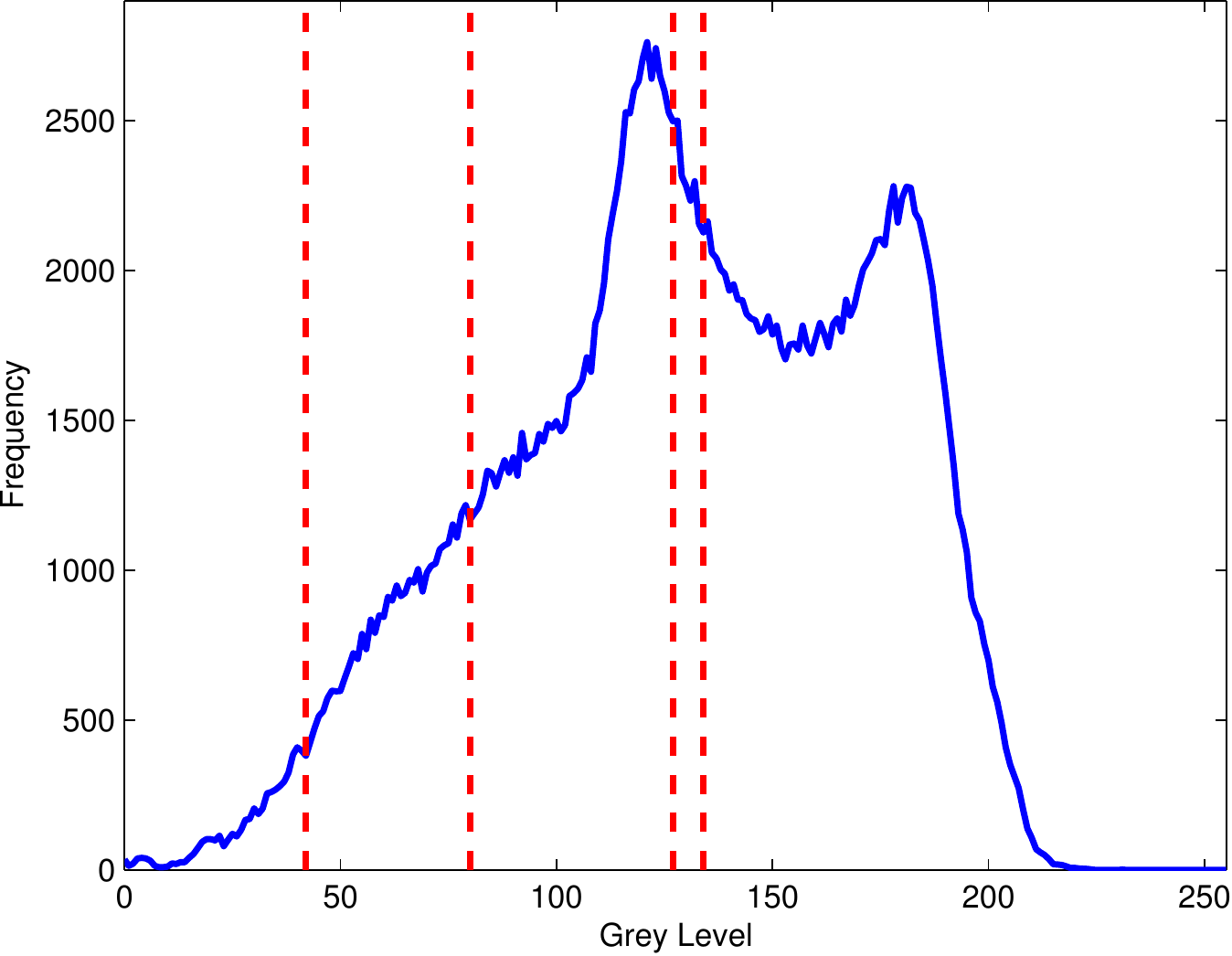}} \:
	\subfloat[threshold=5 \label{fig:Baboonth5Histogram}]
	{\includegraphics[scale=0.21]{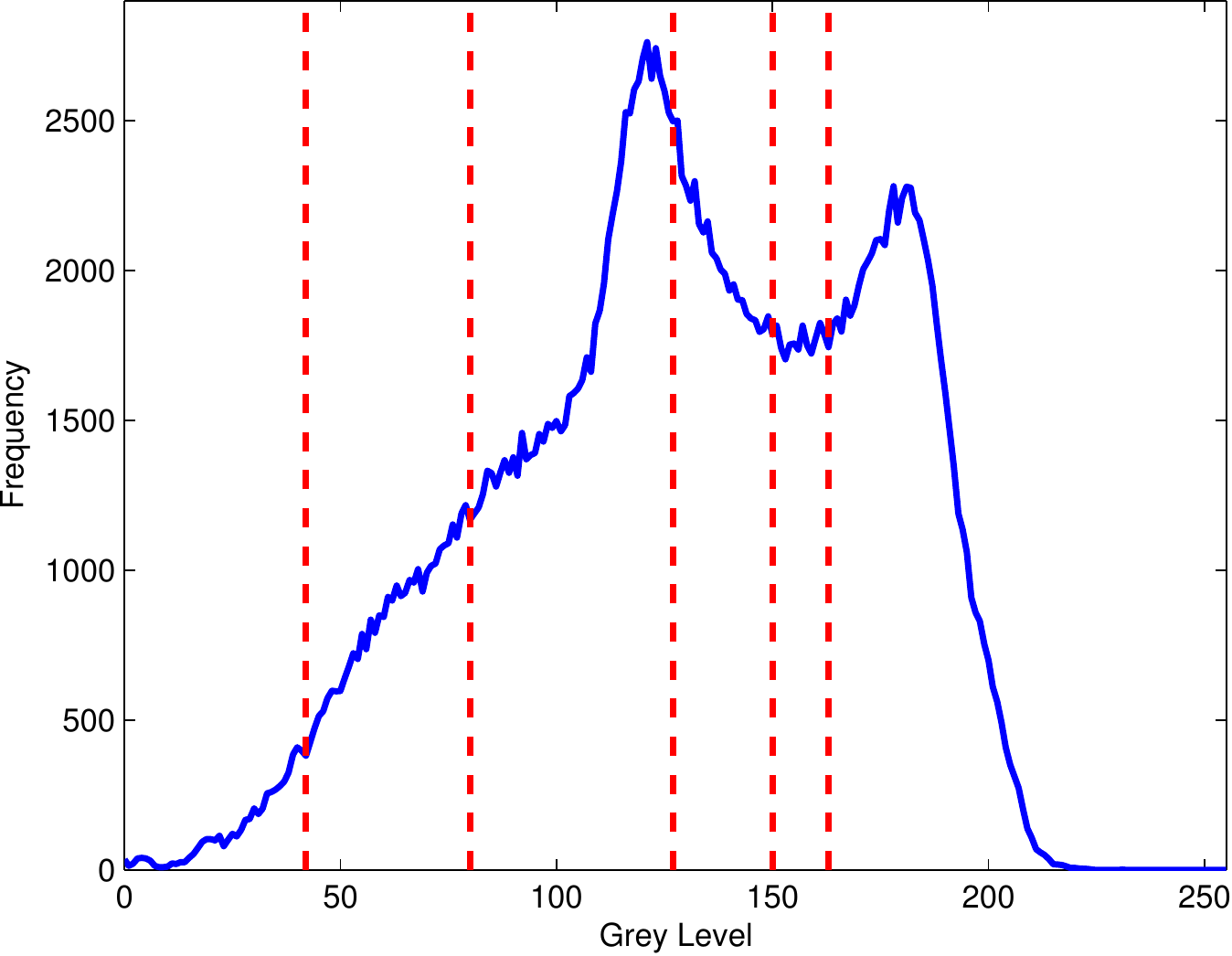}}
	
	\subfloat[threshold=2 \label{fig:Fruitsth2}]
	{\includegraphics[scale=0.192]{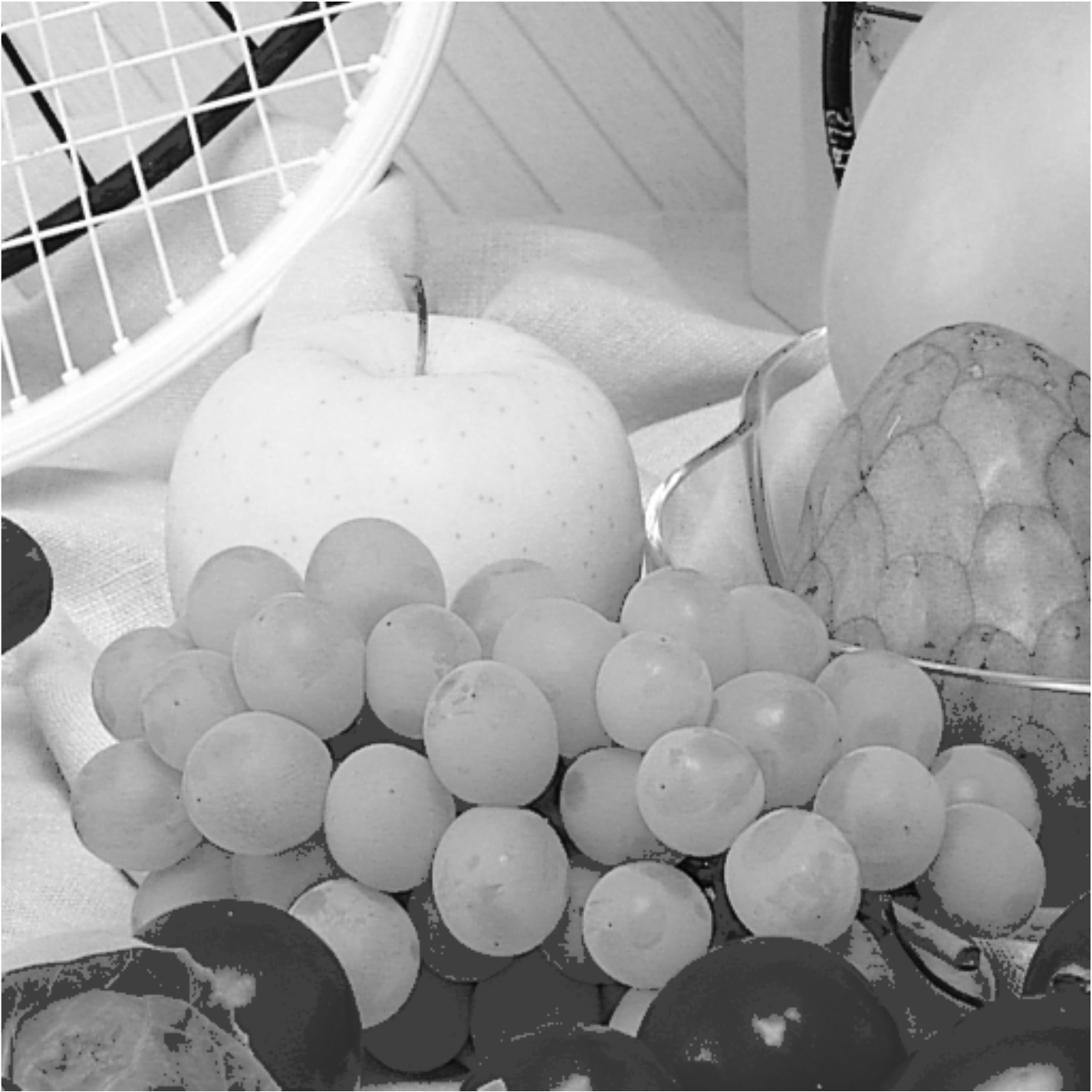}} \:
	\subfloat[threshold=3 \label{fig:Fruitsth3}]
	{\includegraphics[scale=0.192]{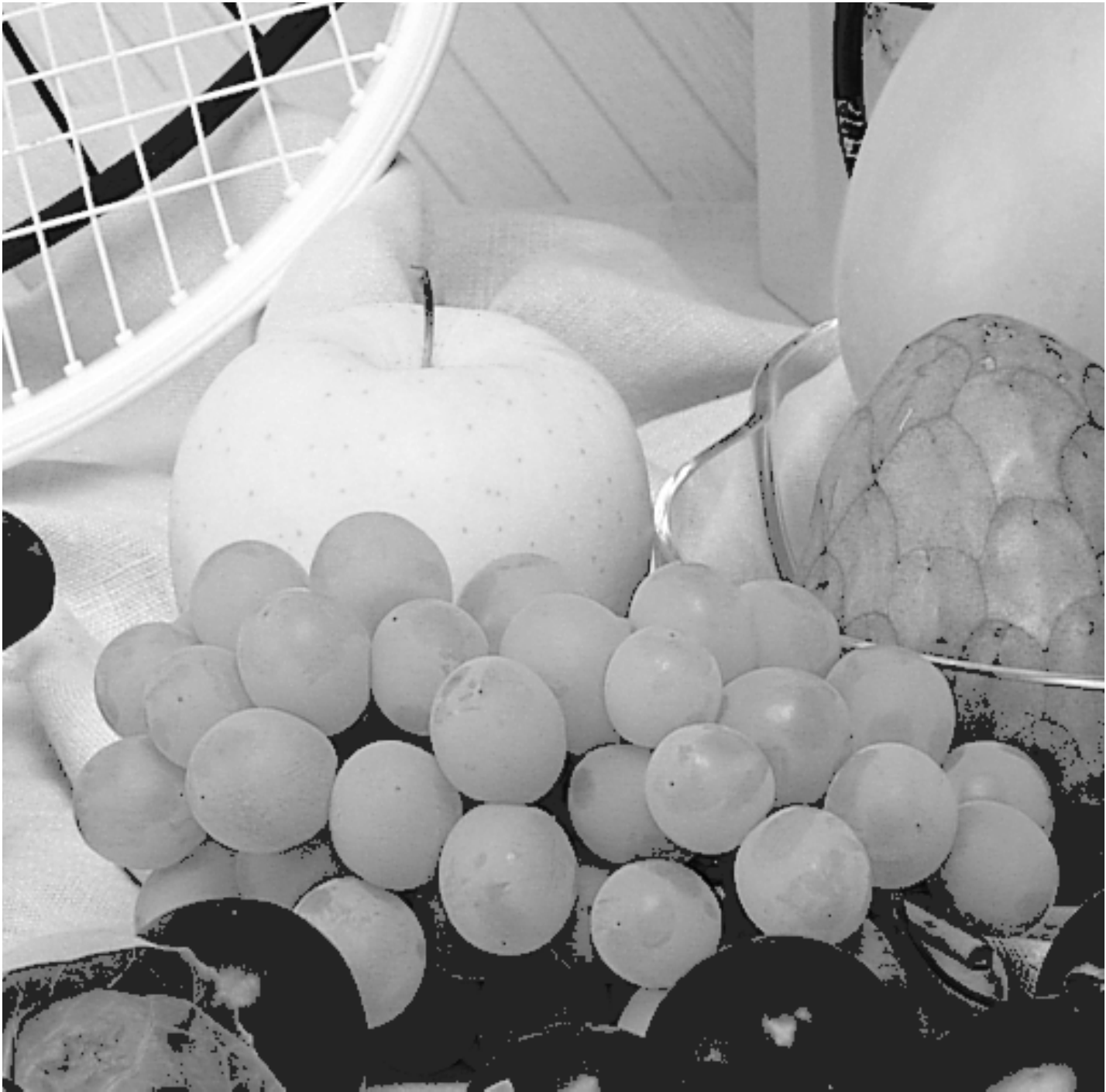}} \:
	\subfloat[threshold=4 \label{fig:Fruitsth4}]
	{\includegraphics[scale=0.192]{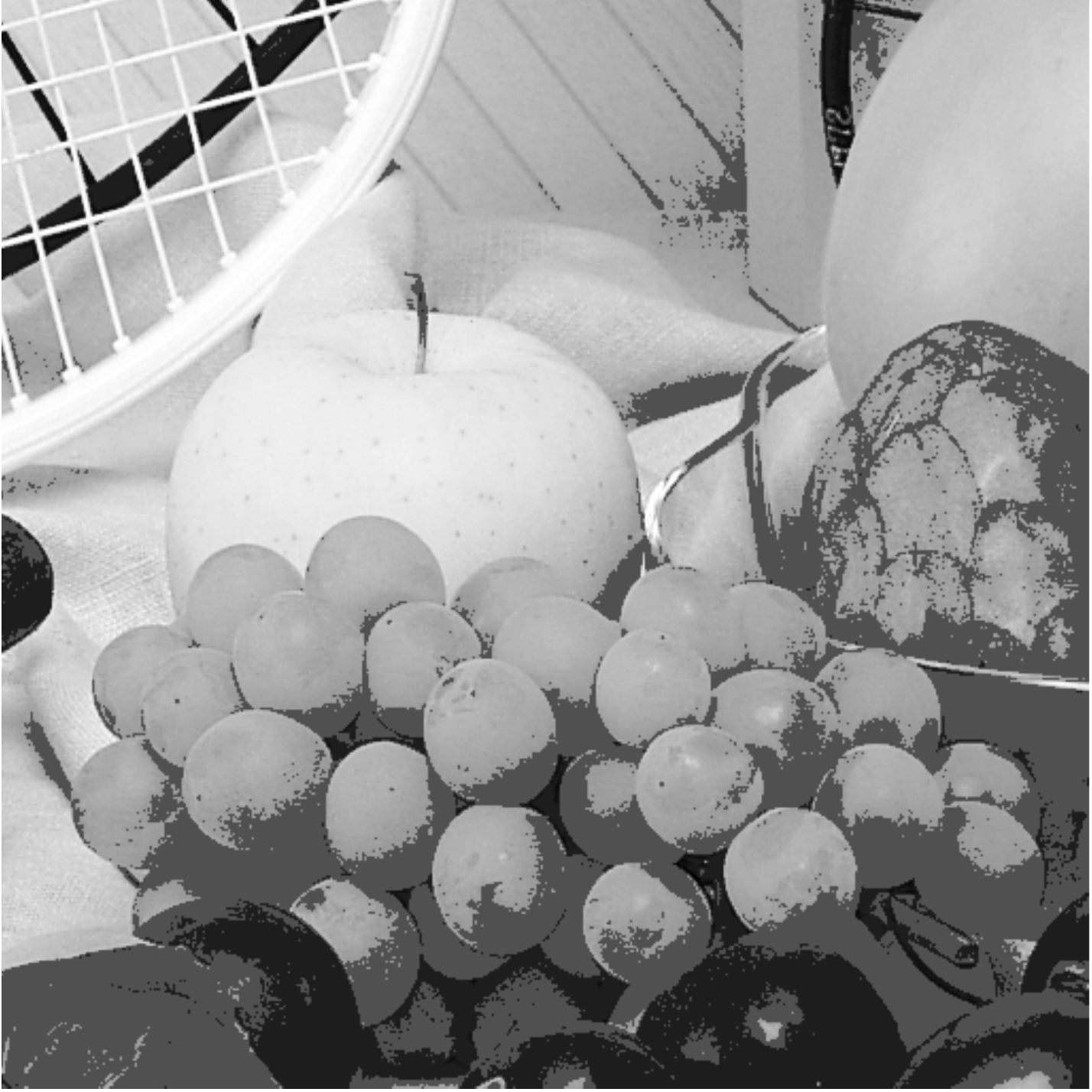}} \:
	\subfloat[threshold=5 \label{fig:Fruitsth5}]
	{\includegraphics[scale=0.192]{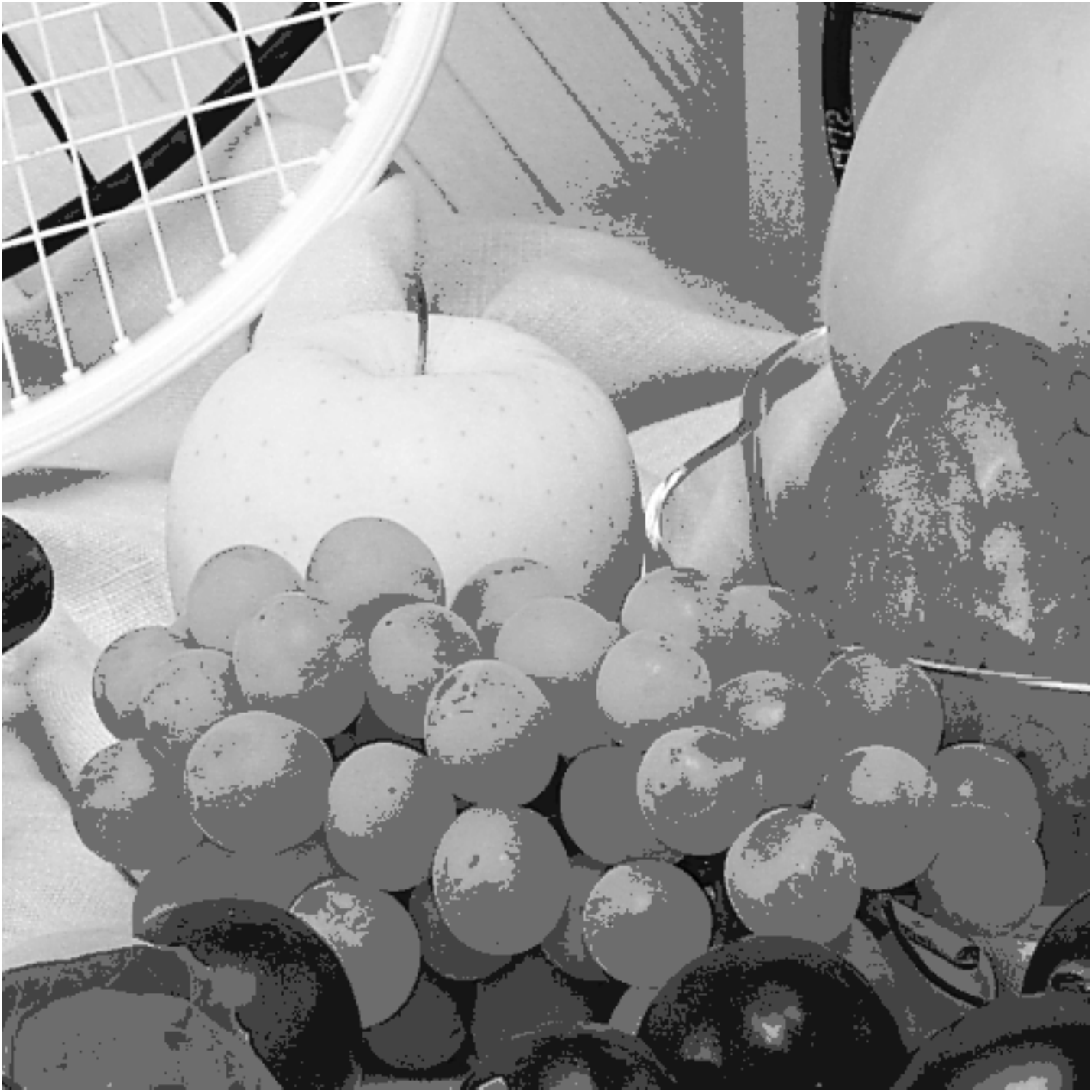}}
	
	\subfloat[threshold=2 \label{fig:Fruitth2Histogram}]
	{\includegraphics[scale=0.20]{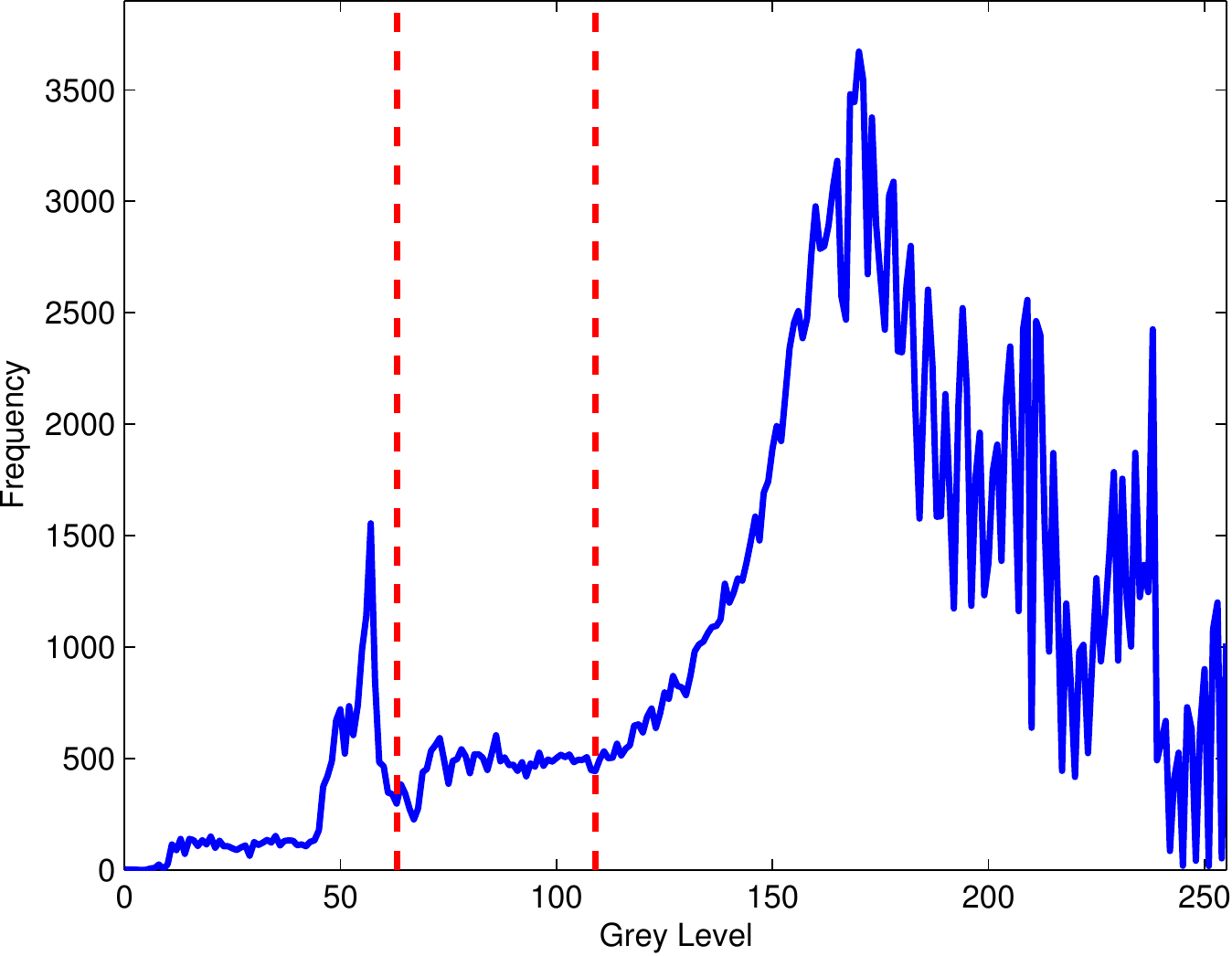}} \:
	\subfloat[threshold=3 \label{fig:Fruitth3Histogram}]
	{\includegraphics[scale=0.20]{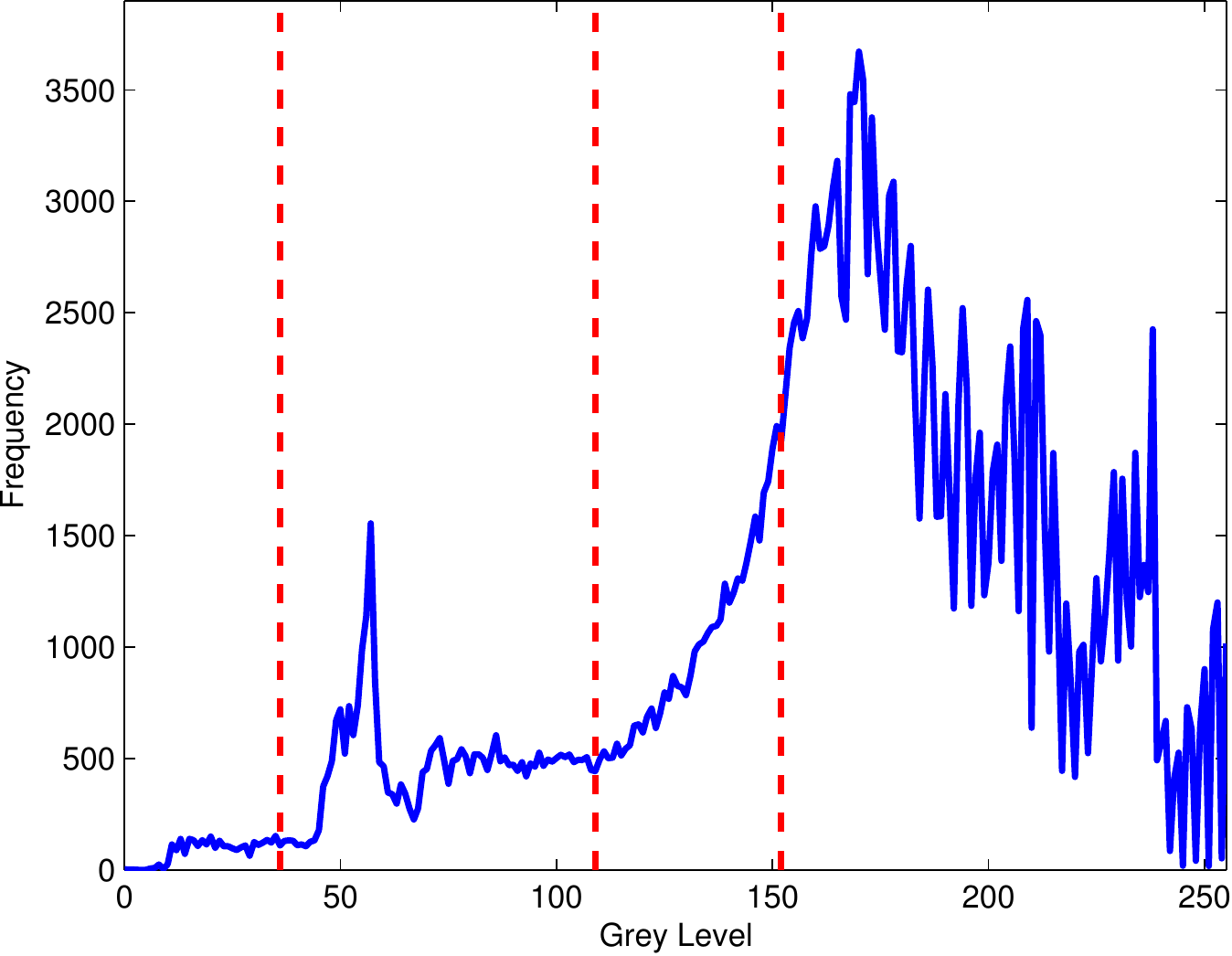}}  \:
	\subfloat[threshold=4 \label{fig:Fruitth4Histogram}]
	{\includegraphics[scale=0.214]{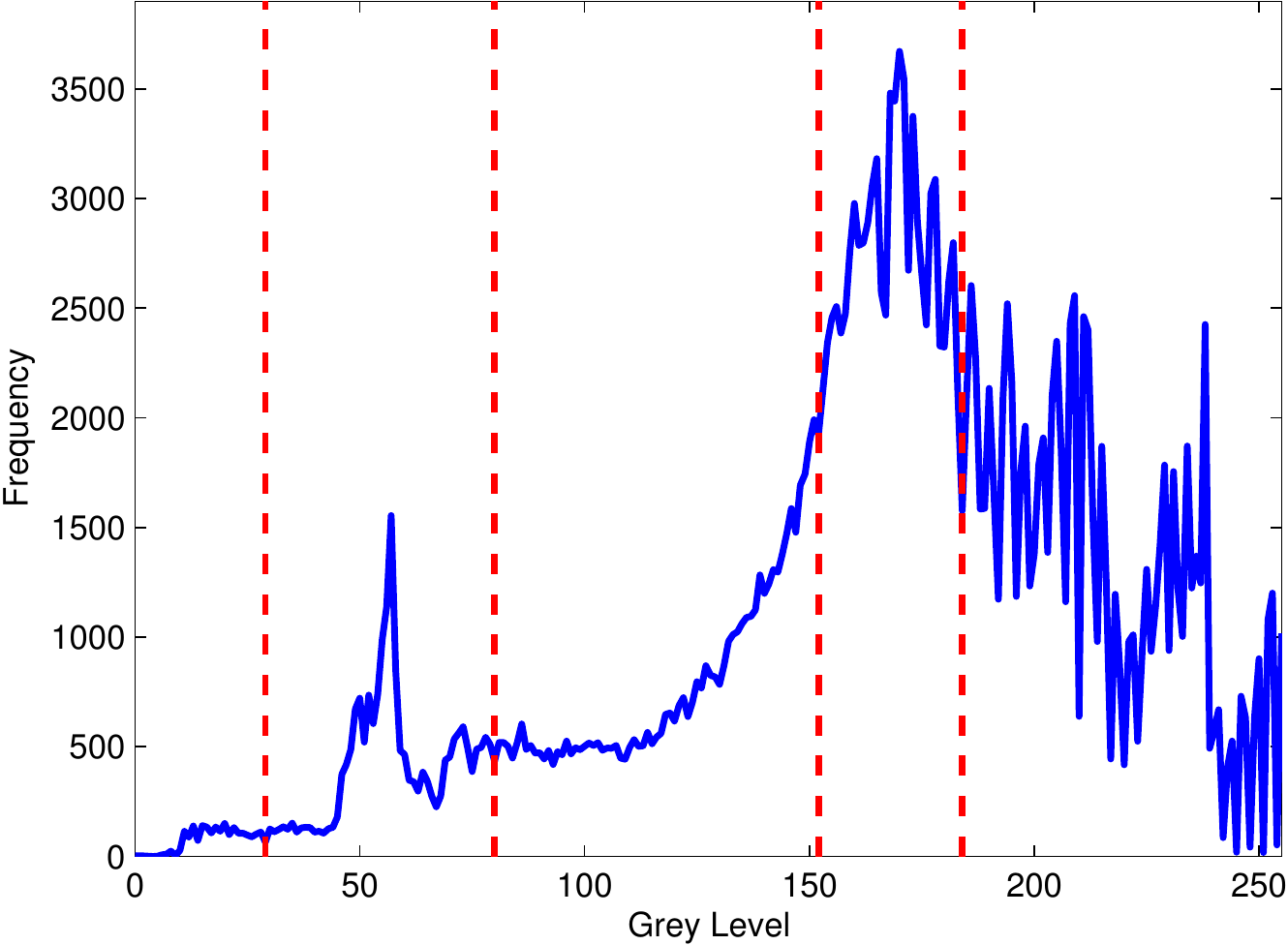}} \:
	\subfloat[threshold=5 \label{fig:Fruitth5Histogram}]
	{\includegraphics[scale=0.20]{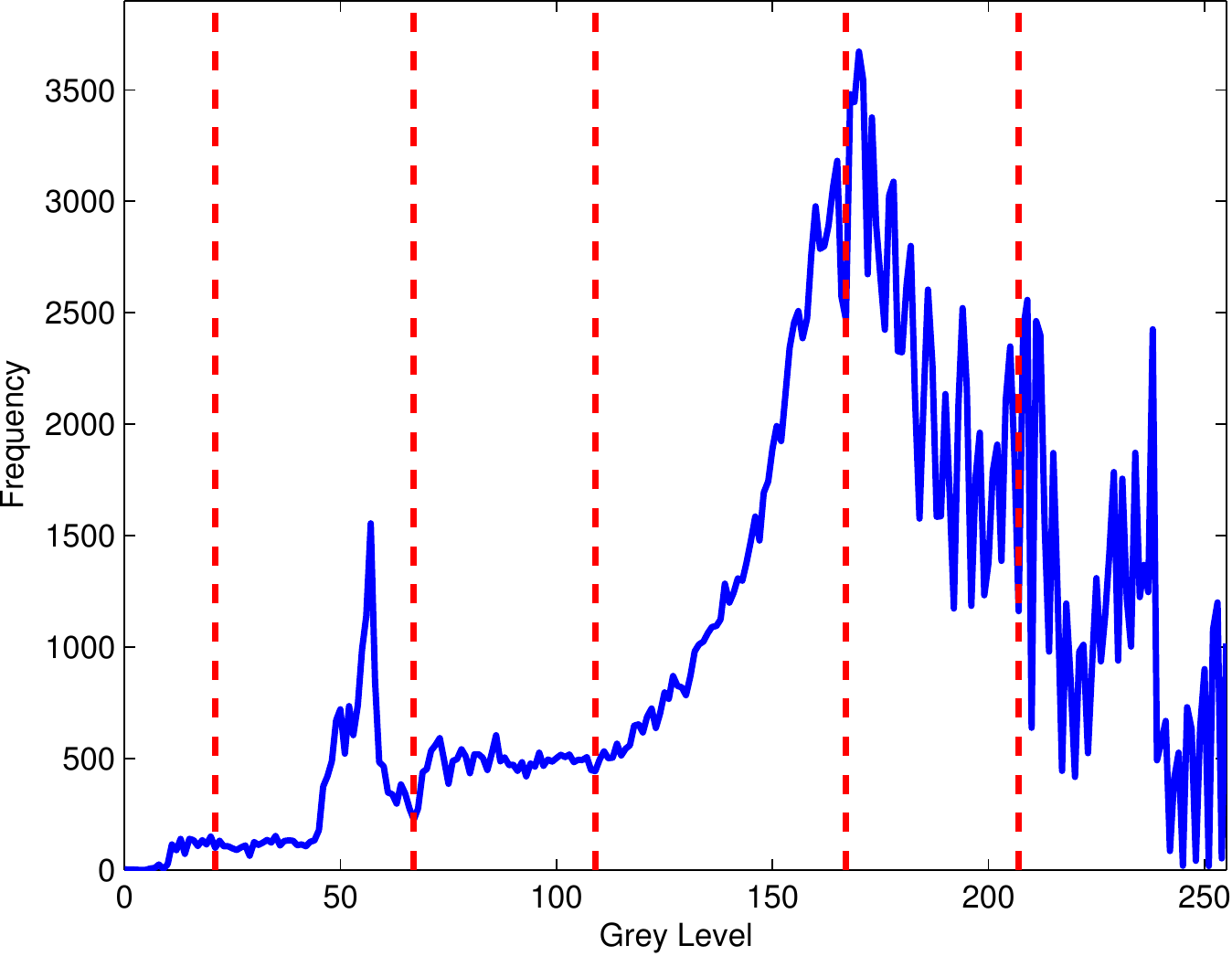}}
	
	\subfloat[threshold=2 \label{fig:Mountainth2}]
	{\includegraphics[scale=0.15]{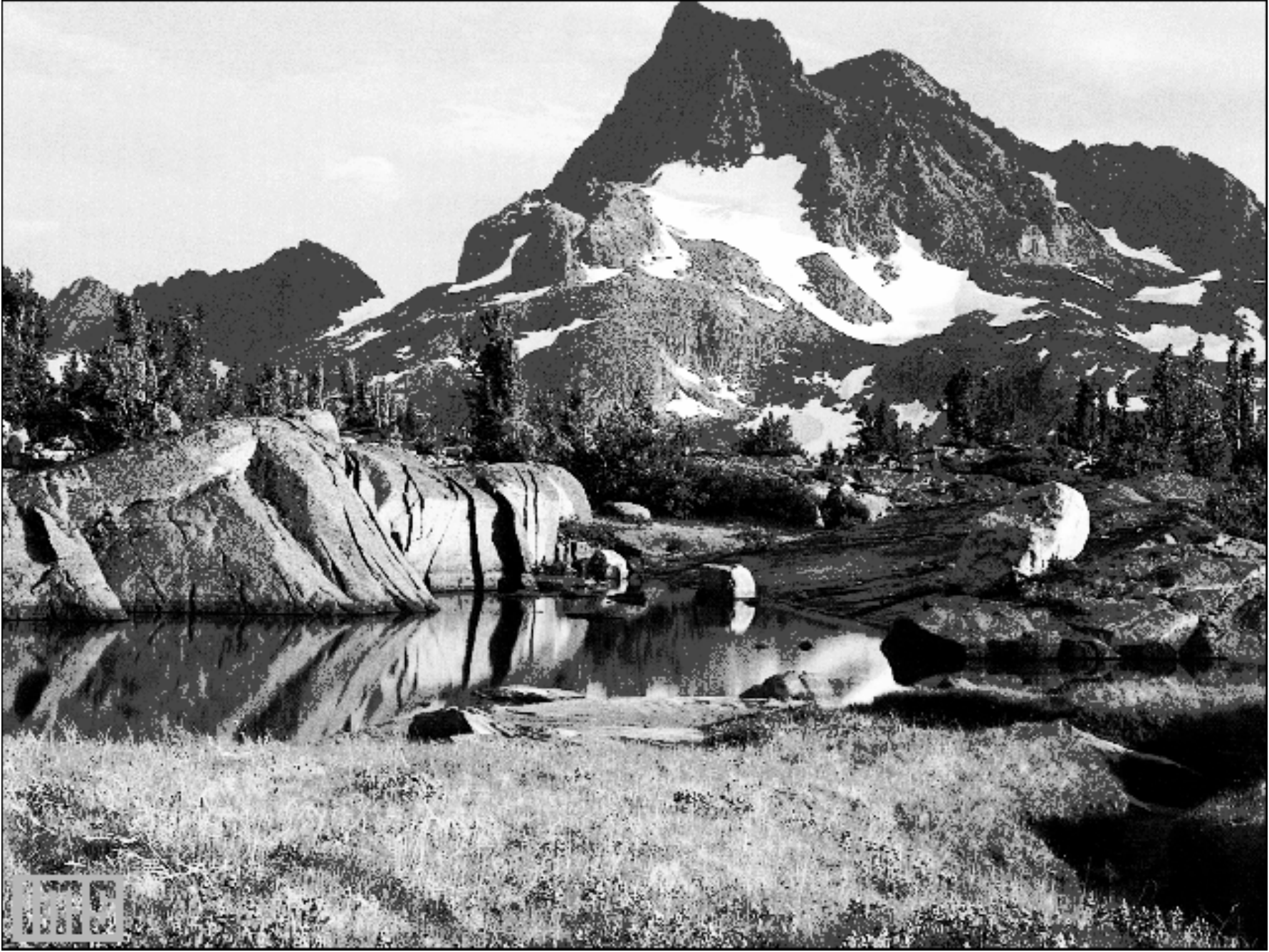}} \:
	\subfloat[threshold=3 \label{fig:Mountainth3}]
	{\includegraphics[scale=0.15]{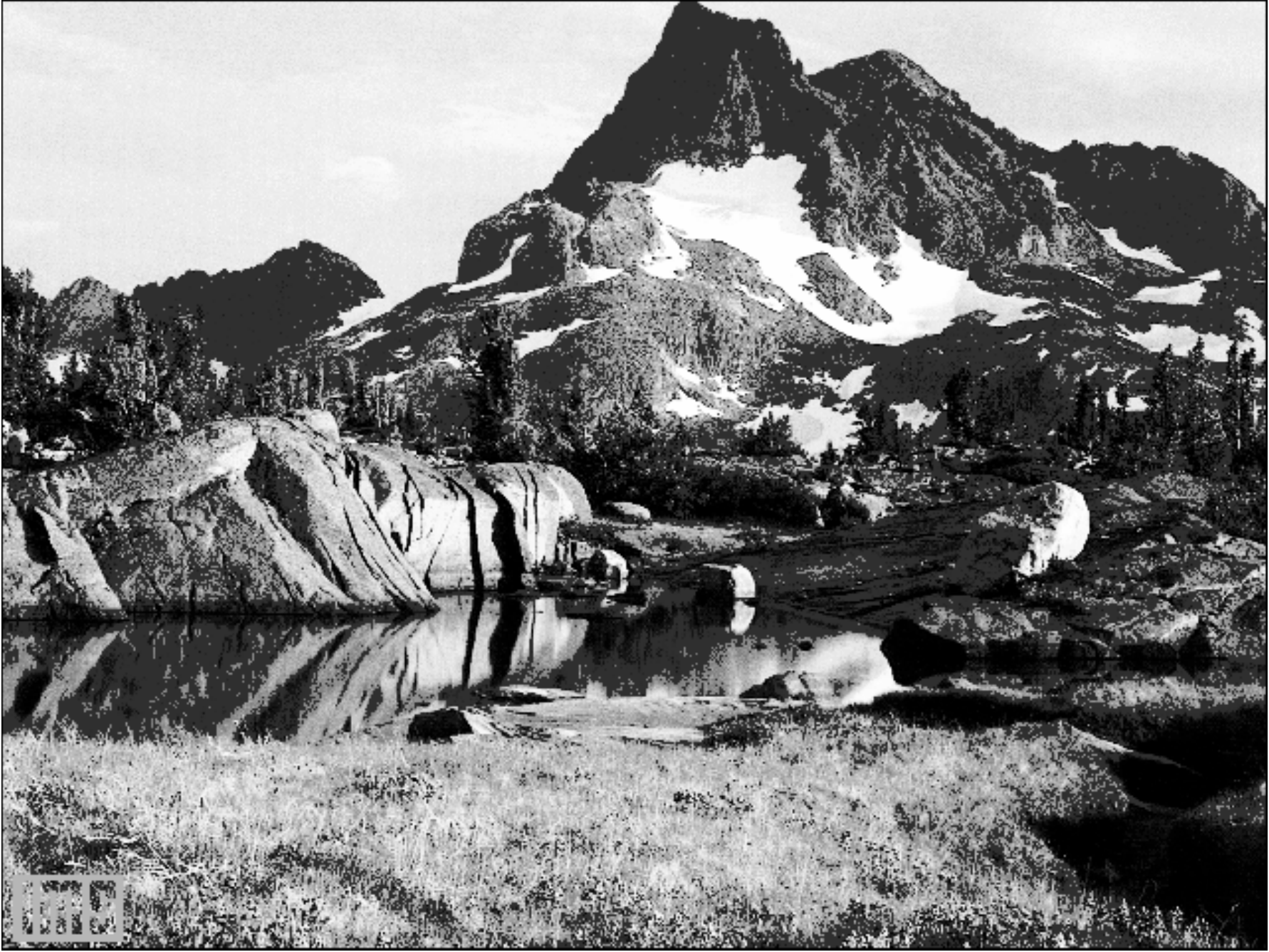}}  \:
	\subfloat[threshold=4 \label{fig:Mountainth4}]
	{\includegraphics[scale=0.15]{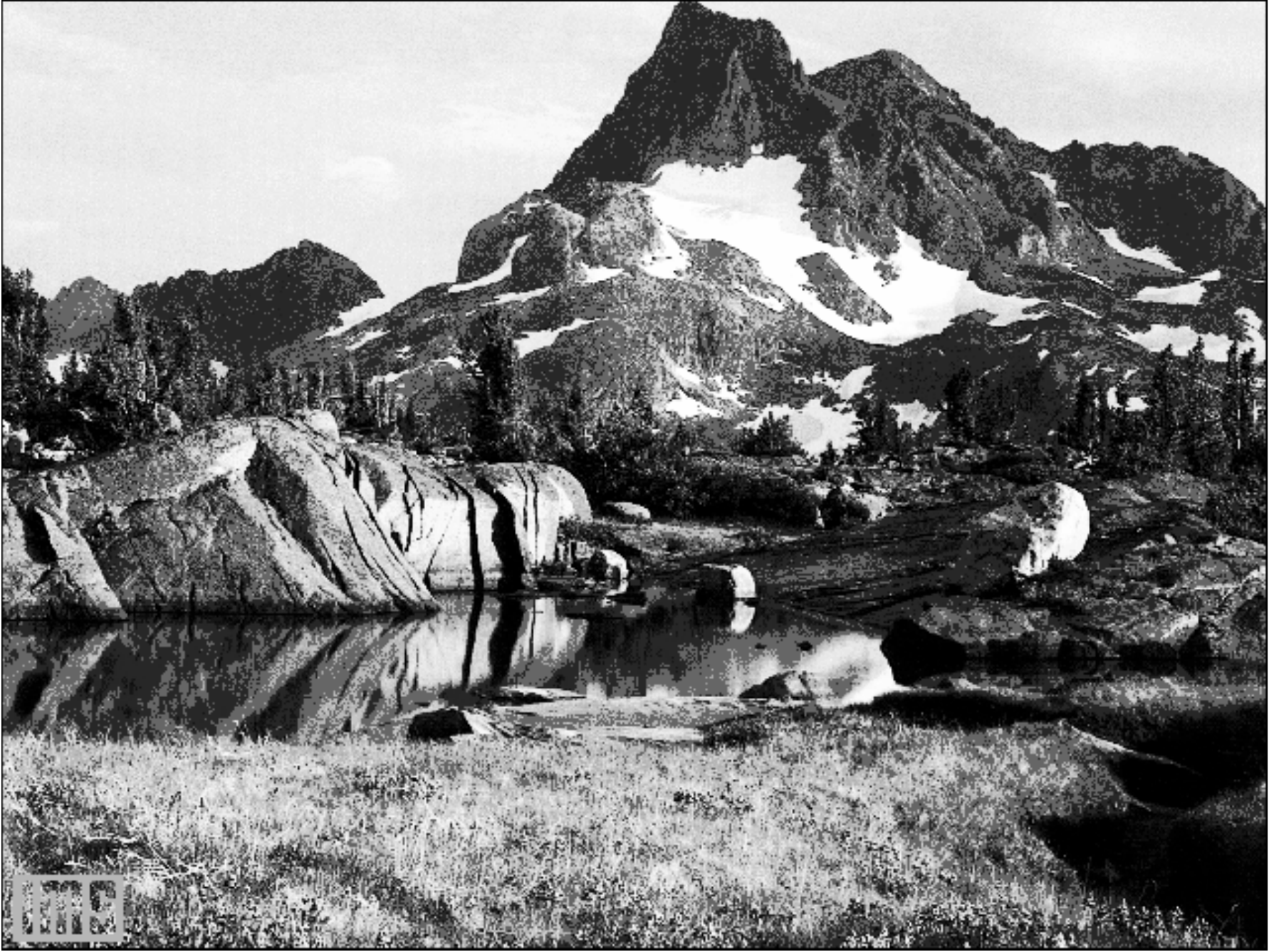}} \:
	\subfloat[threshold=5 \label{fig:Mountainth5}]
	{\includegraphics[scale=0.15]{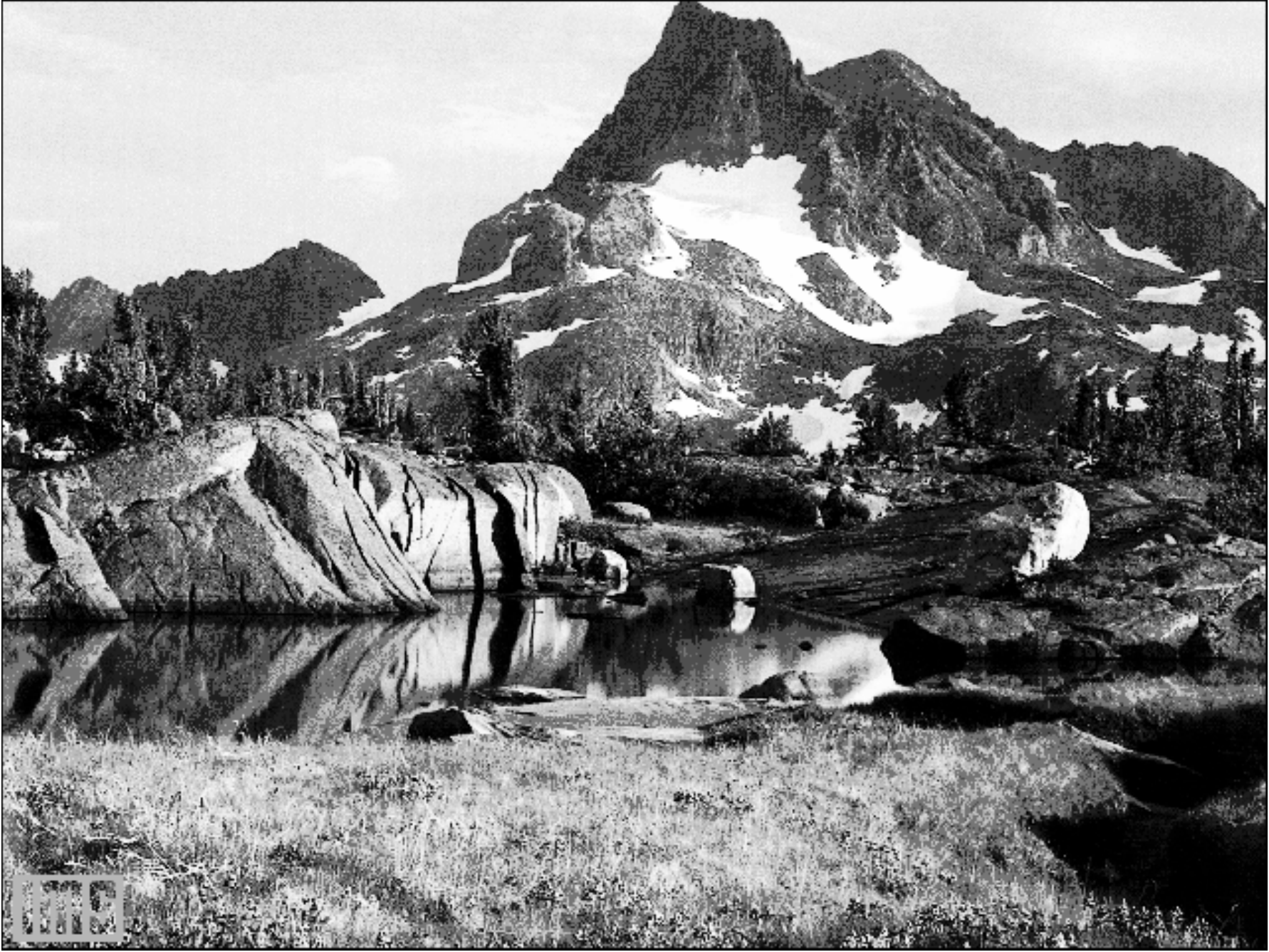}}
	
	\subfloat[threshold=2 \label{fig:Mountainth2Histogram}]
	{\includegraphics[scale=0.203]{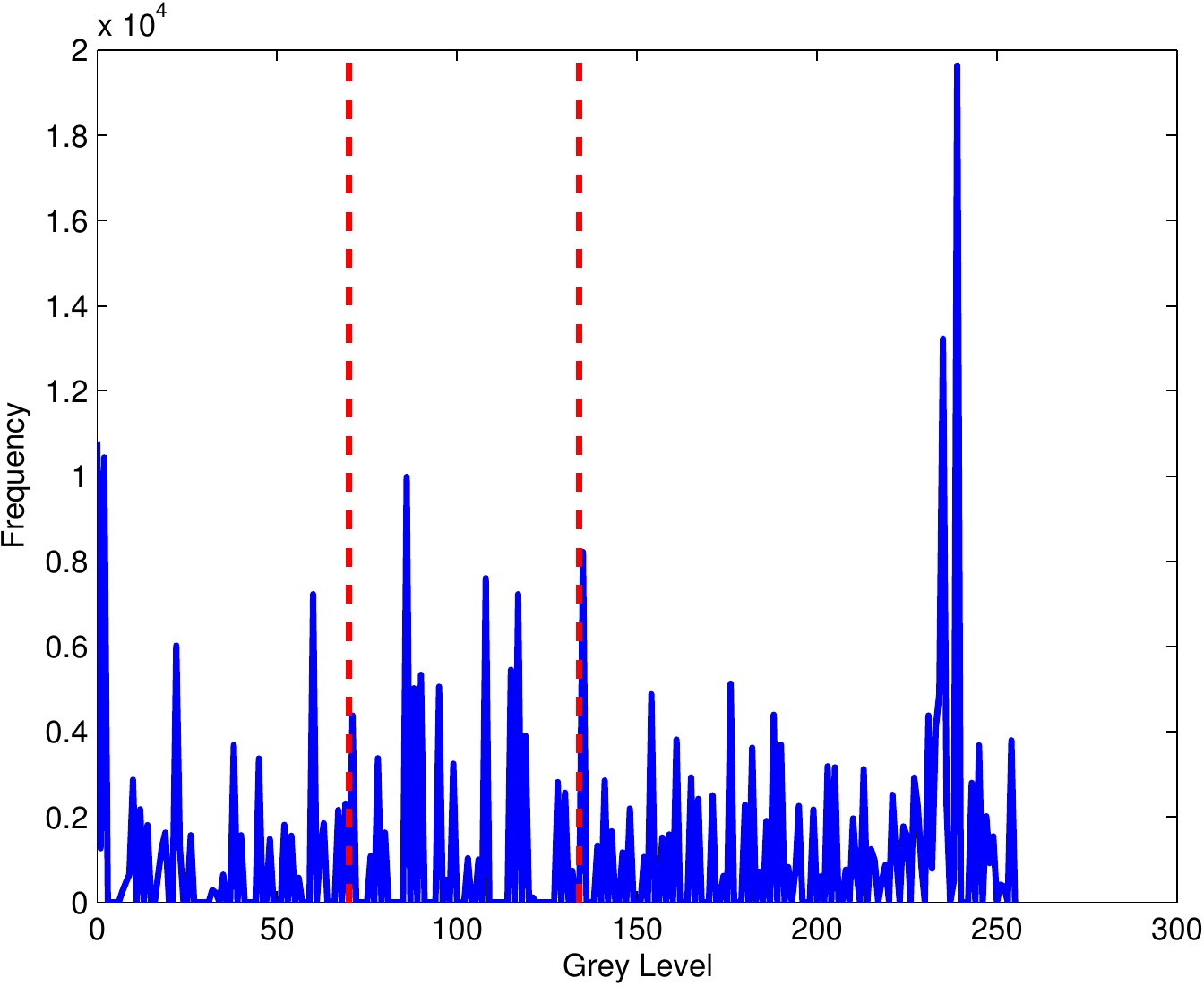}} \:
	\subfloat[threshold=3 \label{fig:Mountainth3Histogram}]
	{\includegraphics[scale=0.203]{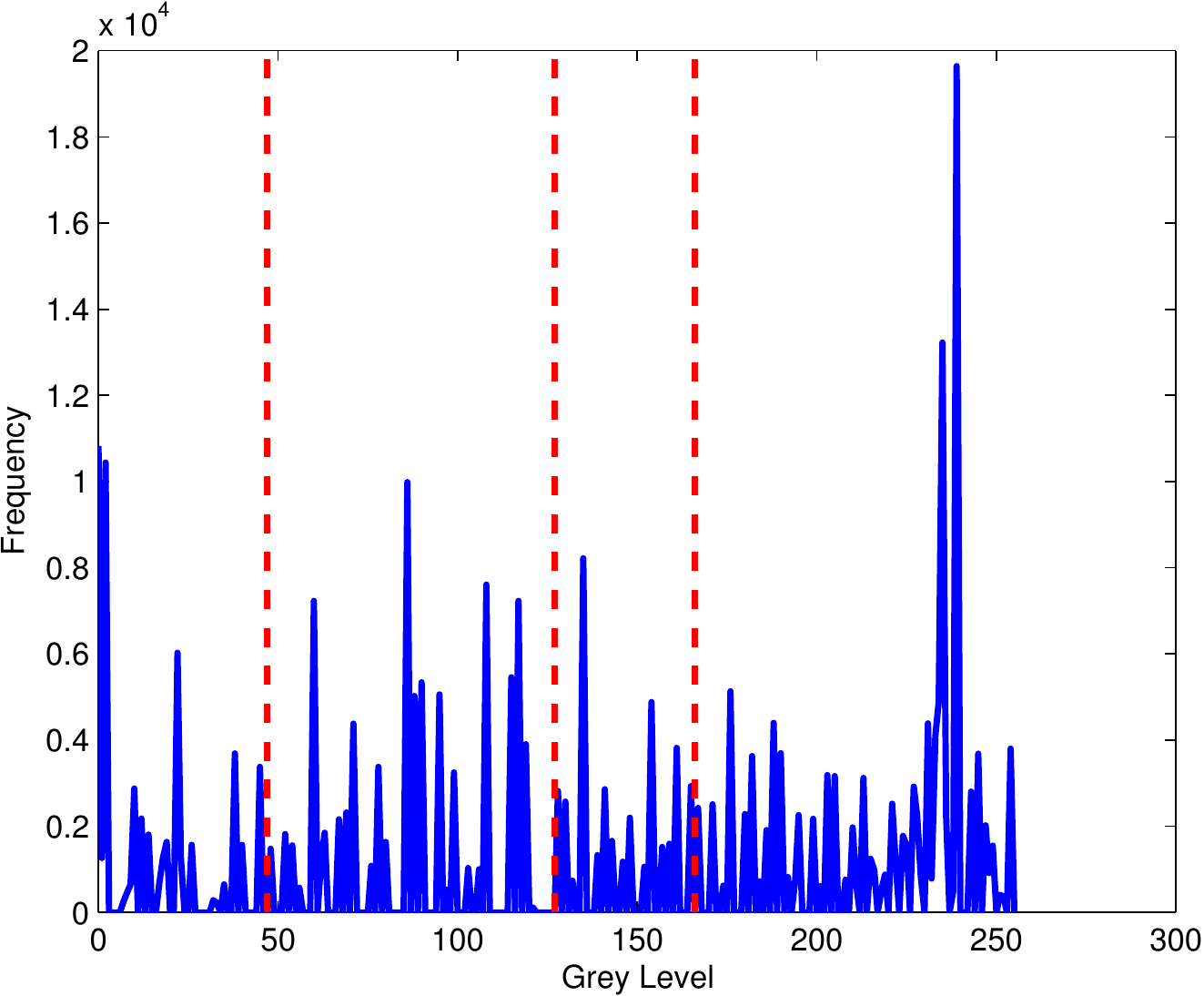}}  \:
	\subfloat[threshold=4 \label{fig:Mountainth4Histogram}]
	{\includegraphics[scale=0.203]{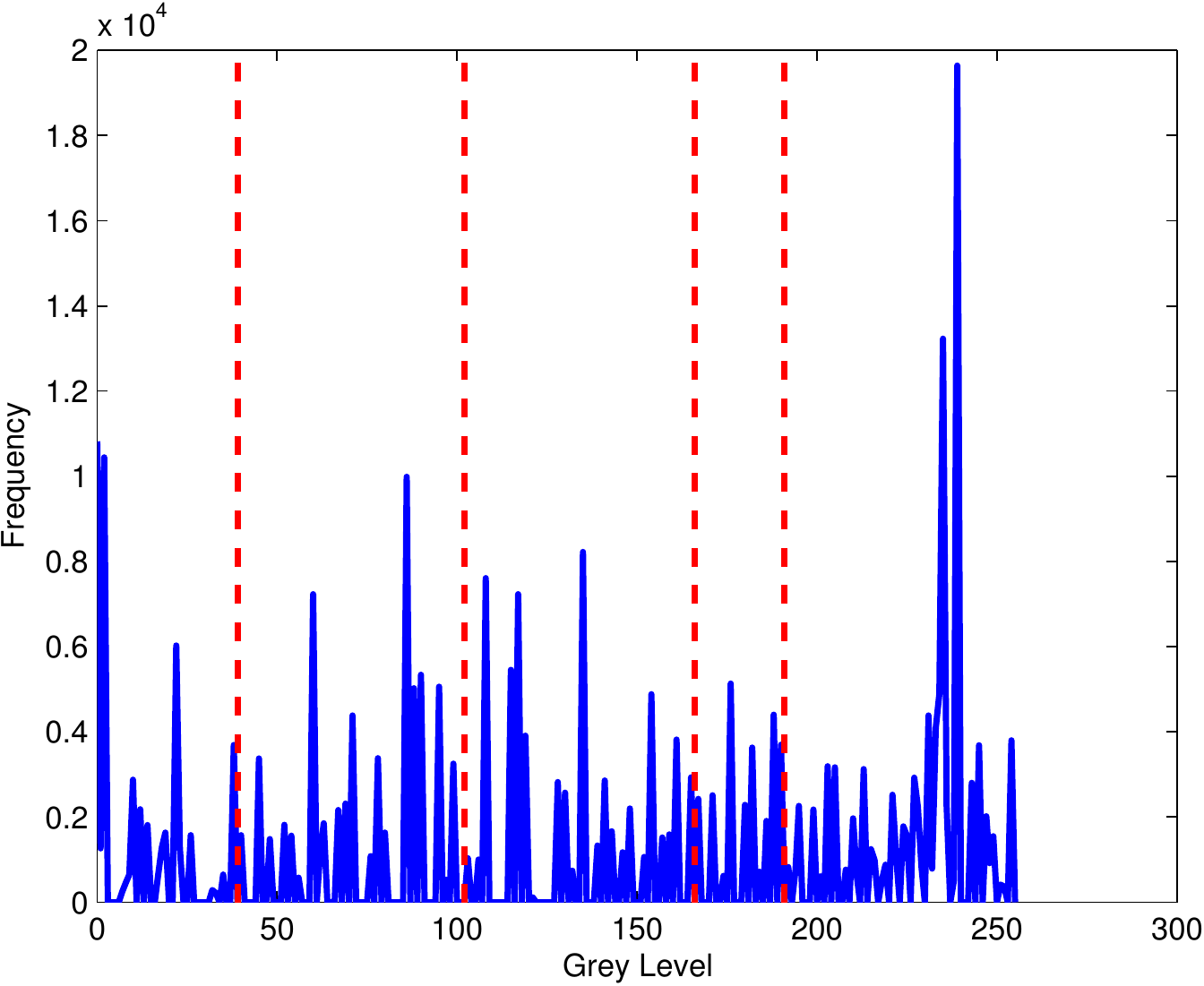}} \:
	\subfloat[threshold=5 \label{fig:Mountainth5Histogram}]
	{\includegraphics[scale=0.203]{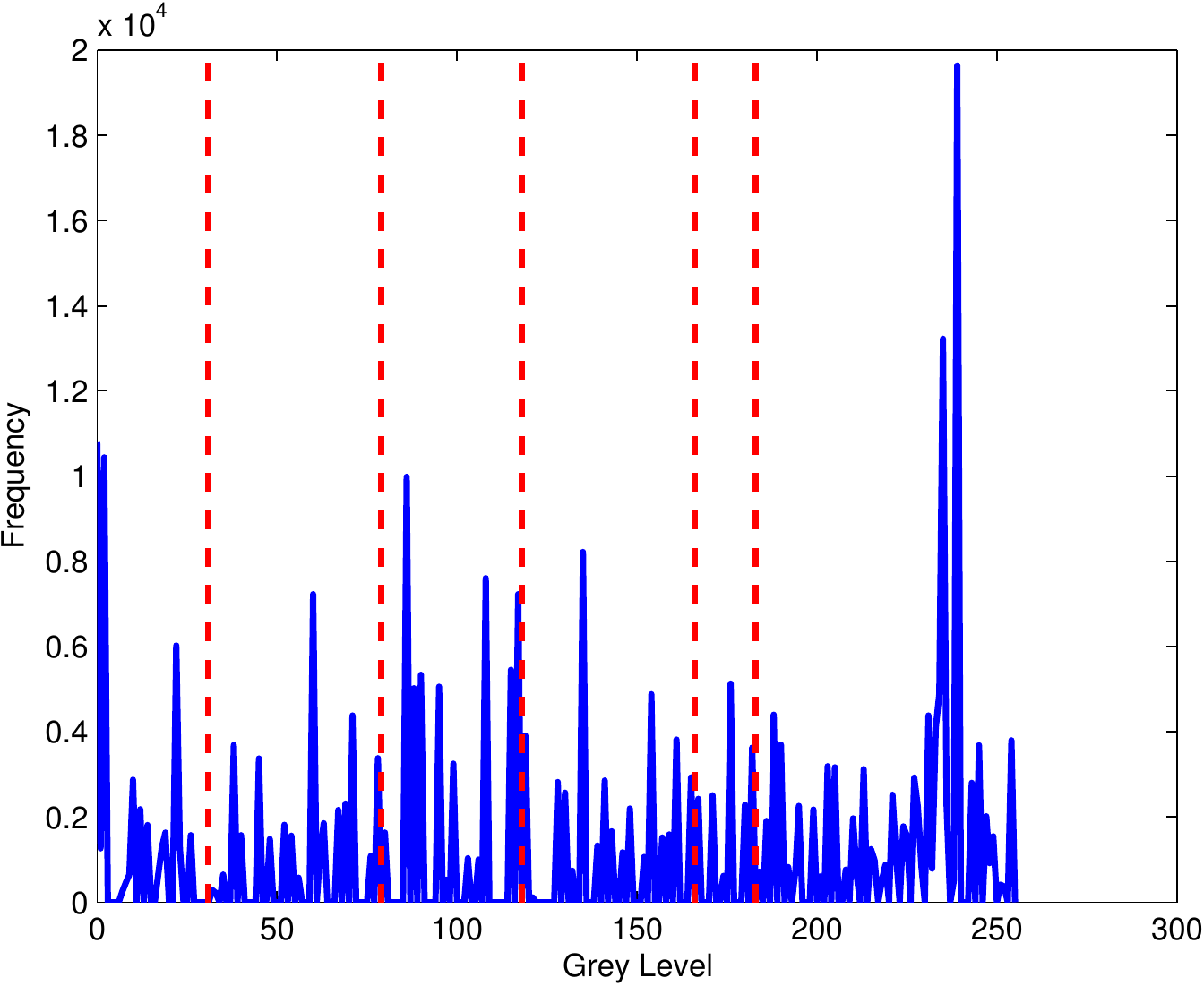}}
	
	\caption{Result obtained using our approach on the benchmark Baboon, Fruits and Mountain. \label{fig:thresholdingContd}}
\end{figure*}

\begin{figure*}[p] 
	\centering{} 	
	\subfloat[threshold=2 \label{fig:Boatth2}]
	{\includegraphics[scale=0.19]{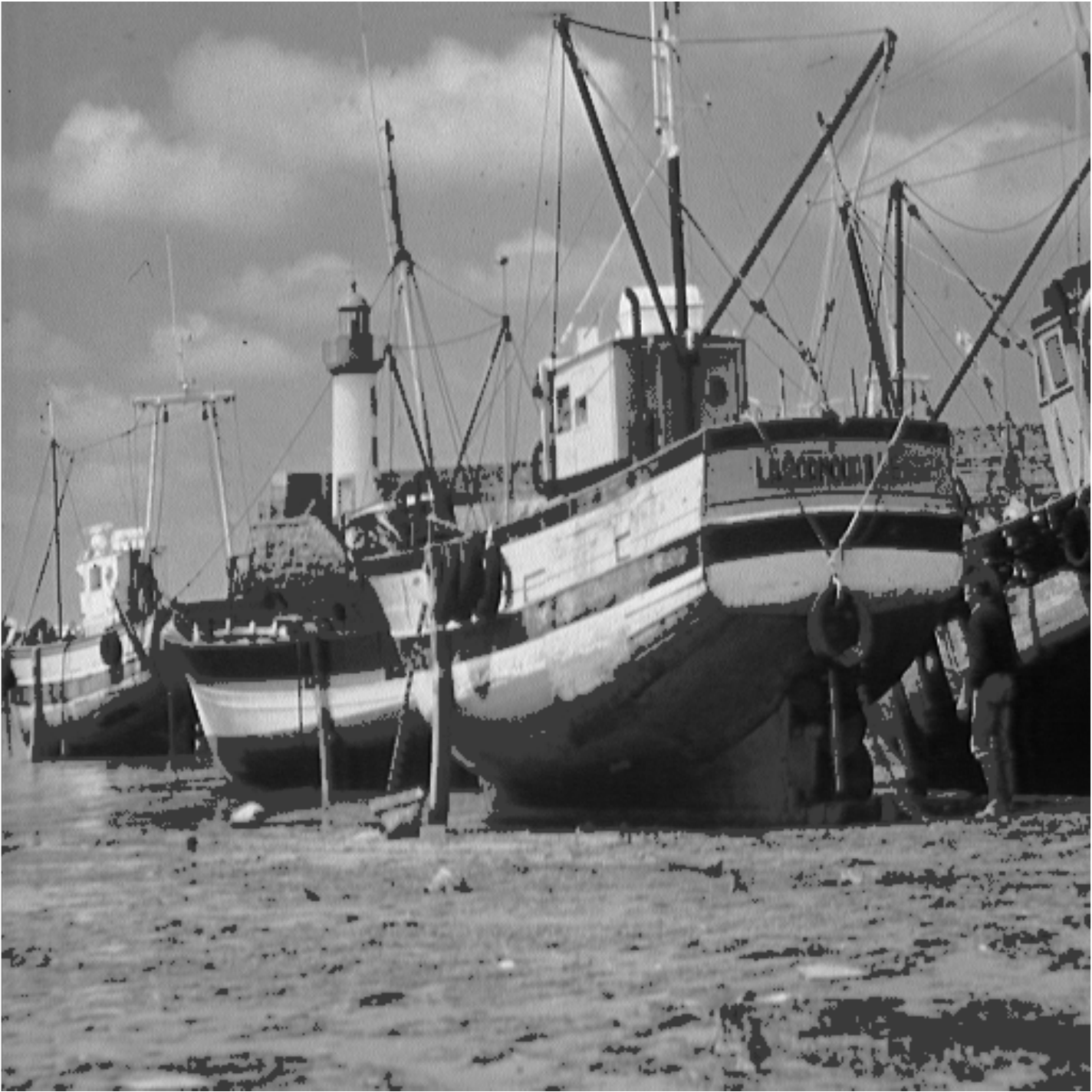}} \:
	\subfloat[threshold=3 \label{fig:Boatth3}]
	{\includegraphics[scale=0.19]{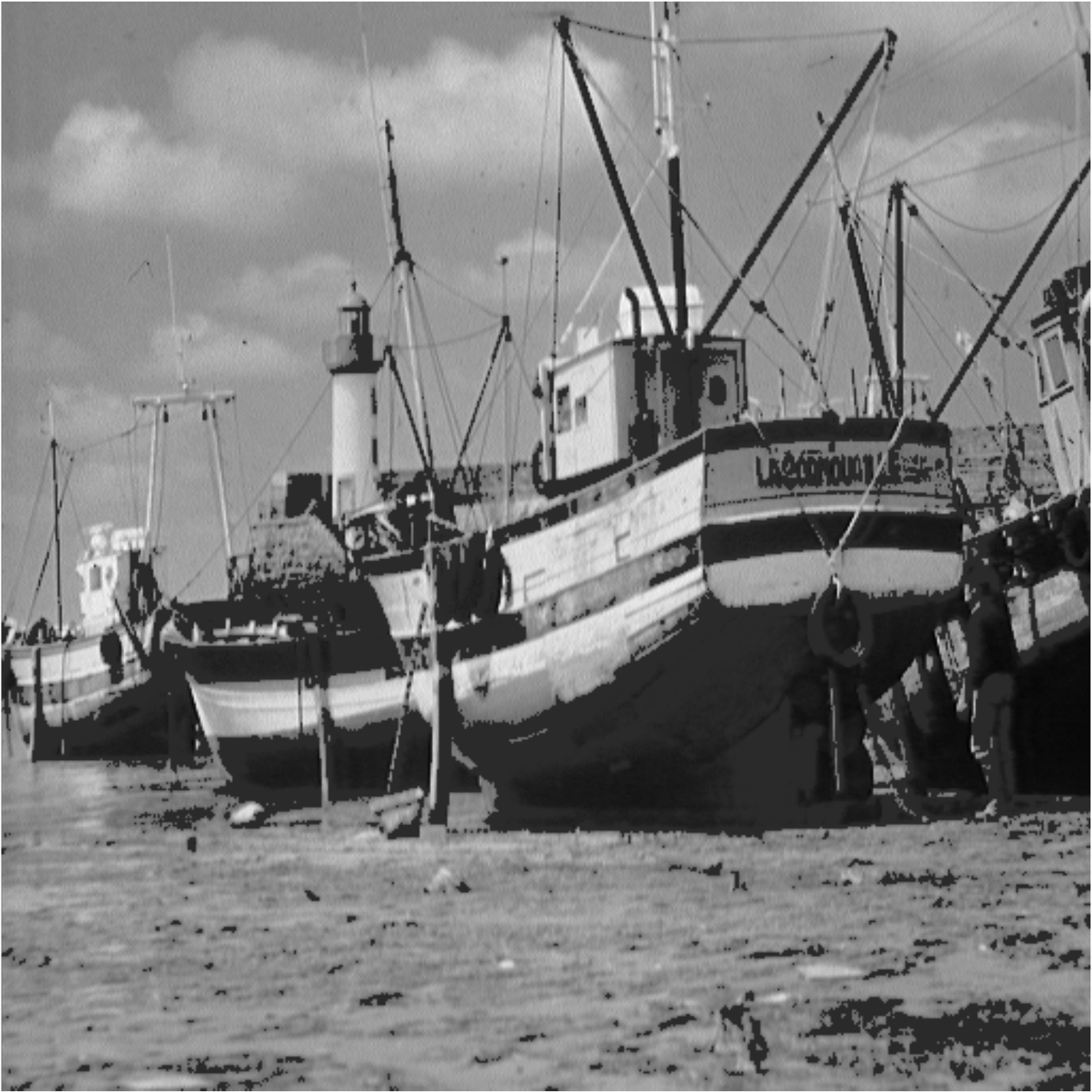}} \:
	\subfloat[threshold=4 \label{fig:Boatth4}]
	{\includegraphics[scale=0.19]{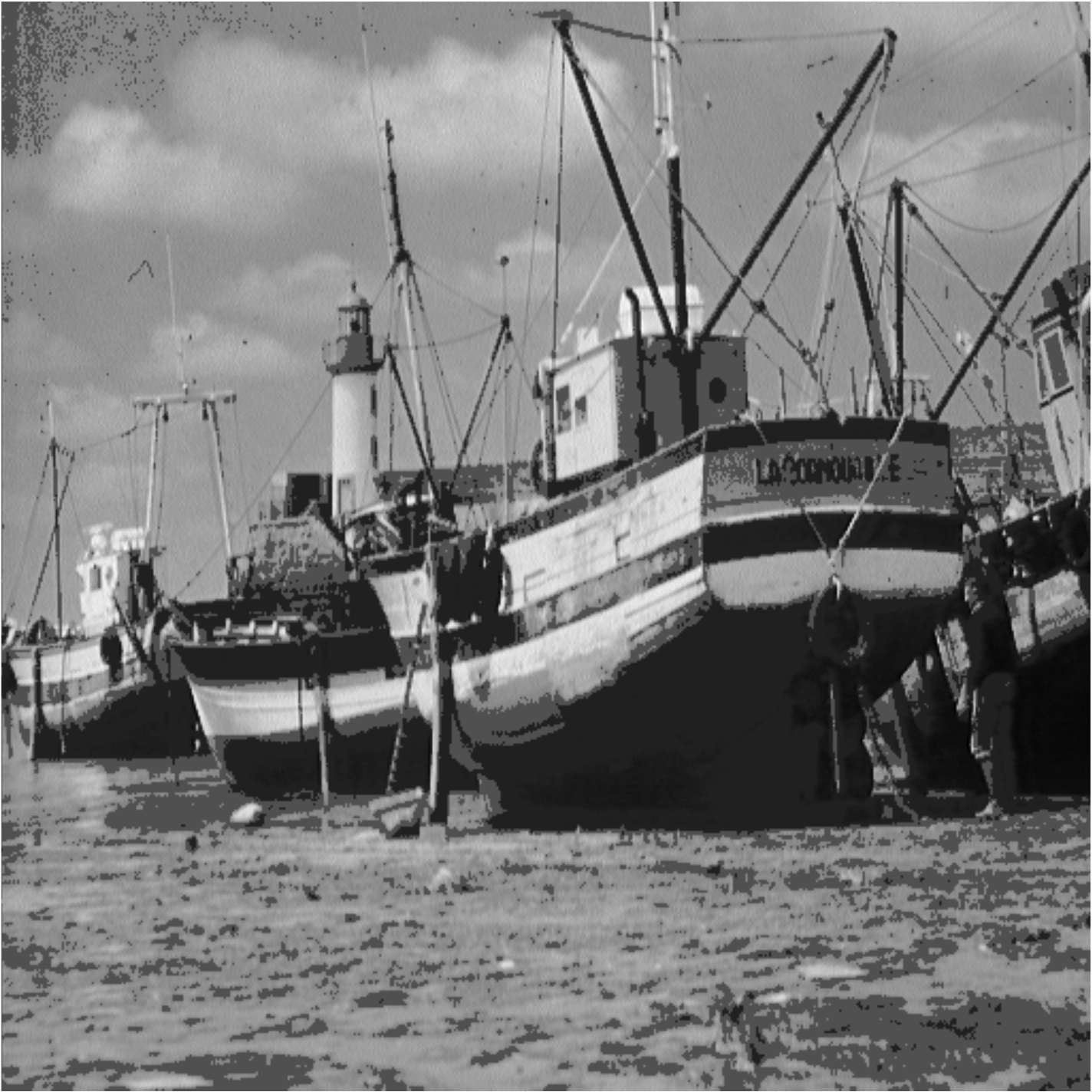}} \:
	\subfloat[threshold=5 \label{fig:Boatth5}]
	{\includegraphics[scale=0.19]{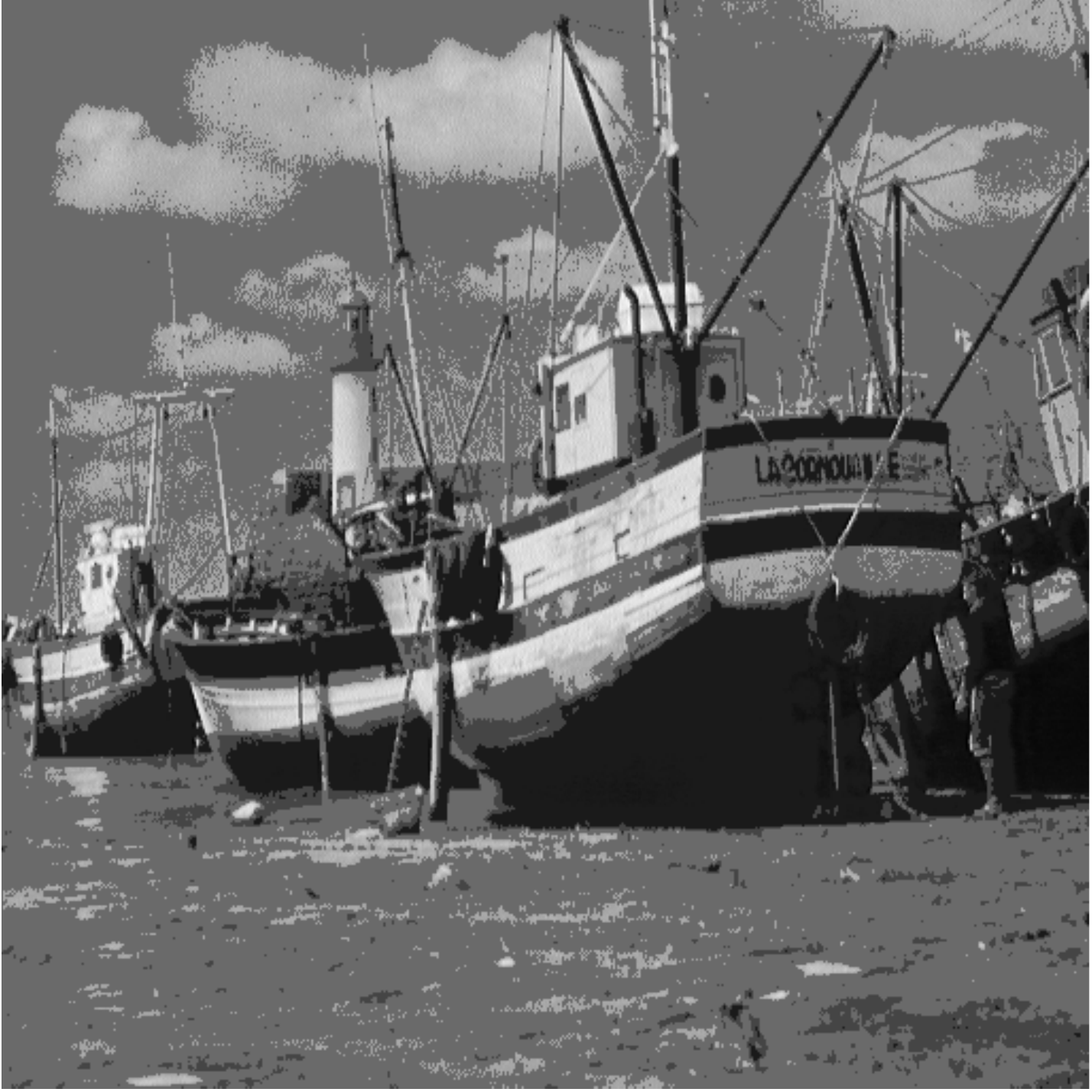}} \:
	
	\subfloat[threshold=2 \label{fig:Boatth2Histogram}]
	{\includegraphics[scale=0.2]{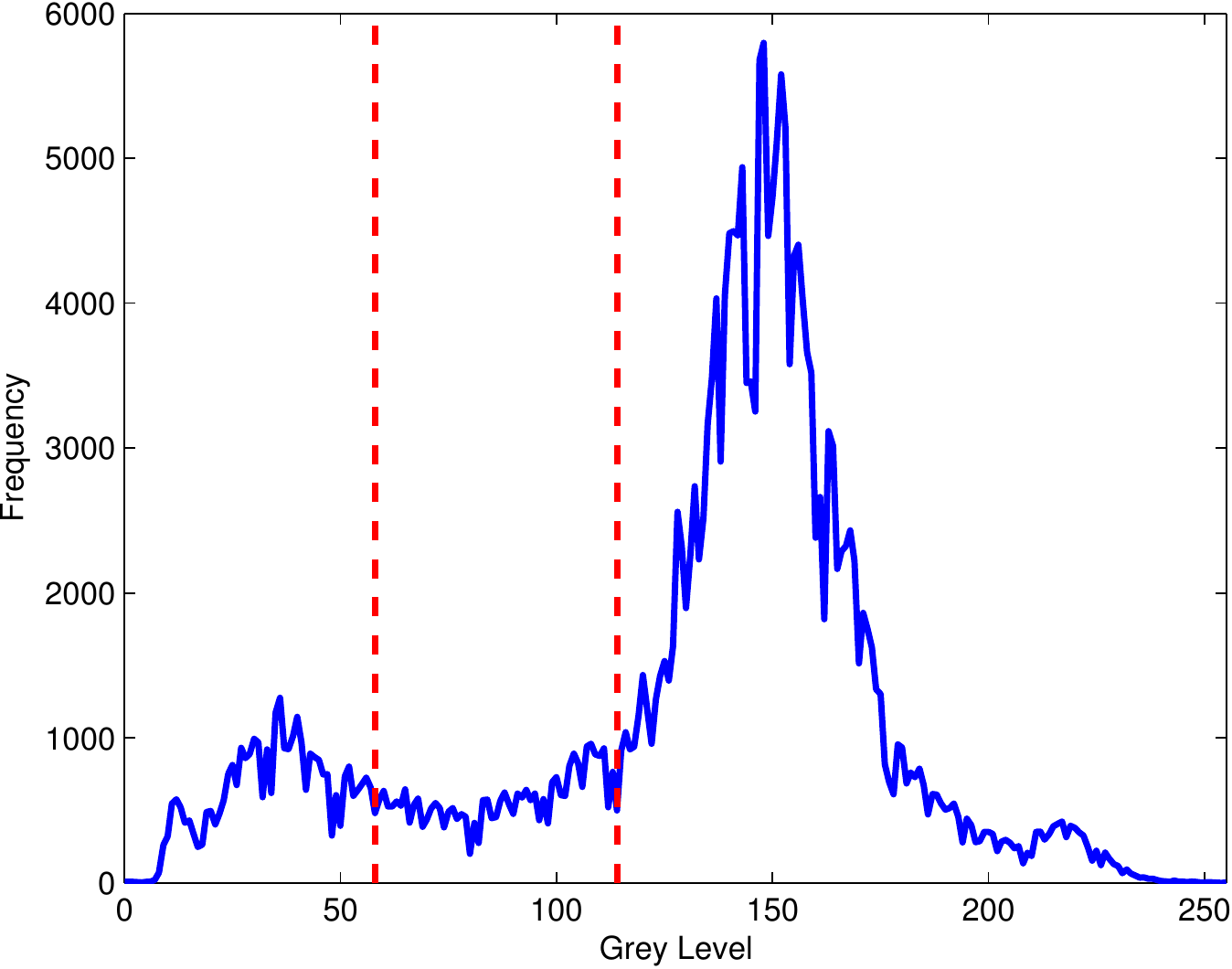}} \:
	\subfloat[threshold=3 \label{fig:Boatth3Histogram}]
	{\includegraphics[scale=0.2]{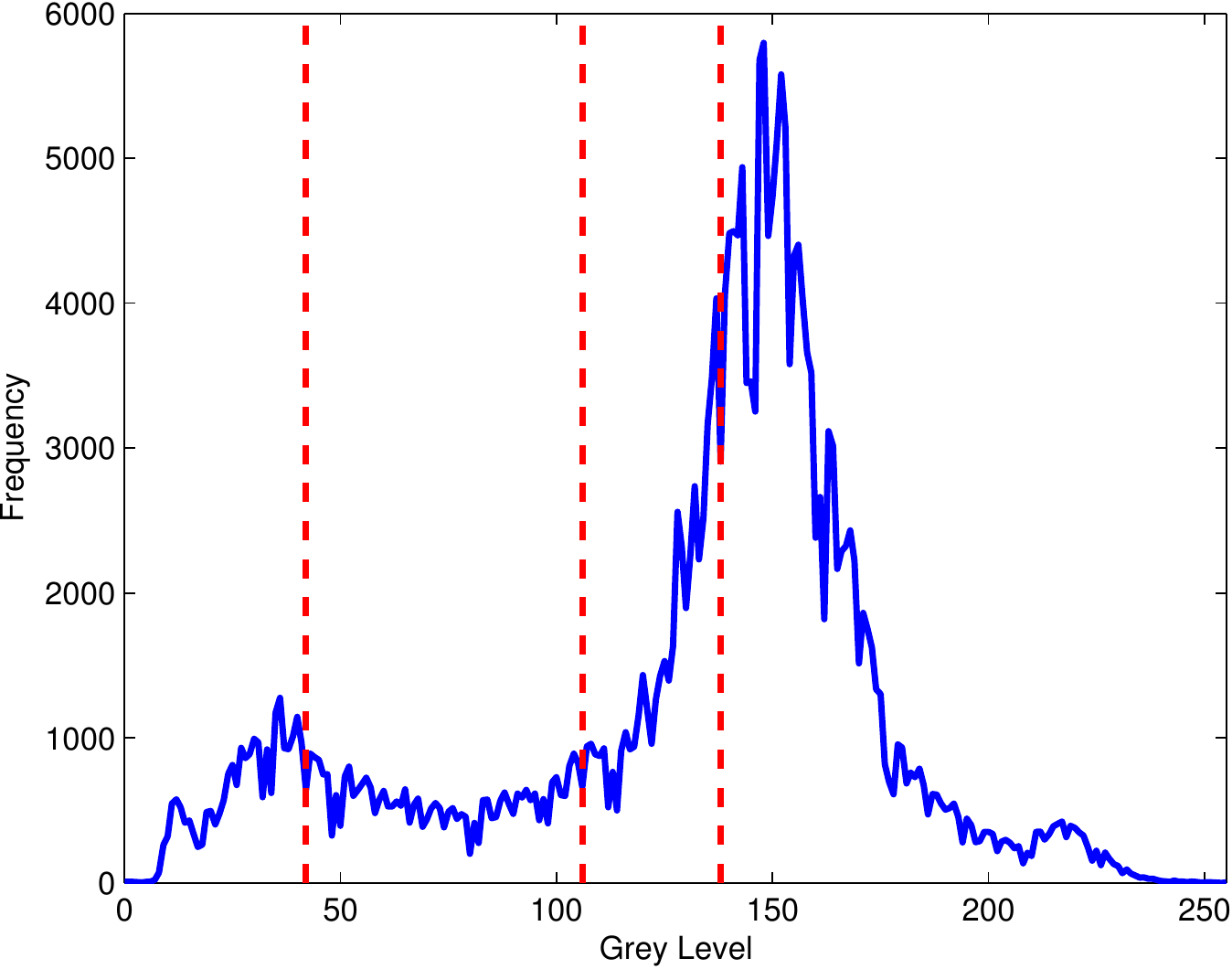}} \:
	\subfloat[threshold=4 \label{fig:Boatth4Histogram}]
	{\includegraphics[scale=0.2]{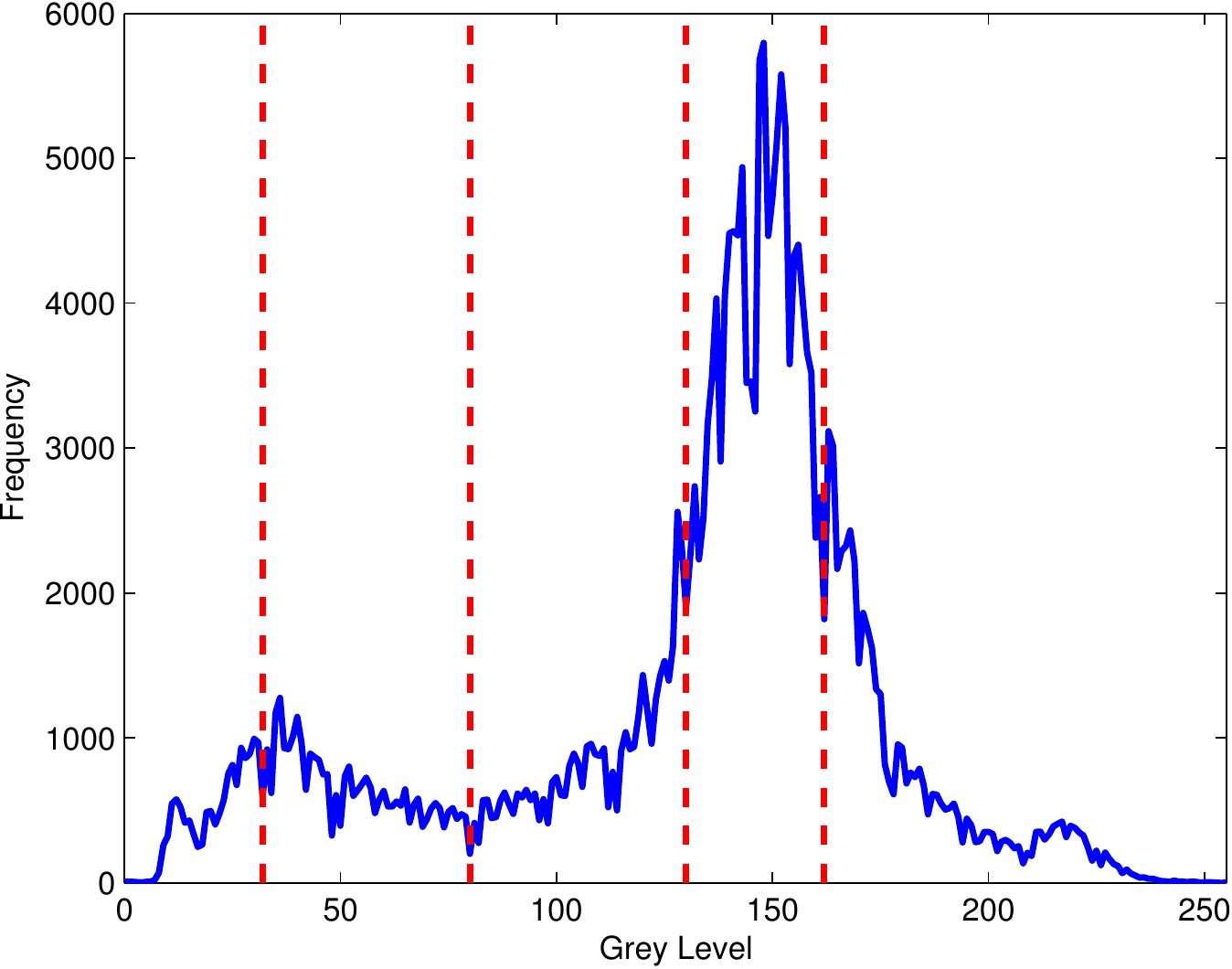}} \:
	\subfloat[threshold=5 \label{fig:Boatth5Histogram}]
	{\includegraphics[scale=0.2]{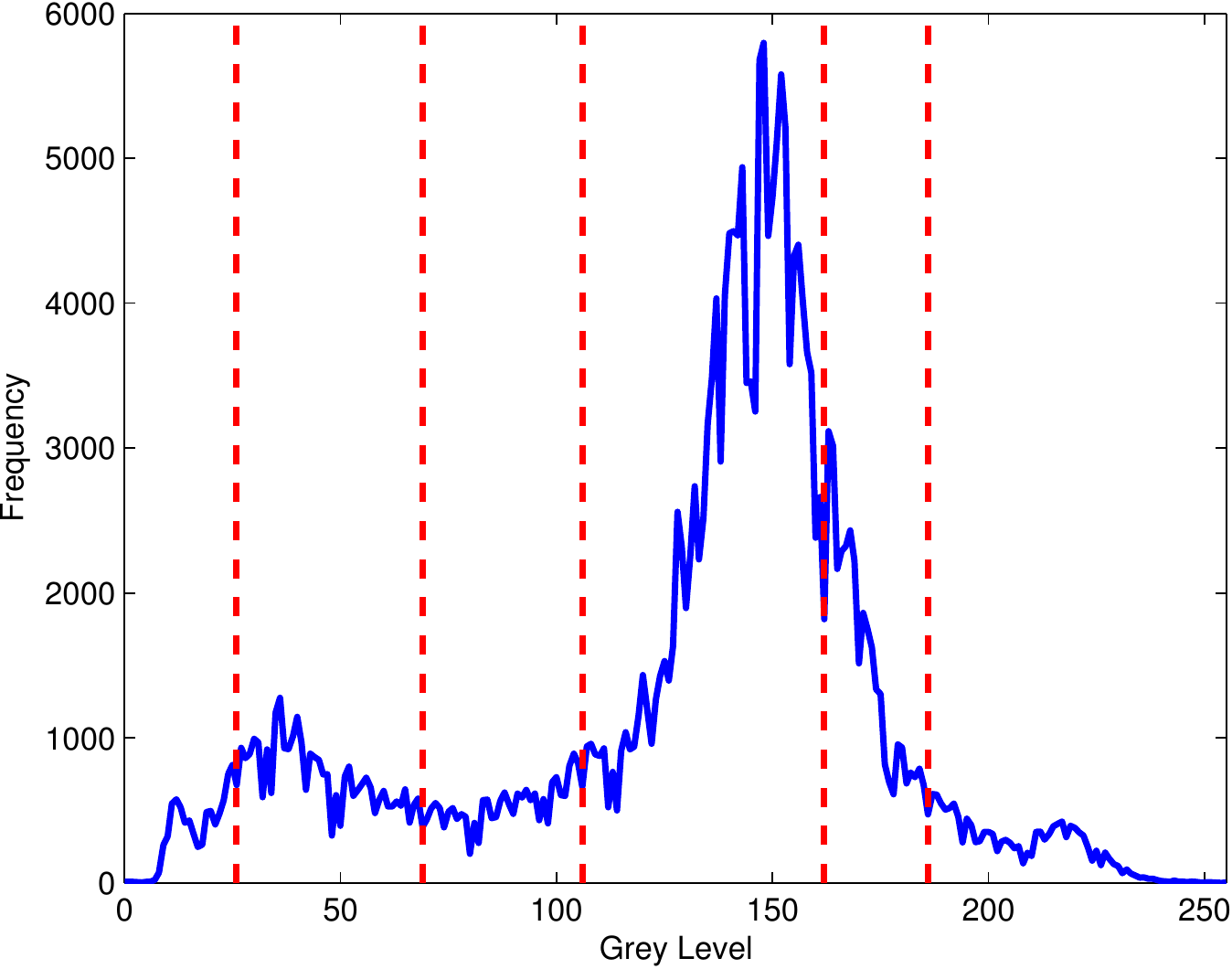}}
	
	\subfloat[threshold=2 \label{fig:Finger101th2}]
	{\includegraphics[scale=0.32]{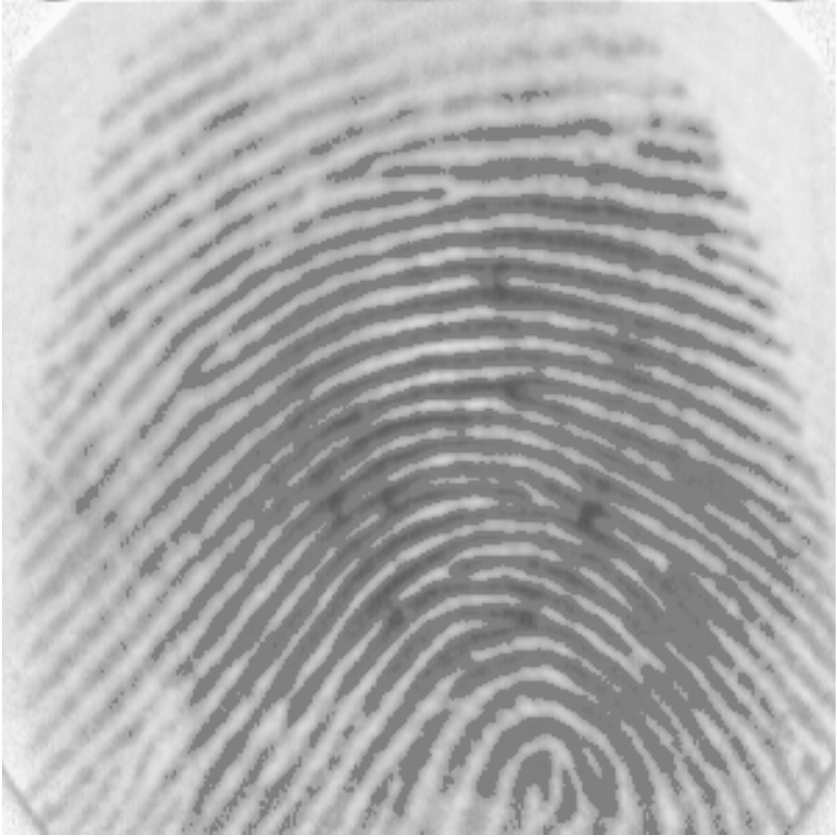}} \:
	\subfloat[threshold=3 \label{fig:Finger101th3}]
	{\includegraphics[scale=0.32]{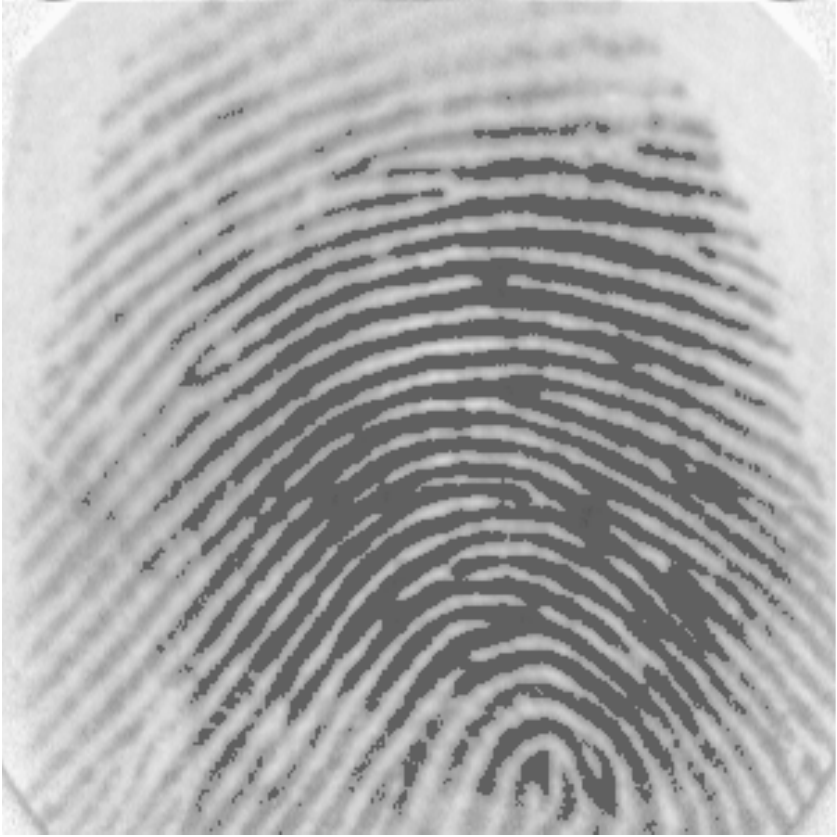}} \:
	\subfloat[threshold=4 \label{fig:Finger101th4}]
	{\includegraphics[scale=0.32]{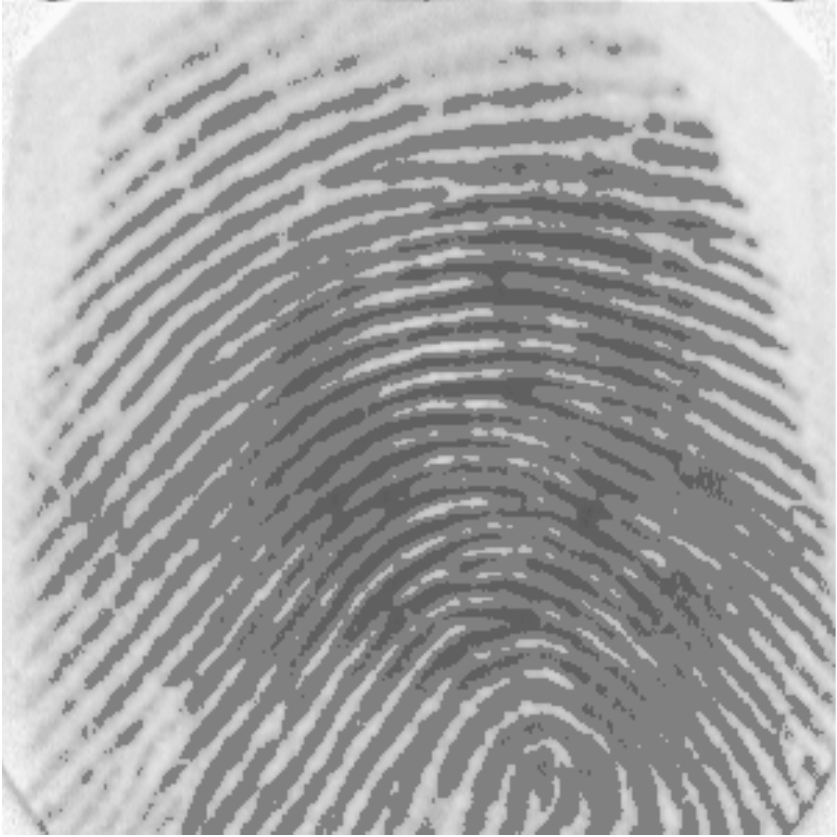}} \:
	\subfloat[threshold=5 \label{fig:Finger101th5}]
	{\includegraphics[scale=0.32]{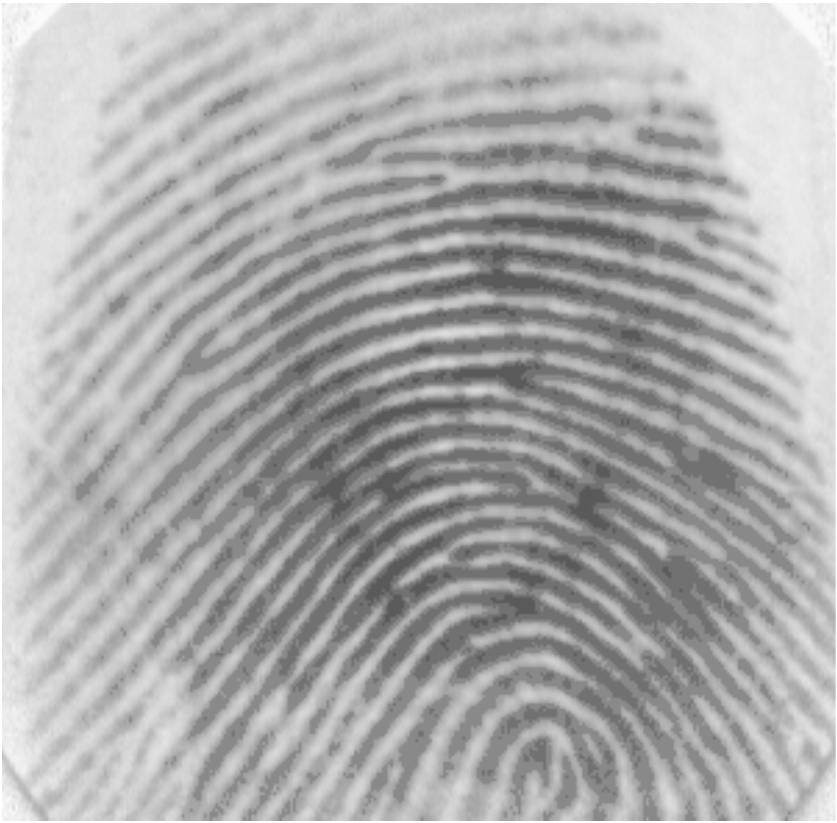}} \:
	
	\subfloat[threshold=2 \label{fig:Finger101th2Histogram}]
	{\includegraphics[scale=0.2]{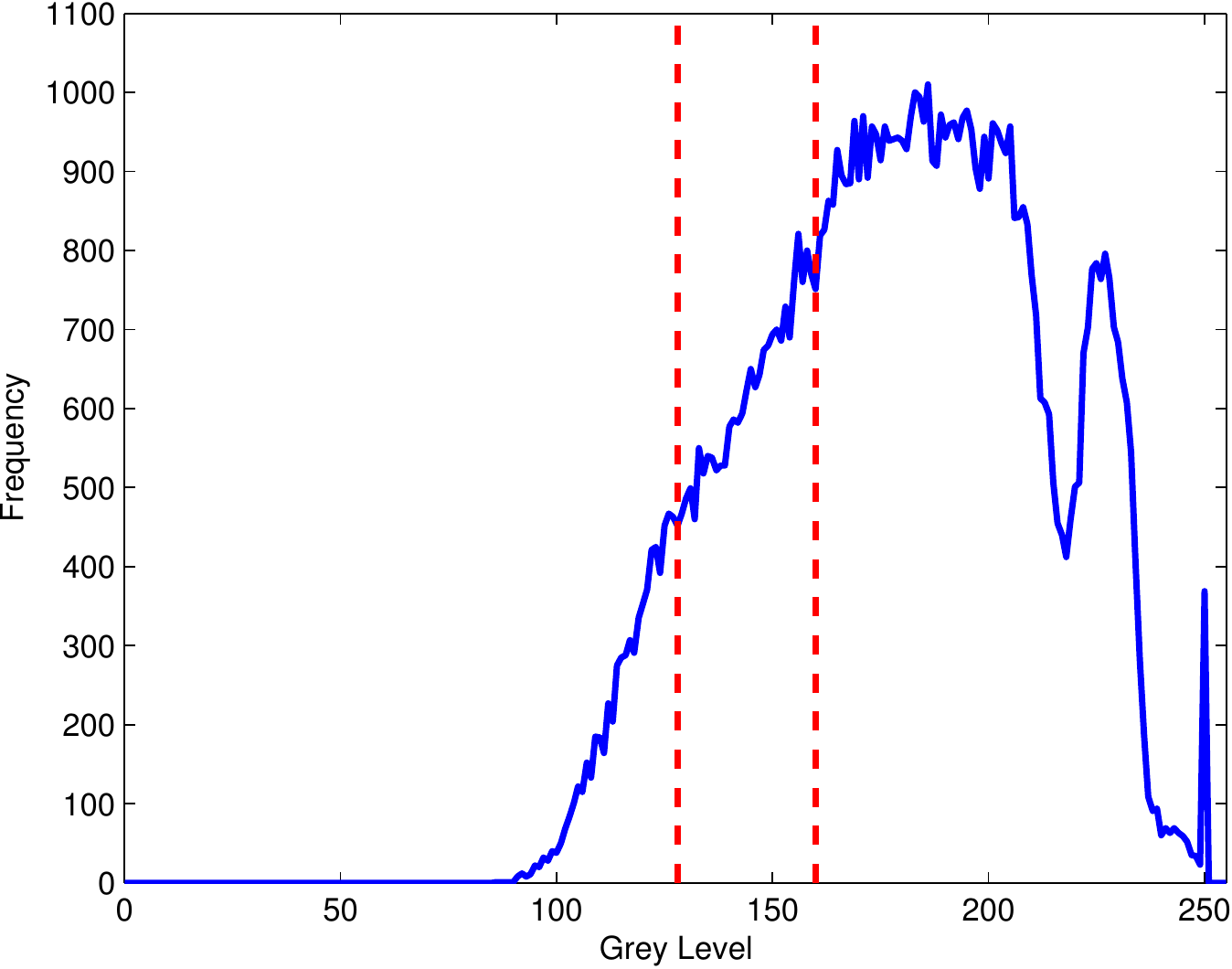}} \:
	\subfloat[threshold=3 \label{fig:Finger101th3Histogram}]
	{\includegraphics[scale=0.2]{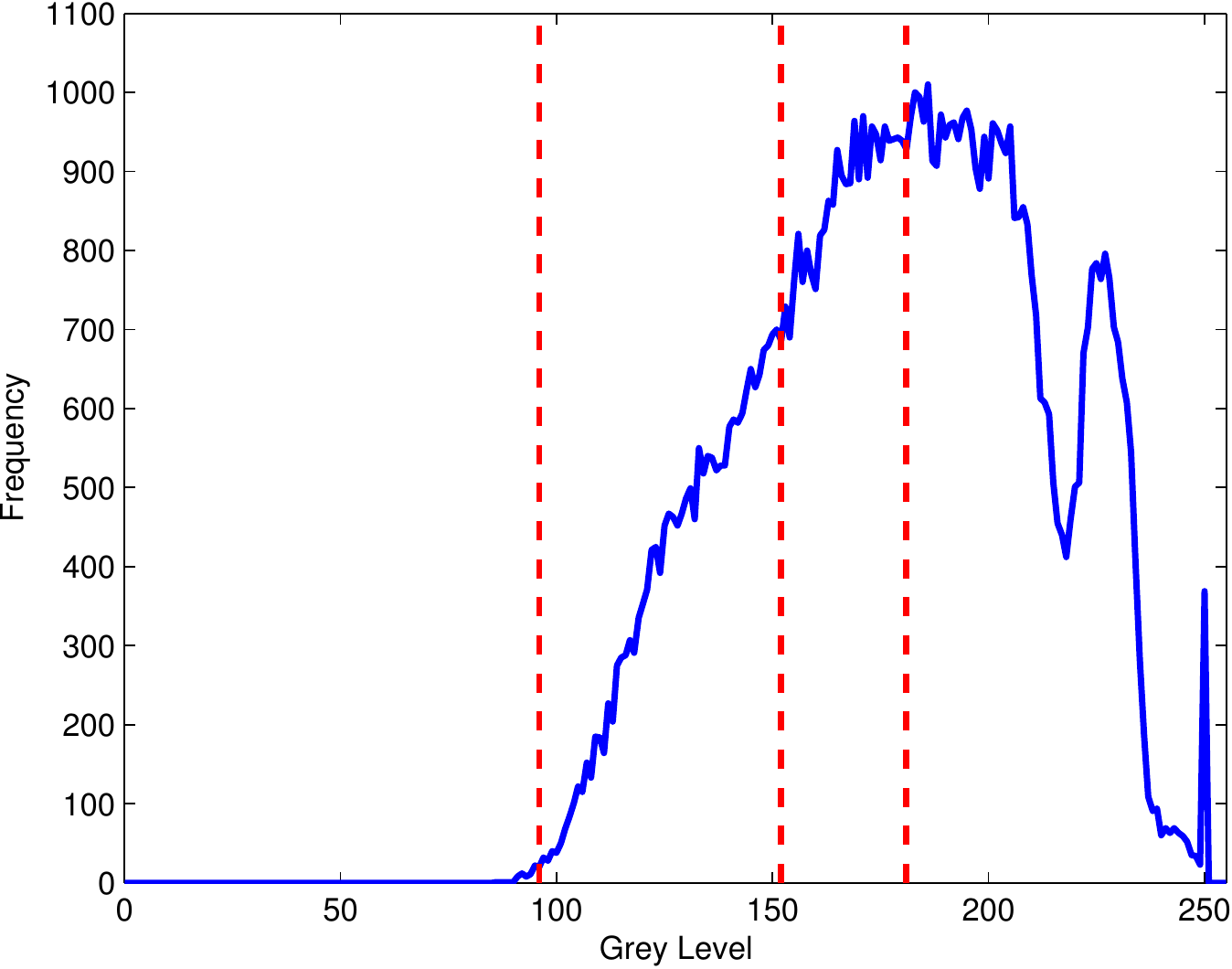}} \:
	\subfloat[threshold=4 \label{fig:Finger101th4Histogram}]
	{\includegraphics[scale=0.2]{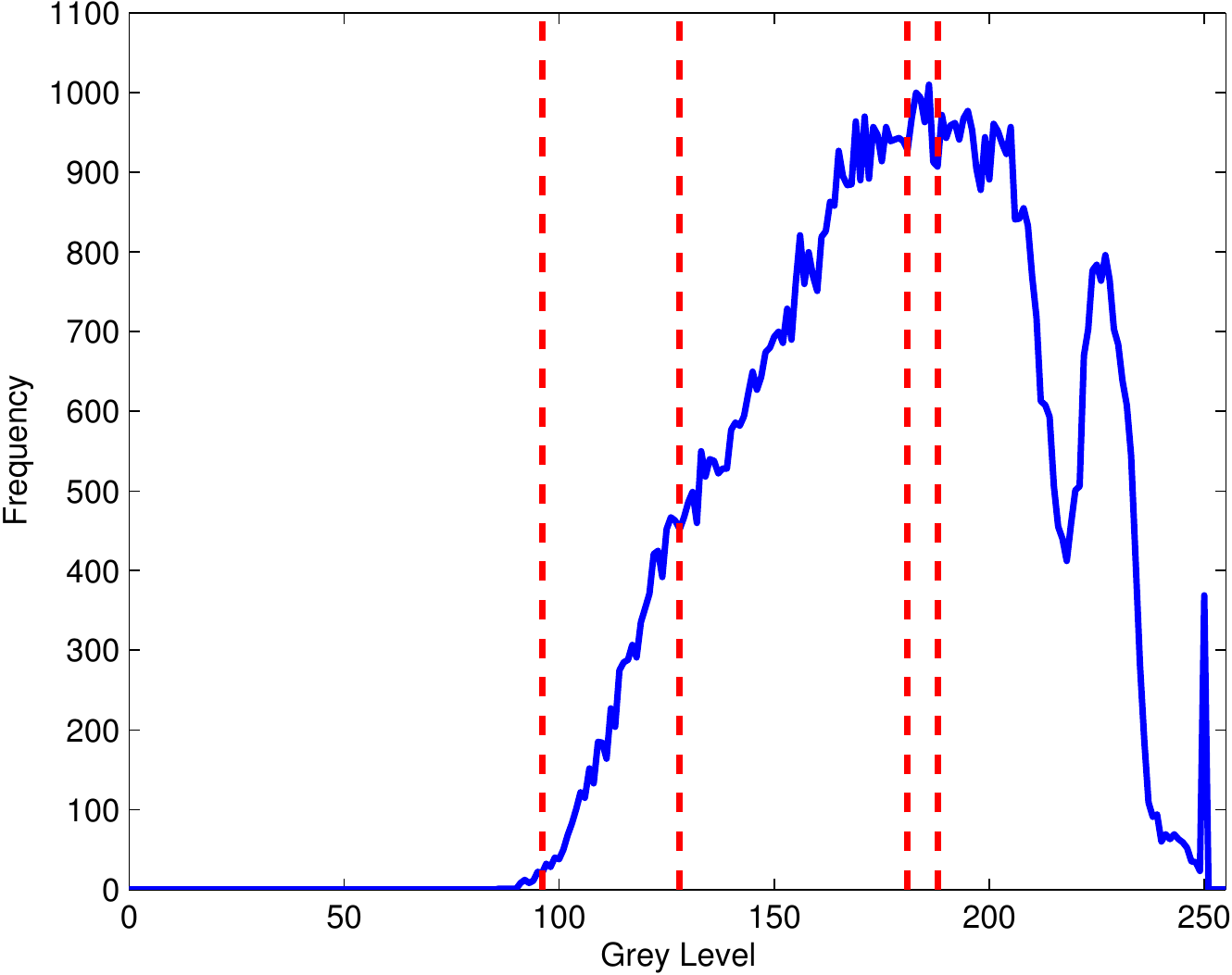}} \:
	\subfloat[threshold=5 \label{fig:Finger101th5Histogram}]
	{\includegraphics[scale=0.2]{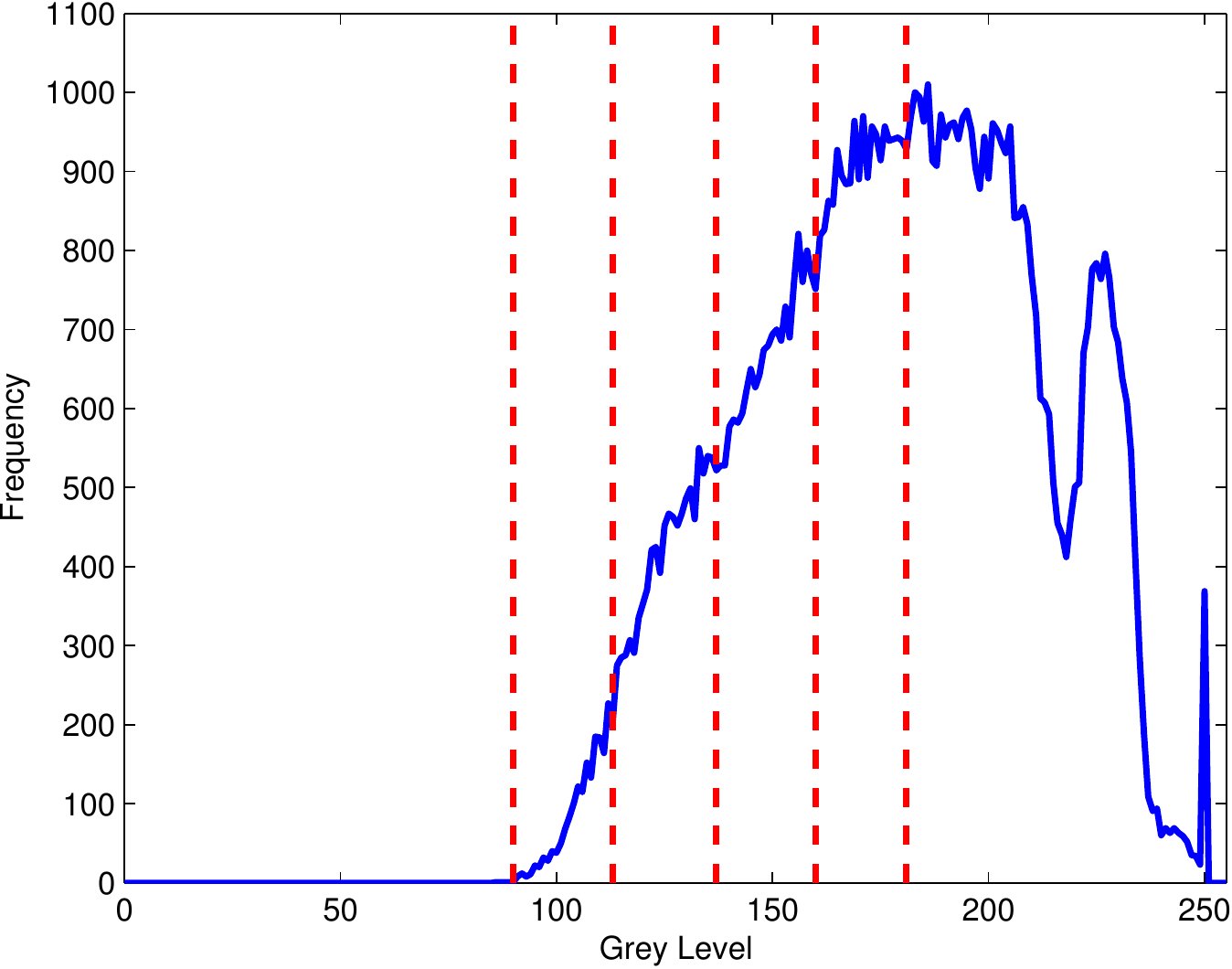}}
	
	\subfloat[threshold=2 \label{fig:Zeldath2}]
	{\includegraphics[scale=0.2]{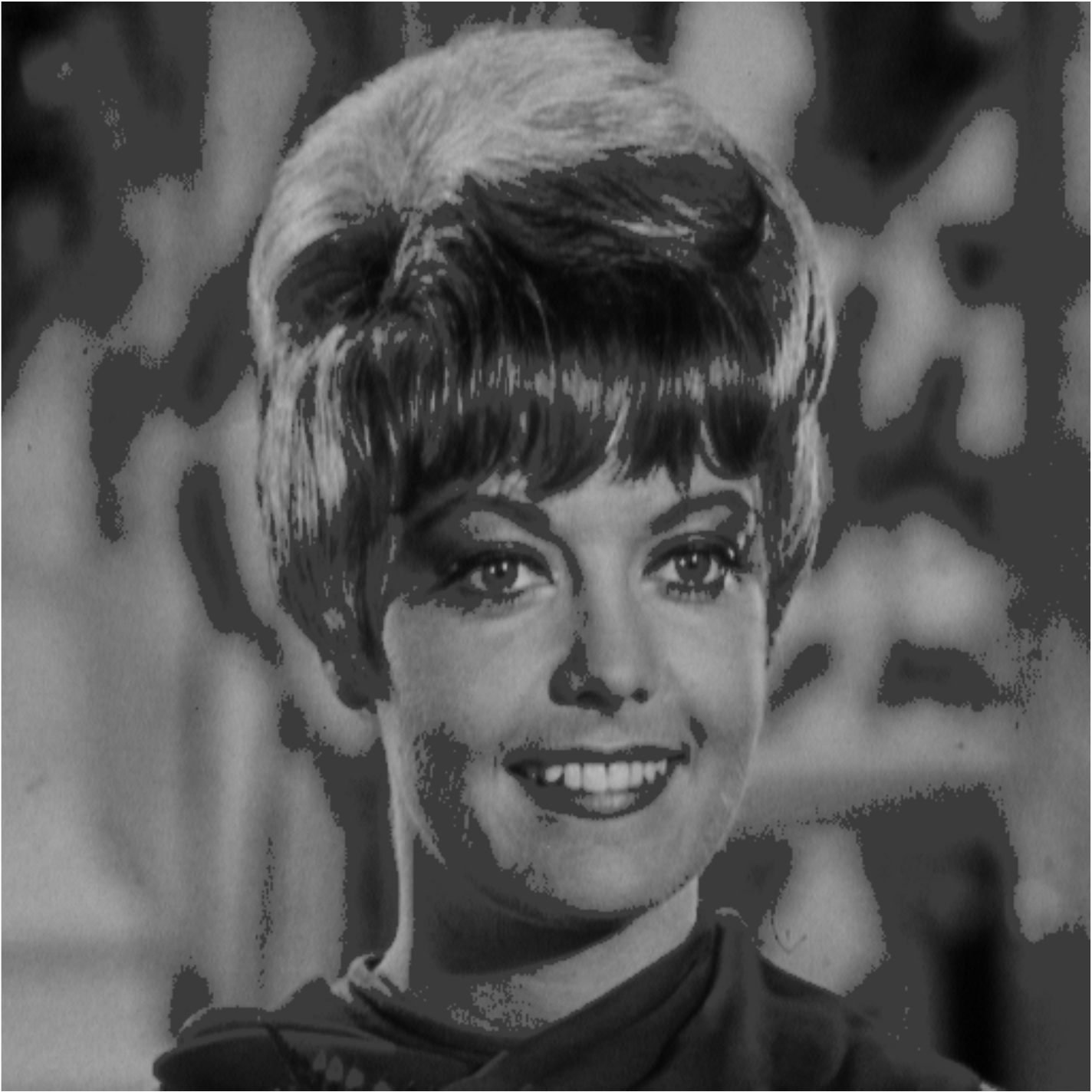}} \:
	\subfloat[threshold=3 \label{fig:Zeldath3}]
	{\includegraphics[scale=0.2]{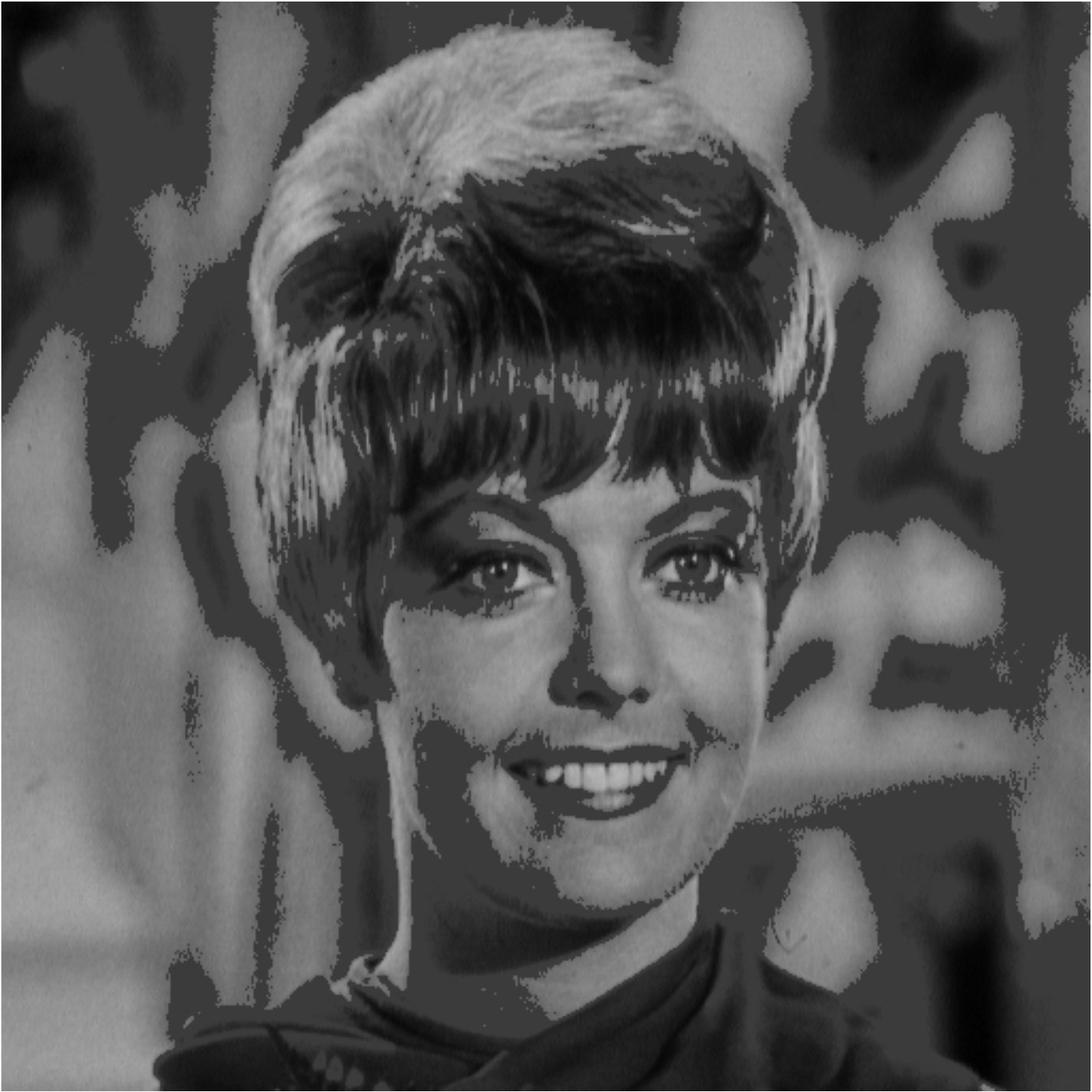}} \:
	\subfloat[threshold=4 \label{fig:Zeldath4}]
	{\includegraphics[scale=0.2]{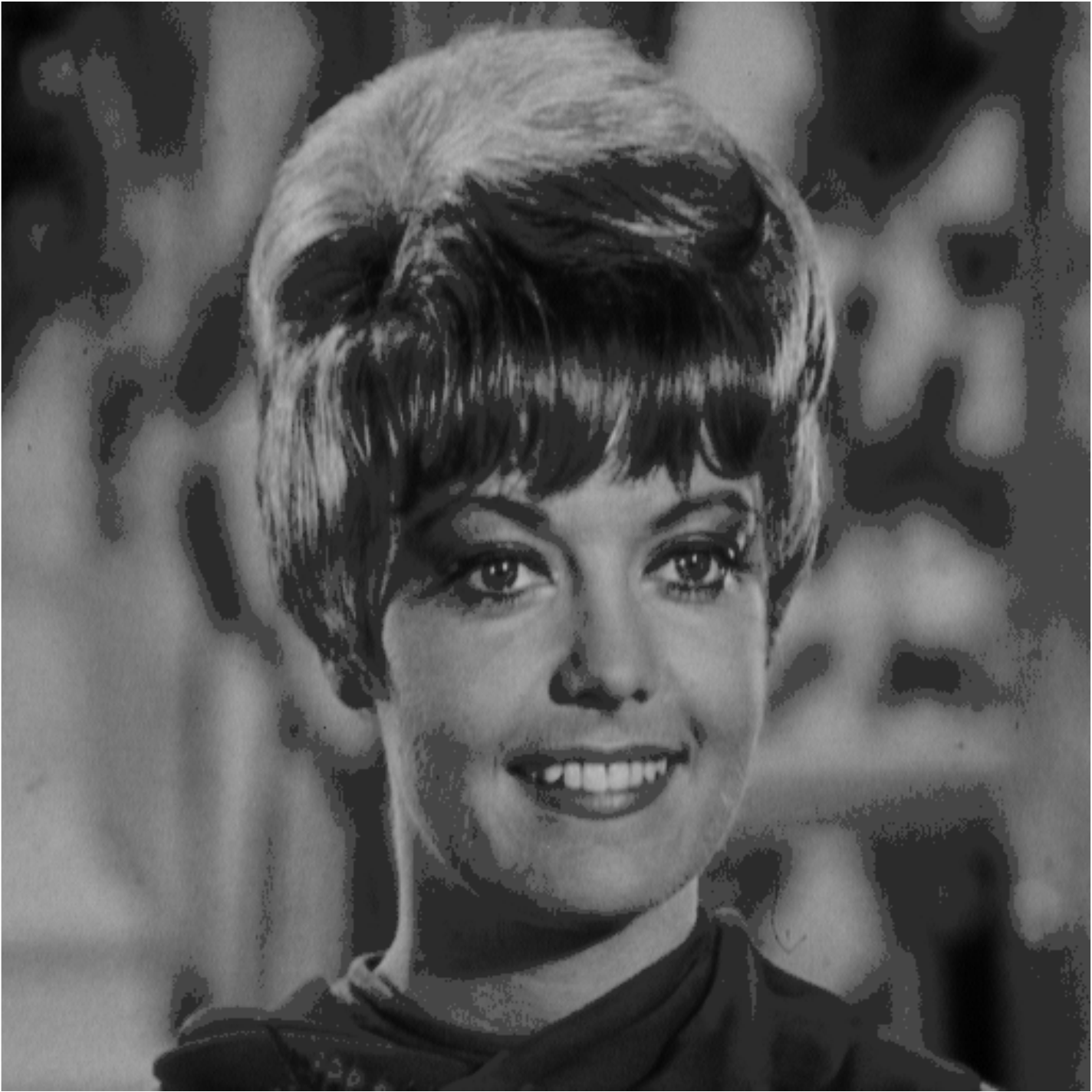}} \:
	\subfloat[threshold=5 \label{fig:Zeldath5}]
	{\includegraphics[scale=0.2]{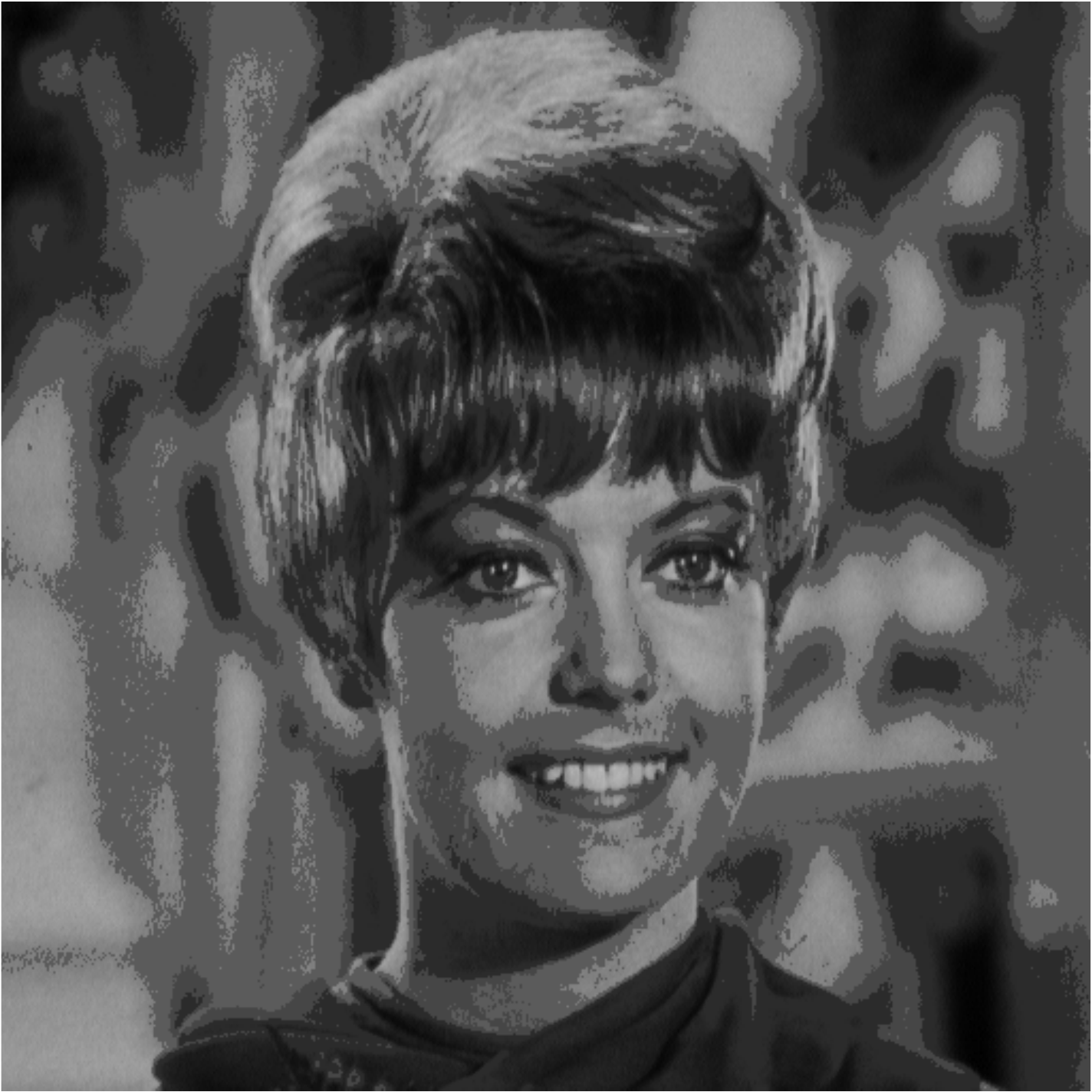}}
	
	\subfloat[threshold=2 \label{fig:Zeldath2Histogram}]
	{\includegraphics[scale=0.2]{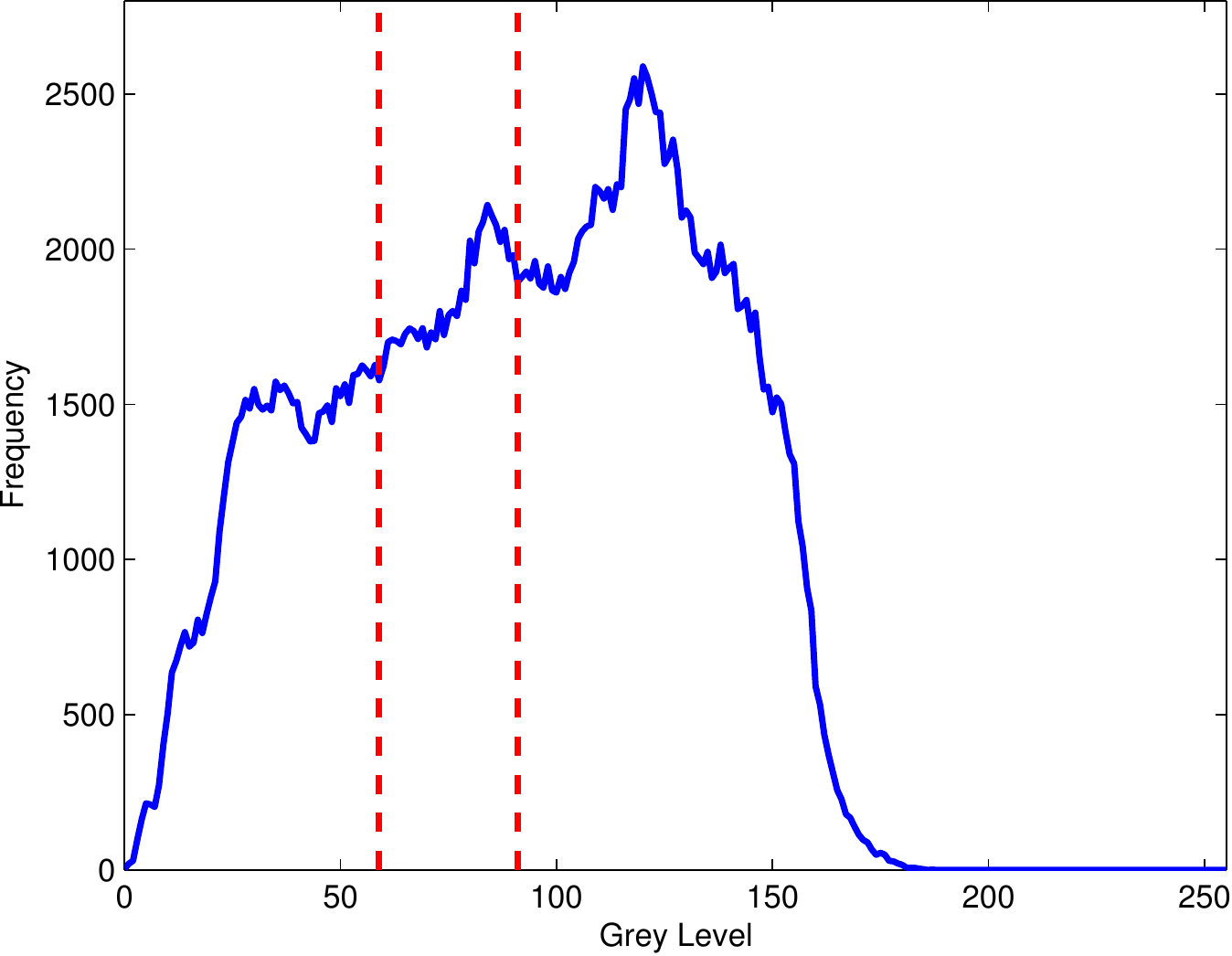}} \:
	\subfloat[threshold=3 \label{fig:Zeldath3Histogram}]
	{\includegraphics[scale=0.2]{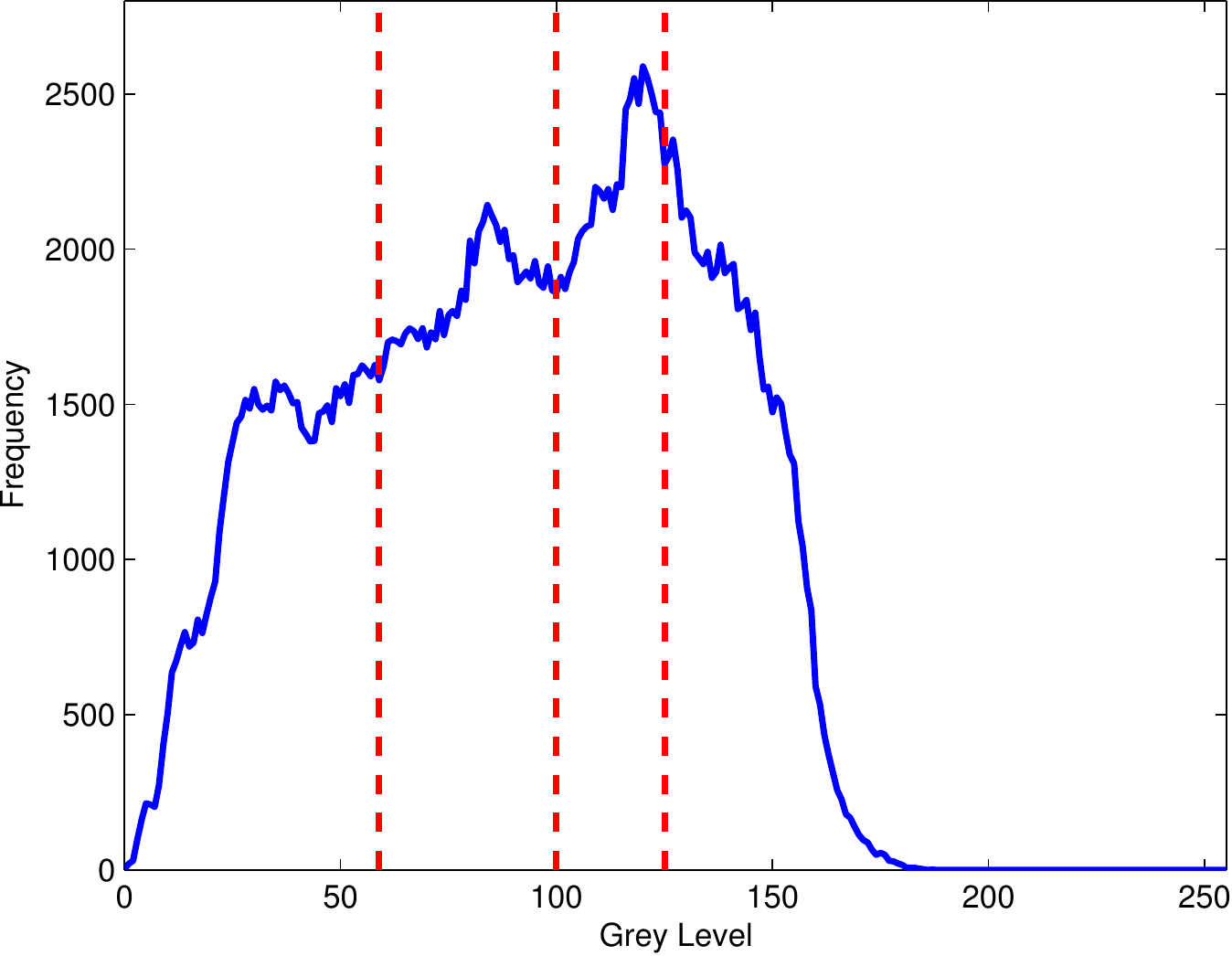}} \:
	\subfloat[threshold=4 \label{fig:Zeldath4Histogram}]
	{\includegraphics[scale=0.2]{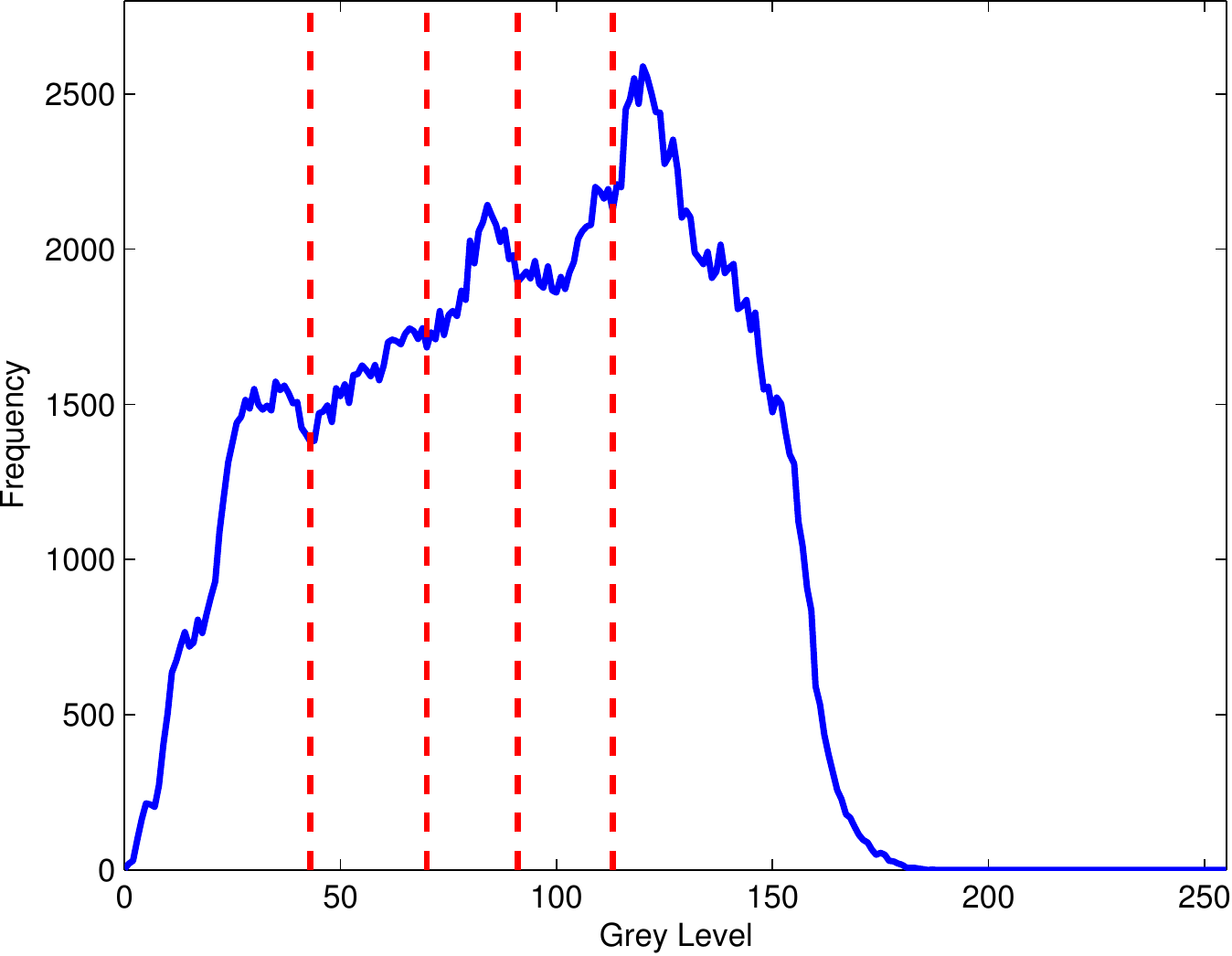}} \:
	\subfloat[threshold=5 \label{fig:Zeldath5Histogram}]
	{\includegraphics[scale=0.2]{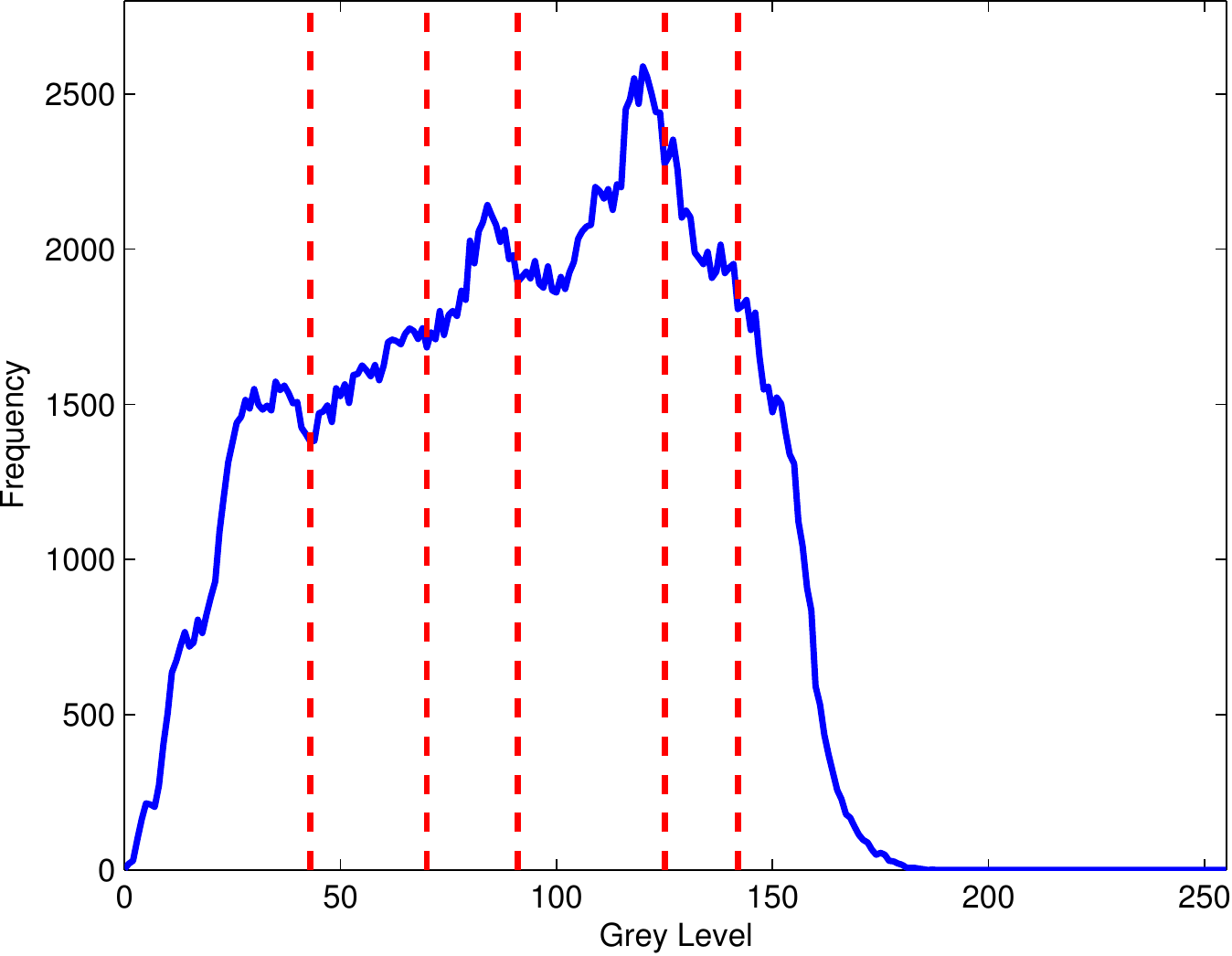}}                                    
	\caption{Result obtained using our approach on the benchmark Boat, Finger-print and Blonde. \label{fig:thresholdingContd2}}
\end{figure*}

Figures \ref{fig:thresholding}, \ref{fig:thresholdingContd}, and \ref{fig:thresholdingContd2} show the outputs of the segmented image and the histogram showing thresholds obtained using AMTIS. The algorithm ensures that chosen thresholds are some valley point in the histogram. We see that images segmented using AMTIS can be easily perceived by human eyes.

\section{Conclusions}\label{sec:conclusion}
We propose a heuristic approach for automatically segmenting an image to determine multilevel thresholds by sampling the histogram of a digital image. The algorithm first employs a gradient descent search to evaluate all valley points in the histogram of the input image. Secondly, the histogram is also partitioned into equal-sized regions to determine minimum frequency within each partition. This partitioning of a histogram is done to obtain candidate threshold values by eliminating multiple local valleys within a local region. It also ensures that candidate values are distributed uniformly in a histogram. Finally, in the third step, we emphasis valley points as optimal thresholds, based on a naive clustering approach. We find that such a naive approach is not very efficient for some benchmarks and required fine-tunning. One such improvement is to select the first candidate threshold instead of taking the mean from the last cluster. As future work, appropriate clustering algorithms can be applied to select optimal threshold values.  We demonstrated that our approach outperforms the popular Otsu's method in terms of CPU computational time. We observed a maximum speed-up of $35.58\times$ and a minimum speed-up of $10.21\times$ on popular image processing benchmarks. The results obtained by our proposed algorithm AMTIS are comparable and better in many cases in comparison to the popular Otu's method. We see that these images, segmented using AMTIS can be easily perceived by human eyes.

\section*{References}

\bibliography{mybiblio}

\end{document}